\documentclass[twoside,letterpaper]{article}

\usepackage[accepted]{aistats2026}
\usepackage[round]{natbib}

\usepackage{graphicx}
\usepackage{subcaption}
\usepackage{algorithm}
\usepackage{algorithmic}
\usepackage[hidelinks]{hyperref}

\usepackage{xcolor}
\usepackage{amssymb}

\newcommand{\AM}{\mathbf{A}}
\newcommand{\nAM}{\mathbf{A^*}}
\newcommand{\ZS}{\mathbf{Z}}
\newcommand{\nZS}{\mathbf{Z^*}}

\newcommand{\beginsupplement}{%
        \setcounter{table}{0}
        \renewcommand{\thetable}{S\arabic{table}}%
        \setcounter{figure}{0}
        \renewcommand{\thefigure}{S\arabic{figure}}%
     }

\begin{document}

%

%

\twocolumn[

\aistatstitle{\MakeUppercase{Regularizing attention scores with bootstrapping}}

\aistatsauthor{ Neo Christopher Chung \And Maxim Laletin}

\aistatsaddress{ University of Warsaw \\ Samsung AI Center, Warsaw \And  University of Warsaw} ]

\begin{abstract}
Vision transformers (ViT) rely on attention mechanism to weigh input features, and therefore attention scores have naturally been considered as explanations for its decision-making process. However, attention scores are almost always non-zero, resulting in noisy and diffused attention maps and limiting interpretability. Can we quantify uncertainty measures of attention scores and obtain regularized attention scores? To this end, we consider attention scores of ViT in a statistical framework where independent noise would lead to insignificant yet non-zero scores. Leveraging statistical learning techniques, we introduce the bootstrapping for attention scores which generates a baseline distribution of attention scores by resampling input features. Such a bootstrap distribution is then used to estimate significances and posterior probabilities of attention scores. In natural and medical images, the proposed \emph{Attention Regularization} approach demonstrates a straightforward removal of spurious attention arising from noise, drastically improving shrinkage and sparsity. Quantitative evaluations are conducted using both simulation and real-world datasets. Our study highlights bootstrapping as a practical regularization tool when using attention scores as explanations for ViT.

Code available: \url{https://github.com/ncchung/AttentionRegularization}
\end{abstract}

\section{INTRODUCTION}

The transformer is a highly influential deep learning architecture that has demonstrated remarkable performance across various domains, including natural language processing \citep{vaswani2017attention,devlin2019bert}, computer vision \citep{dosovitskiy2020vit,wu2020visual}, multi-modal tasks \citep{radford2021clip}, and others. Central to the efficacy of transformers is the attention mechanism, particularly self-attention, which allows the model to weigh the relative significance of each input feature. Despite this built-in capability that could be interpreted as explanations for the model \citep{bahdanau2014nmt,mullenbach2018explainable,serrano2019interpretable,thorne2019explanations}, attention scores generated by these mechanisms can be noisy and often lack sparsity, resulting in poor interpretability and inefficiency. This paper proposes a straightforward yet effective regularization method for attention scores, employing the bootstrapping techniques to suppress noise and enhance meaningful feature selection.

The Transformer architecture is built on self-attention, which enables each token in the input to attend to attend to all other tokens \citep{vaswani2017attention}. This mechanism is expressed as a weighted sum of values, driven by the interactions between query, key, and value projections of the input data. Despite its ability to capture dependencies and relative importances regardless of their distance within the sequence, naive attention scores tend to be diffuse and overly distributed across many features, reducing the model’s ability to highlight the most pertinent features for decision-making. This widespread fat-tailed distribution necessitates regularization techniques to refine the attention scores and improve the model’s interpretability and performance. 

Bootstrapping, a resampling technique in statistics, provides a data-driven approach to estimate the distribution of a statistic by repeatedly sampling with replacement from the data. When a statistical distribution is not well characterized, the bootstrap could prove to be advantageous due to its flexibility and capacity to assess variability and uncertainty in model predictions \citep{efron1979bootstrap,efron1994bootstrap}. In our proposed method, we leverage bootstrapping to generate a baseline distribution of the attention scores by resampling input features to isolate and understand spurious attention attributable to noise rather than informative features. By establishing this distribution, we can estimate the significance of attention scores through p-values and calculate local false discovery rates (lFDR), thereby enhancing the model’s interpretability and reliability from an empirical Bayes perspective \citep{efron2001empirical,storey2001bayesian}.

To validate our approach, we conduct a series of simulation studies that quantitatively assess the accuracy and effectiveness of the proposed regularization method. Our experiments demonstrate how the regularized attention mechanism consistently enhances model performance across several applications. Furthermore, qualitative evaluations reveal marked improvements in the model’s ability to focus on salient features, providing clearer interpretability and richer insights into the decision-making process. These empirical findings underscore the effectiveness of bootstrapping as a tool for refining attention mechanisms within the Transformer architecture.

The remainder of this paper is organized as follows: Section~\ref{sec:related_works} discusses related works in the field, detailing existing methods of attention score regulation and bootstrapping applications. In Section~\ref{sec:method} we elaborate on the proposed, detailing the application of bootstrapping to attention scores, 
and describe in detail the methodology of our simulation study. Section~\ref{sec:results} presents the evaluation results, highlighting the improvements in both accuracy and model interpretability. Finally, Section~\ref{sec:conclusion} offers a discussion on the implications of our findings and potential avenues for future research.  

\section{RELATED WORKS}
\label{sec:related_works} 

In the transformer architecture \citep{vaswani2017attention}, attention mechanisms provides relative importance of different input features when transformers generate outputs, emphasizing the significance of attention scores in capturing contextual relationships within data. Attention allows transformer models to weigh the relevance of various input features dynamically, leading to improved performance in language, vision, and other domains. The effectiveness of attention scores is further enhanced by their potential ability to facilitate explainability, enabling researchers to understand which parts of the input data contribute most to the model’s predictions \citep{bahdanau2014nmt,mullenbach2018explainable,serrano2019interpretable,thorne2019explanations}. However, whether the attention mechanisms provides comprehensive explanations is debated \citep{serrano2019interpretable, wiegreffe2019attention, jain2019attention}.

In most of modern transformer models, a multiple set of attention mechanisms are used, such that multi-head attentions may learn different types of relationships between input representations. Different types of relationships may even represent semantic across a single hidden layer. Nonetheless, some studies have found that not all attention heads are necessary or important. Thus, there are methods to estimate importance of attention heads and prune the neural networks \citep{michel2019sixteenheads,voita2019analyzing,molchanov2019importance}. While we have similar motivation, our approach is orthogonal and complementary to the neural network pruning. Specifically, they want to identify and potentially remove unimportant attention heads. Our methods regularize attention scores, at a more fine-grained level (i.e., pixel-level), that makes of each attention heads. Our motivation to regularize attention scores arises from a need to improve interpretability of transformers, which is broadly called explainable artificial intelligence (XAI). 

For convolutional neural networks (CNNs), saliency maps\footnote{Also called importance, attribution, or relevance maps. When writing about general architectures and different methods, we call them an importance estimator.}   back-propagate gradients from the output score to the input features \citep{simonyan2013saliency}. Extensions of saliency maps – modified back-propagation or aggregation of multiple saliency maps – are developed due to the noisy and unreliable nature of vanilla saliency maps in certain conditions \citep{sundararajan2017axiomatic, selvaraju2017gradcam, smilkov2017smoothgrad, shrikumar2016deeplift}. Gradient-based methods are also developed and incorporated to improve attention scores of transformers in language \citep{voita2019analyzing,abnar2020quantifying} and computer vision \citep{chefer2021interpretability,brocki2019,brocki2024class}. The proliferation of saliency maps and related methods has then spawned a number of evaluation frameworks 
\citep{brocki2023fidelity, brocki2023fpa}. In most of these studies, regularization is implicitly prized as it related to interpretability, feature selection, visual contrast, and human-centric evaluation.

In machine learning, regularization techniques are utilized to mitigate overfitting and improve explainability, that are especially well developed for weights of deep neural networks (DNNs). Classically, network pruning was used to improve generalization \citep{lecun1989optimal,thodberg1991pruning}. Popular regularization methods, such as L$_1$ \citep{tibshirani1996lasso}, L$_2$ \citep{hoerl1970ridge}, and elasticnet \citep{zou2005elasticnet} regularization add penalties to the loss function to constrain the model’s complexity. L$_2$ penalties can help regularize model weights towards 0 \citep{krogh1991decay,ash2020warmstart} or initialization \citep{kumar2023plasticity}. Such modified loss functions has been also used for feature selection \citep{rahangdale2019dnn,lemhadri2021lassonet}. Dropout randomly deactivates neurons during training to promote robustness and reduce co-adaptation among units \citep{srivastava2014dropout}. MixUp trains a DNN by using similar samples as data augmentation \citep{zhang2018mixup}. Total variation regularization and related methods have been proposed \citep{bredies2010tgv,ren2013fractional,kobler2020tdv,zhang2023regularization}. Peer-regularized networks (PeerNet) and related methods leverage dependent samples to increase the model performance and mitigate adversarial attacks \citep{svoboda2019peernets,sun2019patchlevel}. The theoretical foundations of regularization in deep learning have also been an area of active research \citep{achille2018information,taheri2020statistical}. Regularization methods not only enhance the stability of the training process but also improve the model’s ability to generalize \citep{hernandez2018augmentation,zhang2020local}. 

By resampling with replacement, the bootstrap mimics the data generation process which help estimate the sampling distribution of a statistic without relying on traditional assumptions \citep{efron1979bootstrap,efron1994bootstrap}. The computational advancements in recent years have made it feasible to apply the bootstrap methods to large datasets and to increase the downstream model performance \citep{kleiner2014scalable}. Bootstrapping has been shown to be effective in scenarios where labeled data is scarce \citep{chai2024deeplearning}. The bootstrapping has been applied to neural processes (NP) to improved the classification and regression models to be more robust and generalizable. Particularly, bootstrapping neural processes (BNP) estimate the residuals and obtaining the paired residual-bootstrapped samples, followed by bagging. Instead of sampling the residual of the data, NP with stochastic attention replaces deterministic attention modules with Bayesian Attention Modules (BAM) to sample attention weights from the Weibull distribution. By modeling the uncertainty directly, they show that the stochastic attention improves NP compared to the deterministic baselines.

In contrast, we aim to regularize attention scores by directly generating or resampling the input features. To the best of our knowledge, we are introducing for the first time the use of bootstrapping to regularize attention scores.

\section{METHODS AND MATERIALS}
\label{sec:method}

\subsection{Proposed Methods}
We propose \emph{Attention Regularization} by the Bootstrap (ARB) to better estimate attention scores and enhance the explainability of transformers. In brief, we bootstrap an input sample $\textbf{X}$ to obtain bootstrapped samples $\textbf{X}^*$. With a pre-trained transformer model $f$, we compute attention scores that would arise by chance $\textbf{A}^*$. Attention scores derived from a bootstrap sample, which is constructed to not contain systematic structure, do not exhibit meaningful patterns. By estimating a null distribution of such unimportant attention scores through repeated bootstrapping, we can evaluate the observed attention scores. This comparison allows us to determine whether the observed attention scores significantly deviate from null attention scores that would be expected by random chance, thus identifying truly relevant features. 

\subsubsection*{The Bootstrap} 

\begin{figure}
    \centering
    \includegraphics[width=0.45\textwidth]{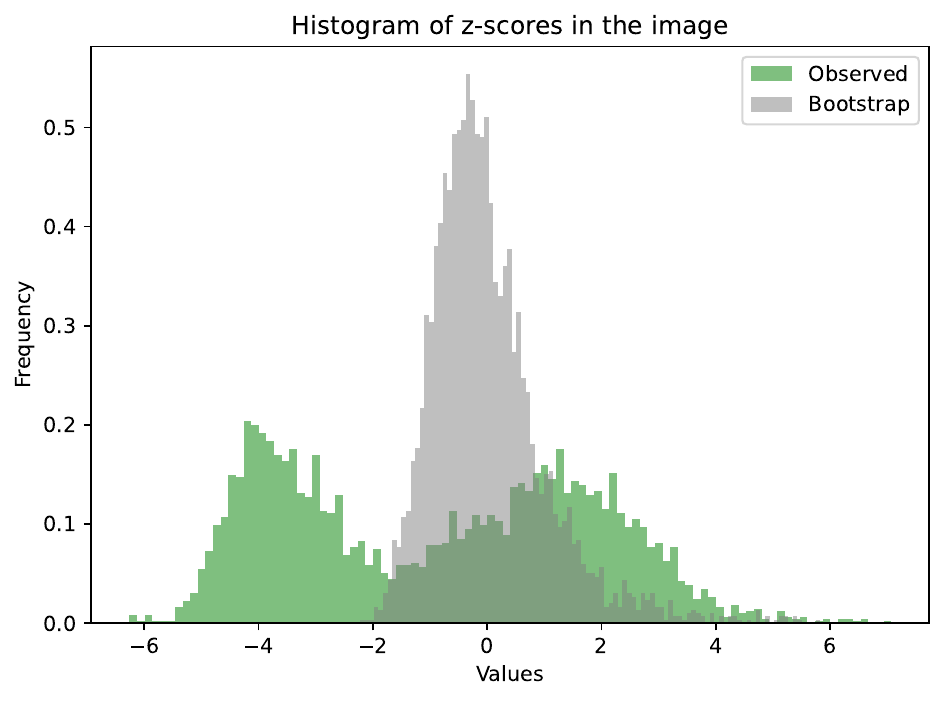}
    \caption{Histogram of $z$-statistics from the observed attention score sample corresponding to one of the images we use in our analysis and from the derived bootstrap sample. 
    }
    \label{fig:example_histograms}
\end{figure}

Our bootstrapping approach offers flexibility in its implementation, allowing for both (a) nonparametric and (b) parametric bootstrap sampling techniques (Step 2 in Algorithm~\ref{alg:uncertainty_attention}). In the non-parametric bootstrap, input features are randomly resampled with replacement \citep{efron1979bootstrap,efron1994bootstrap}. This relies on the robustness provided by resampling the actual dataset, thereby preserving first-order characteristics of observed samples. In contrast, the parametric bootstrap constructs input features from a distribution, effectively simulating feature sets under a specific probabilistic model assumption \citep{davison1997bootstrap}. Specifically, we demonstrate constructing bootstrap samples by drawing values from a normal distribution with mean and variance corresponding to that of each RGB channel of an image (Fig.~\ref{fig:example_histograms}).

\subsubsection*{Uncertainty Measures}

Attention scores computed from the observed sample $\AM$ and the bootstrap samples $\nAM$ are converted
to $z$-statistics $\ZS$ and $\nZS$, respectively. The mean and standard deviation of the bootstrap sample $\nAM$ are estimated by
\begin{align}
\hat{\mu} &= \frac{1}{mB} \sum_{i=1}^{m} \sum_{j=1}^{B} a^{\ast}_{ij}, \\
\hat{\sigma}^2 &= \frac{1}{mB} \sum_{i=1}^{m} \sum_{j=1}^{B} (a^{\ast}_{ij} - \hat{\mu})^2.
\end{align}
where index $i$ goes over all the attention scores in the image sample of size $m$ and index $j$ goes over all bootstrap samples with the total number $B$. Then, attention scores are converted to $z$-statistics as follows:
\begin{equation}
z_{i} = \frac{a_{i} - \hat{\mu}}{\hat{\sigma}}, 
\quad
z^{\ast}_{ib} = \frac{a^{\ast}_{ib} - \hat{\mu}}{\hat{\sigma}}.
\label{eq:zstat}
\end{equation}

We measure the uncertainty of observed attention scores by comparing the distributions $\ZS$ and $\nZS$ (Fig.~\ref{fig:example_histograms}). 
First, note that the $z$-statistics are standardized by attention scores from the bootstrap samples 
(mean $0$ and standard deviation $1$), such that their magnitudes and signs are meaningful. 
For example, negative $z$-statistics indicate attention scores smaller than the mean that would arise by chance. 

Second, given an attention score, the statistical significance ($p$-value) is estimated as the probability 
that $z_{i}$ is lower than a value in the bootstrap distribution established by $\nZS$:
\begin{equation}
p_{i} = \frac{\#\{ (z_j + z^{\ast}_{jb}) > z_{i} ; \; j=1,\ldots,m,\; b=1,\ldots,B \}}{mB}.
\label{eq:pvalue}
\end{equation}

Third, we compute the local false discovery rate (LFDR), defined as the probability that a given 
attention score is attributable to the bootstrap distribution~\citep{efron2001empirical,storey2001bayesian}. 
LFDR can be computed directly from the observed $z$-statistics as
\begin{equation}
l_{i} = \Pr(z_{i} = 0 \mid \nZS),
\label{eq:lfdr}
\end{equation}
which assumes the most conservative hyper-parameter $\pi_{0} = 1$. 
More accurate estimates of the probability of null scores $\pi_{0} \in [0,1]$ can be obtained 
following the approach of~\cite{storey2003genomewide}.





\begin{algorithm}[tbh]
\caption{Estimating Uncertainty of Attention Scores}
\label{alg:uncertainty_attention}
\begin{algorithmic}[1]
\REQUIRE Input sample $\textbf{X}$ and pretrained model $f$
\STATE Compute the observed attention scores 
    $\AM = (a_{1}, \ldots, a_{m})^{T}$ using model $f$
\STATE Create a bootstrap sample $\textbf{X}^{\ast}_{b}$ by
    \begin{enumerate}
        \item a) resampling $\textbf{X}$ with replacement, or
        \item b) sampling from a parametric distribution
    \end{enumerate}
\STATE For $\textbf{X}^{\ast}_{b}$, compute the null attention scores 
    $\textbf{A}^*_b = (a^{\ast 1}_{b}, \ldots, a^{\ast m}_{b})^{T}$
\STATE Repeat steps 2--3 for $b = 1, \ldots, B$ to obtain 
    $B \times m$ null attention scores 
    $\nAM = (a^{\ast}_{1}, \ldots, a^{\ast}_{B})$
\STATE Estimate mean $\mu$ and standard deviation $\sigma^{2}$ of $\nAM$
\STATE Compute standardized $z$-statistics for $\AM$ and $\nAM$ according to Eq.~\eqref{eq:zstat}
\STATE Compute $p$-values for the observed attention scores from Eq.~\eqref{eq:pvalue}
\STATE Compute LFDR for the observed attention scores from Eq.~\eqref{eq:lfdr}
\end{algorithmic}
\end{algorithm}

\subsubsection*{Regularization Techniques}

Based on uncertainty statistics, we introduce several shrinkage approaches. As an initial step, we apply the simple $z$-statistics by setting to zero any attention scores with $z \leq 0$ and then apply one of the methods described below\footnote{This $z$-statistics procedure can be regarded as a regularization technique on its own with a threshold $z_{\rm th}$. While we use $z_{\rm th} = 0$ here, identical behaviors can be achieved by considering $z$-statistics in combination of p-values and LFDRs.}.

While we present those methods for clarity, it is possible to develop related shrinkage methods. 

\textbf{$p$-thresholding:} We threshold attention scores based on $p$-values using a threshold $p_{\rm th}$

\begin{equation}
\tilde{a}_i^p = \begin{cases}
    0, & \text{if } p_i > p_{\rm th}\\
    a_i, & \text{otherwise}
\end{cases}
\end{equation}

\textbf{$l$-thresholding:} Similarly, $l_i$ is used to regularize attention scores.

\begin{equation}
\tilde{a}_i^l = \begin{cases}
    0, & \text{if } l_i > l_{\rm th}\\
    a_i, & \text{otherwise}
\end{cases}
\end{equation}

In the aforementioned $p$- and $l$-thresholding methods, it becomes apparent that choosing $p_{\rm th}$ and $l_{\rm th}$ controls regularization. To automate threshold selections, we 
can estimate the proportion of null scores $\pi_{0} \in [0,1]$ using the method of \cite{storey2003genomewide} and zero out attention scores below the corresponding percentile. We call it the \textbf{$\pi_0$ thresholding}.



\begin{figure*}[th!]
\centering
\includegraphics[width=\linewidth]{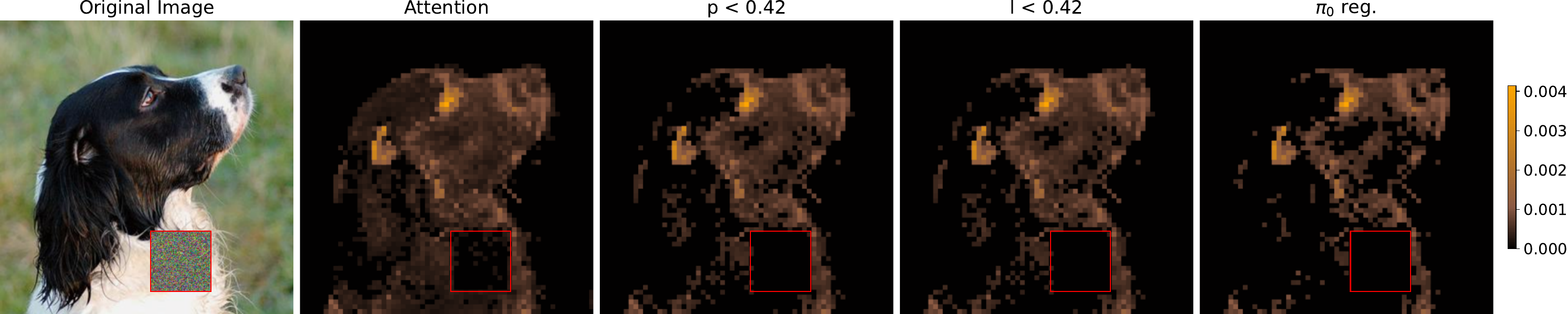}\\[0.5cm] 
\includegraphics[width=\linewidth]{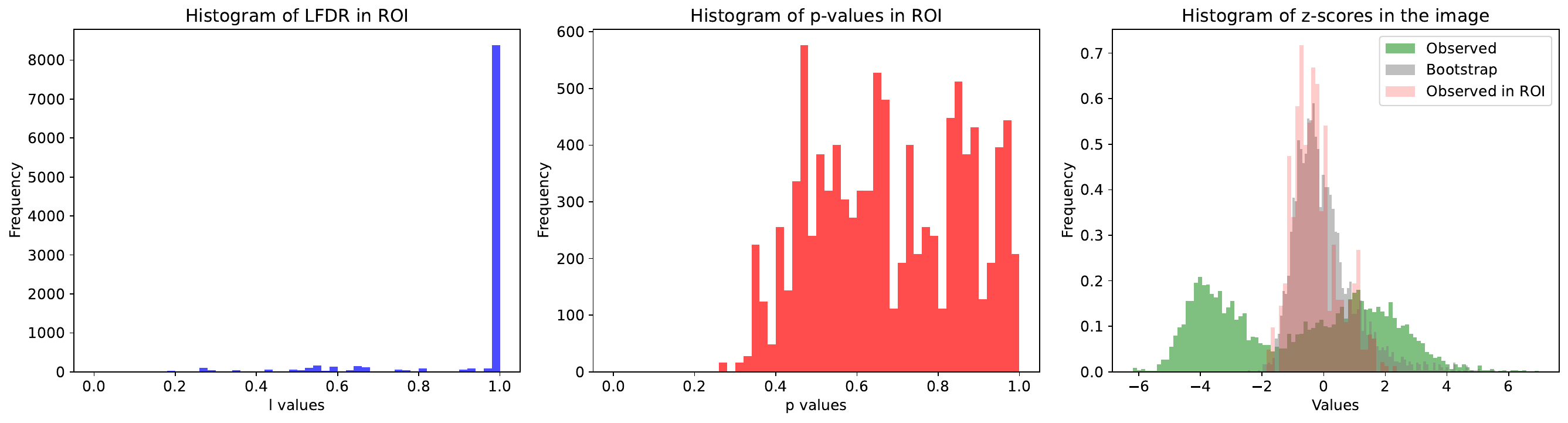}
\caption{Example of a perturbed image (\texttt{n02102040\_821.JPEG}) with the attention map before and after regularization via different shrinkage methods: $p$-thresholding and $l$-thresholding with thresholds set at the 10-th percentile for $p$-values and LFDR respectively.}
\label{fig:perturbed_image_attention_and_hist}
\end{figure*}

\subsection{Quantitative Evaluation}
\label{sec:noise_simulation}

\subsubsection*{Simulation Studies}
\label{sec:simulation_method}
We conduct simulation studies to create random i.i.d. pixels and evaluate the corresponding attention scores. By knowing which pixels are purely noise, we can evaluate whether regularization helps to obtain attenuated attention scores. After applying the proposed regularization methods, we quantify shrinkage, sparsity, and suppression factor. Furthermore, sensitivity and specificity are measured while varying the regularization thresholds.


The images are perturbed by an injection of a $100 \times 100$ size square consisting of generated noise pixels into a random location in the image such that the square fits in the image completely. Alternatively, we have also conducted simulation study with a different noise pattern which is randomly scattered around the image (diffuse) (Appendix~\ref{app:diffuse_noise}). Thus, we can easily frame the noisy part of the image, which we call the region of interest (ROI). Pixels are sampled independently for each channel from a normal distribution whose mean ($
\mu$) and standard deviation ($\sigma$) match those of the corresponding image pixel values. The DINO ViT backbone \citep{caron2021emerging} with the patch size of $8 \times 8$ pixels and the backbone fine-tuned on the ImageNet subsets as described in \cite{brocki2024class}. 
In Appendix \ref{app:dino_v2} we show how the proposed regularization method works with a different type of ViT (DINOv2 with the patch size of $14 \times 14$).

The attention map is constructed from the attention weights for the CLS token extracted from the last transformer encoder layer of the ViT and averaged over all attention heads. The attention map consisting of patches is further refined to match the image size using the nearest-neighbor interpolation scheme and rescaled using min-max normalization. We construct our bootstrap sample from a normal distribution with mean and standard deviation corresponding to those of distribution of pixels in each RGB channel. We consider the effects of using non-parametric bootstrap method in Appendix~\ref{app:non_parametric_bootstrap}: both parametric and non-parametric bootstrap methods result in highly comparable operating characteristics. Generally we use one bootstrap sample $(B = 1)$ 
and study the impact of other values of $B$ hyperparameter (as well as the width of the distribution) in Appendix~\ref{app:hyperparameters}.




The patches may have an internal \textit{structure} that are quite different from the structure of the rest of the image in some cases, while the null attention scores are computed for a homogeneous bootstrap image. Hence in general the distribution of attention scores in ROI deviate from the null distribution. Since our goal is to simulate the noise in the images, we filter out those perturbed images that has large mean $z$ scores in ROI and concentrate on the region $|z| \leq 1$. In Appendix~\ref{app:mean_z_scores_roi} we show the distribution of mean $z$ scores for all the perturbed images in our study, as well as the distribution of selected images in the $z$ range of interest. 



 
\subsubsection*{Analysis and Evaluation Metrics}

To analyze the regularization effectiveness in our simulation quantitatively we rely on the percentiles $q$ of attention scores in ROI w.r.t. the scores in the rest of the image. The \textit{mean percentile} $\langle q \rangle$ of scores in ROI before and after regularization ($\langle \tilde{q} \rangle$) is a point estimate that reflects the strength of the regularization method for different noise levels. 

To evaluate the power of different shrinkage methods cumulatively over sets of images we introduce the \textit{average suppression factor} $D$, which is computed as a sum of mean percentiles in ROI after regularization divided by the sum of mean percentiles before regularization 

\begin{equation}
    D = \frac{\sum_k \langle \tilde{q} \rangle_k}{\sum_k \langle q \rangle_k} \, ,
    \label{eq:D_factor}
\end{equation}
where index $k$ denotes a single image and the sum goes over all the images in a set. The closer $D$ is to 1 the worse is the efficiency of the shrinkage method on average for a given set, while $D = 0$ is the perfect case in terms of noise reduction. 

To measure how regularization affects the rest of the image, which is treated as a positive signal, we employ the concepts of \textit{sensitivity} and \textit{specificity}. We define sensitivity as the measure of how well a shrinkage method removes the noise (ROI) for a given image and calculate it according to the following formula

\begin{equation}
\text{Se} = 1 - \frac{\sum_i \tilde{a}_i}{\sum_i a_i} \Bigg|_{\rm ROI} ,
\end{equation}
where $a$ denotes the attention scores before regularization and $\tilde{a}$ the attention scores after regularization. Thus, the sensitivity is equal to 1 if the noise is completely removed and is equal to 0 if it remains exactly the same. In its turn the specificity is defined as the measure of how intact is the rest of the image after regularization and is calculated as follows

\begin{equation}
    \text{Sp} = \frac{\sum_i \tilde{a}_i}{\sum_i a_i} \Bigg|_{\rm Rest} .
\end{equation}
We use these metrics to illustrate the balance between noise reduction and loss of important features controlled by the regularization parameter $p_{\rm th}$ or $l_{\rm th}$.

\subsection{Application to Medical Images}
\label{sec:medical_data_applications}

The proposed \emph{Attention Regularization} can be used to denoise attention maps for various real-world applications. Besides the natural images in the ImageNet, we apply these regularization techniques to the images from lung cancer screening dataset IQ-OTH/NCCD \citep{alyasriy2020diagnosis} containing 1097 images of normal, malignant and benign cases (a resolution of $512 \times 512$). In Appendix~\ref{app:applications} we show several examples of the attention maps obtained with different shrinkage methods for different types of images and also briefly address the efficiency of regularization in Fig.~\ref{fig:reg_efficiency_medical}.

\section{RESULTS}
\label{sec:results}

In Fig.~\ref{fig:perturbed_image_attention_and_hist} we demonstrate a simulation study example from the Imagenette validation dataset and visualize the corresponding attention maps before and after regularization. We further show the distributions of the quantities that are used for regularization, namely LFDR, $p$-values and $z$-scores. The top row of the figure shows from left to right the original image with a noise patch at $(350,250)$, the attention map extracted for that image followed by the attention map after various regularization methods: $p$-thresholding, $l$-thresholding and $\pi_0$ regularization. We use $p_{\rm th} = 0.42$ and $l_{\rm th} = 0.42$ that correspond to the 10th-percentile of $p$-values and LFDR in ROI. 

The mean percentile of attention scores in ROI w.r.t. the whole attention map before regularization is $16.24$, while for the displayed shrinkage methods these values are $[1.90, 1.90, 0.54]$ in the corresponding order. The bottom row shows from left to right the histograms of LFDR values in ROI, the histogram for $p$-values in ROI and the superimposed histograms of $z$-scores from the attention scores (green), from the bootstrap distribution (gray) and from the attention scores in ROI (red), which correspond to noise. Several other examples of simulation and the corresponding attention maps (including the ones that show the results of regularization for different threshold values) can be found in Appendix~\ref{app:noise_examples}.

\begin{figure*}[t]
    \centering
    \begin{subfigure}{0.47\textwidth}
        \centering
        \includegraphics[width=\textwidth]{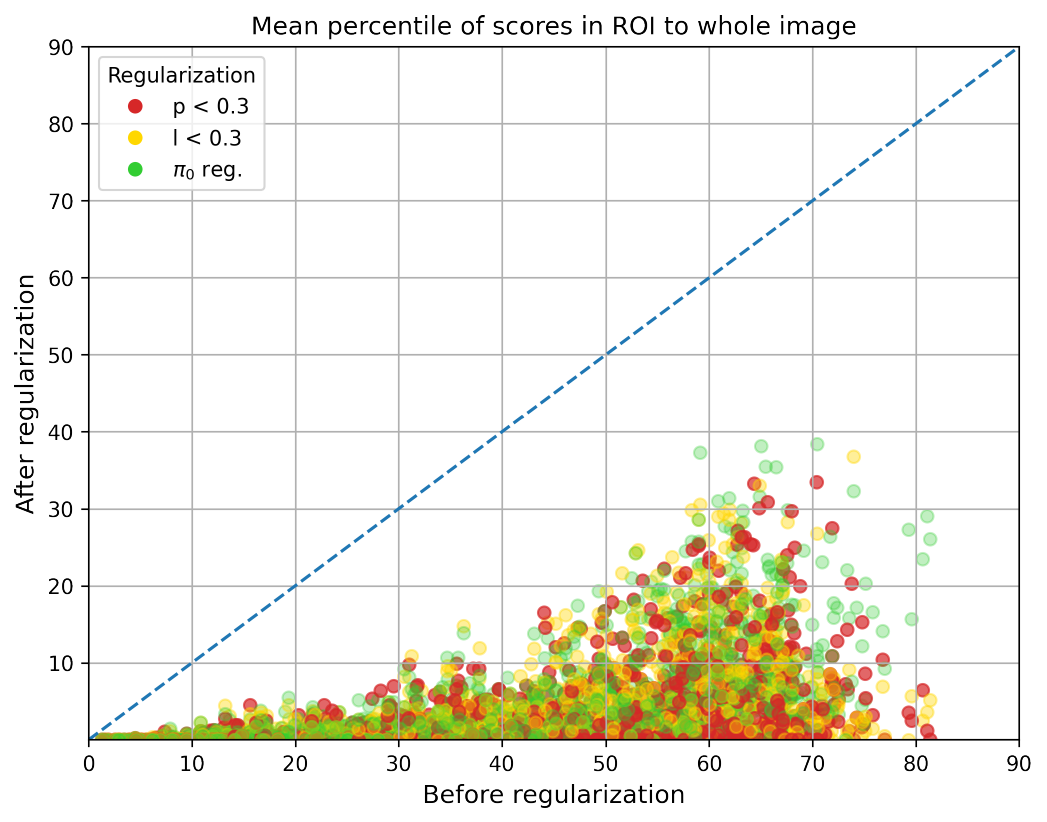}
        \caption{}
        \label{fig:mean_perc_roi_reg}
    \end{subfigure}
    \hfill
    \begin{subfigure}{0.47\textwidth}
        \centering
        \includegraphics[width=\textwidth]{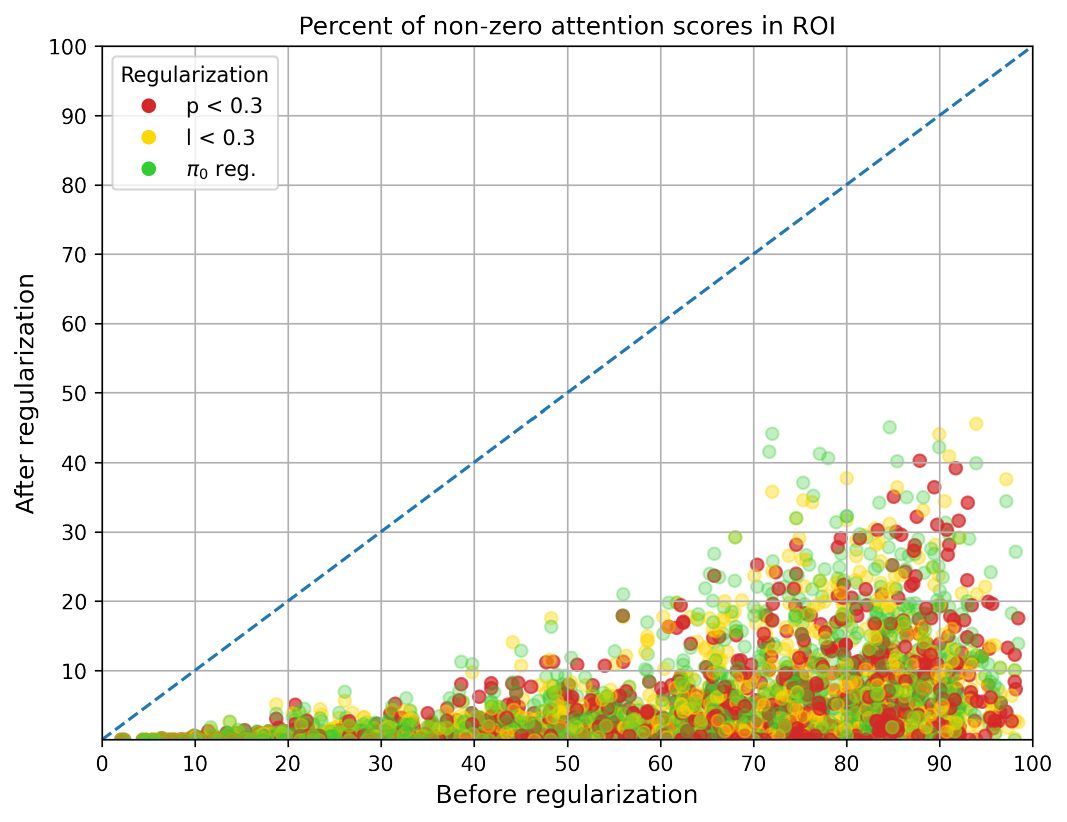}
        \caption{}
        \label{fig:non_zero_roi_reg}
    \end{subfigure}
    \caption{Regularization efficiency in ROI for various images from all the categories of the Imagenette validation subset expressed in terms of: a) mean percentile of scores in ROI w.r.t. the whole image; b) percentage of non-zero attention scores in ROI. Each dot denotes the results of using a method for a perturbed image. Thresholding values $p_{\rm th} = 0.3$ and $l_{\rm th} = 0.3$. The blue dashed line indicates no regularization.}
    \label{fig:reg_efficiency}
\end{figure*}

\begin{figure}[h]
    \centering
    \includegraphics[width=0.5\textwidth]{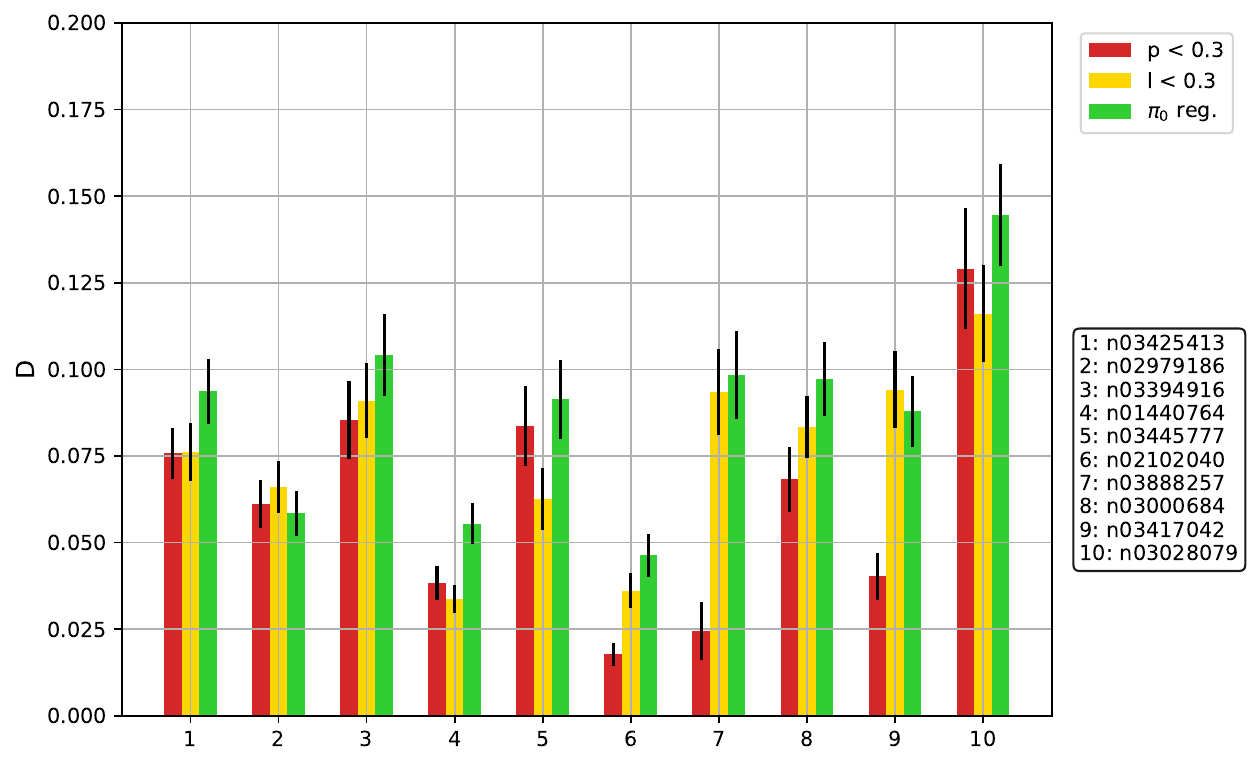}
    \caption{The suppression factor $D$ for different categories of the Imagenette validation subset and different shrinkage methods. Lower $D$ corresponds to better regularization.
    }
    \label{fig:D_histogram}
\end{figure}
In Fig.~\ref{fig:mean_perc_roi_reg} we show the mean percentile of attention scores in ROI with respect to the attention scores in the whole image before and after different regularization methods. Fig.~\ref{fig:non_zero_roi_reg} shows the percentage of non-zero attention scores in ROI before and after these regularization methods. Each dot denotes the results of using one method for one perturbed image. Red dots show the results of $p$-thresholding shrinkage method for $p_{\rm th} = 0.3$, 
yellow dots -- $l$-thresholding with $l_{\rm th} = 0.3$ and green dots -- $\pi_0$-thresholding for $p$-values. The threshold values (except for the $\pi_0$ threshold) are taken as rather arbitrary non-extreme values and do not have any specific meaning. The blue dashed line corresponds to the same values before and after regularization, thus any successful regularization method is expected to produce points below that line. 

Overall, our evaluation suggests that our bootstrap regularization works well 
for different image categories and that the results of different regularization methods correlate with each other. 
This can be also seen in Fig.~\ref{fig:D_histogram}, in which we show the average suppression factor $D$ for different categories and methods. While some categories of images are regularized a bit more efficiently than the others, overall we achieve $\sim 10\%$ noise level w.r.t. the case before regularization. One can notice that different regularization methods give quite similar results as expected since the p-values and LFDR are monotonically increasing. It is worth noting that quantitatively similar results to the ones presented in Figs.~\ref{fig:reg_efficiency} and~\ref{fig:D_histogram} are achieved even if we include noise simulation cases with mean $z > 1$, as demonstrated in Appendix~\ref{app:non_filtered_results}.

Figure~\ref{fig:se_sp_p_reg} depicts the specificity-sensitivity curves for different images that undergo $p$-regularization (left subplot) and $l$-regularization (right subplot) with different threshold values. For this study we extracted 5 random images from each category of Imagenette validation subset (denoted with different colors) which have the mean $z$-score in ROI between $-1$ and $1$. The curves connect the results for 50 threshold values that span between $0$ and $1$.

\begin{figure*}[tbh!]
    \centering
    \begin{subfigure}{0.49\textwidth}
        \centering
        \includegraphics[width=\textwidth]{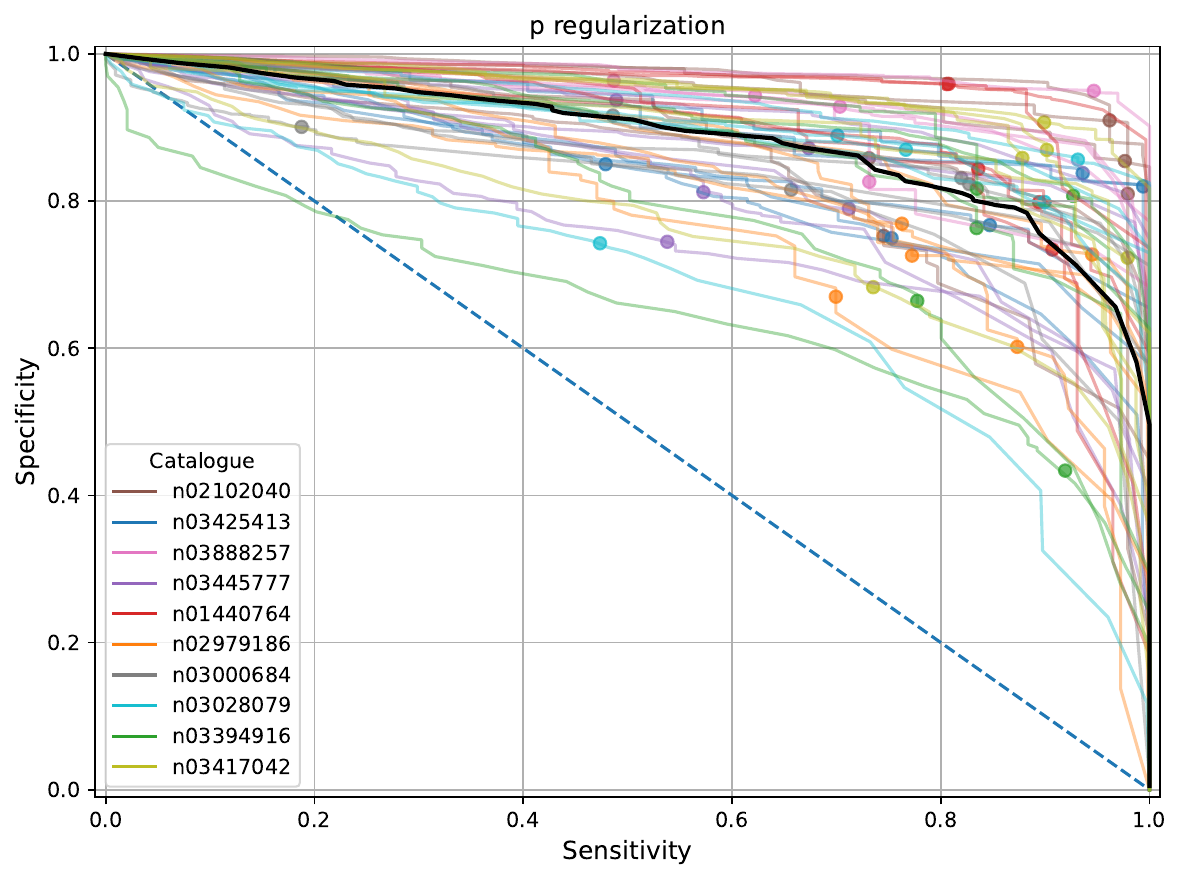}
        \caption{}
    \end{subfigure}
    \hfill
    \begin{subfigure}{0.49\textwidth}
        \centering
        \includegraphics[width=\textwidth]{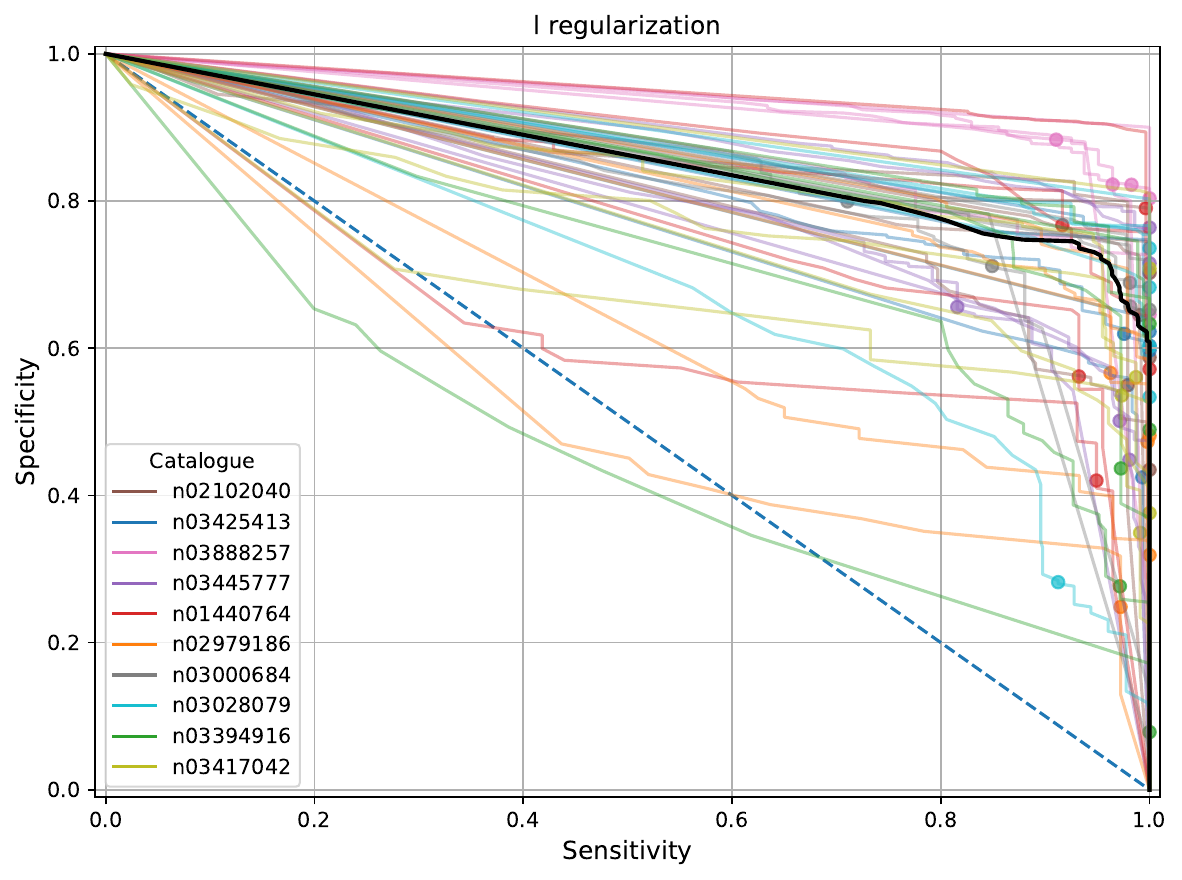}
        \caption{}
    \end{subfigure}
    \caption{Sensitivity vs. specificity curves for 5 random images from each of the categories of the Imagenette validation subset (denoted with different colors) regularized via $p$-thresholding (\textit{left}) and $l$-thresholding (\textit{right}) for 50 values of the corresponding threshold spanning logarithmically from $0$ to $1$. The dots correspond to the $\pi_0$-threshold value for each curve. The black solid line shows the median of all the curves.
    }
    \label{fig:se_sp_p_reg}
\end{figure*}

\begin{figure*}[htb!]
\centering
\begin{subfigure}{\linewidth}
    \includegraphics[width=\textwidth]{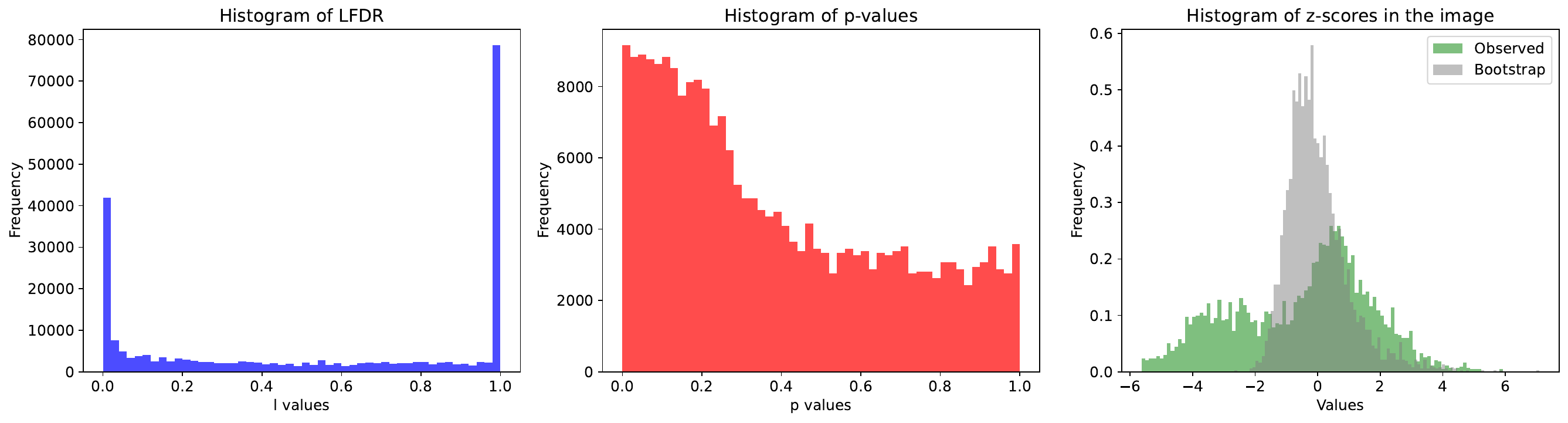}
\end{subfigure}
\vfill
\begin{subfigure}{\linewidth}
    \includegraphics[width=\textwidth]{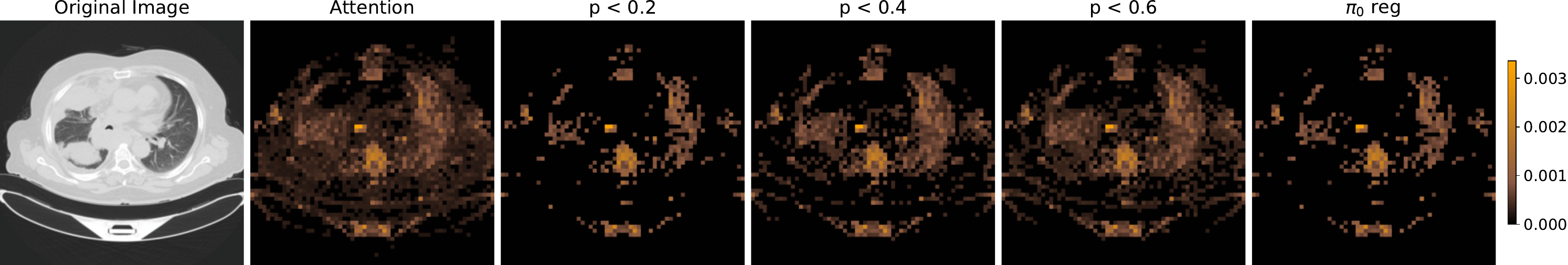}
\end{subfigure}
\caption{Results of the proposed attention regularization on the lung cancer medical images (malignant case 21). The top row shows the histograms of LFDRs, p-values, and z-scores. In the bottom row, regularized attention maps with various $p$-thresholds are shown alongside the CT scan and the original attention map. In the rightmost, a $\pi_0$ threshold is automatically estimated. }
\label{fig:example_med}
\end{figure*}

The dots correspond to the sensitivity-specificity values for each presented image that are regularized with the $\pi_0$ threshold estimated for $p$-values and $l$-values accordingly. The black solid curve shows the median of all the displayed curves. The blue dashed line corresponds to the regularization process that shrinks all the attention scores in the image equally, hence any successful regularization curve should fall above that line. We see that the $p$-thresholding regularization is a bit more conservative than the $l$-thresholding and the regularization efficiency changes less steeply with the change of the threshold. This pattern follows from the fact that $p$-values are distributed more uniformly than the LFDR values, as can be seen in the example of Fig.~\ref{fig:perturbed_image_attention_and_hist} and similar figures in Appendix~\ref{app:noise_examples} (we also demonstrate the uniformity of $p$-values and LFDR in Fig.~\ref{fig:uniformity} in Appendix~\ref{app:p_lfdr_uniformity}).

At last, we demonstrate the proposed regularization methods on the CT scans of lungs from IQ-OTH/NCCD \citep{alyasriy2020diagnosis}. Fig.~\ref{fig:example_med} shows the distribution of attention scores and their corresponding p-values and LFDRs from one malignant case of lung cancer. More examples and analyses are provided in Appendix~\ref{app:applications}.

\section{CONCLUSION}
\label{sec:conclusion}

We consider the attention score in transformer architectures as a noisy realization. Given all input features are assigned non-zero attention scores, we have developed a regularization method to suppresses noise while preserving salient features. We detailed how the bootstrap can be used in several different approaches that can be tailored to achieve the desired level of noise suppression. Due to the non-linearity and complexity of attention scores, the bootstrap is suitable to estimate the null distribution which can not be obtained otherwise.

We demonstrate the efficiency of these methods by simulating images. Our results highlight that both $p$-thresholding and $l$-thresholding regularization methods can achieve strong noise suppression across a wide range of thresholding hyper-parameters with $p$-regularization being slightly more conservative in terms of preserving the image features. The $\pi_0$-threshold estimation provides a remarkable balance between sensitivity and specificity without manual tuning. 

This study opens several directions for future work. Although we focused on particular realizations of parametric and non-parametric bootstrapping, further research is needed to investigate other forms of bootstrap sampling. Given a long history of the bootstrap \citep{efron1994bootstrap}, we expect to see even more accurate and efficient implementation. An interesting avenue for future studies would be to explore this regularization framework for different types of images, as different applications may require calibration to be useful in real world. 

By grounding attention refinement in statistical principles, this work advances interpretability and robustness in Transformer models and opens the door to more reliable deployment of attention-driven architectures across diverse domains.

\subsubsection*{Acknowledgements}
This work was funded by the SONATA BIS [2023/50/E/ST6/00694]
from Narodowe Centrum Nauki, with computational resources of Interdisciplinary Centre for Mathematical and Computational Modelling University of Warsaw [GDM-3540].






\bibliography{references}
\bibliographystyle{apalike}

\clearpage

\section*{Checklist}



\begin{enumerate}

  \item For all models and algorithms presented, check if you include:
  \begin{enumerate}
    \item A clear description of the mathematical setting, assumptions, algorithm, and/or model. [Yes]
    \item An analysis of the properties and complexity (time, space, sample size) of any algorithm. [Yes/No/Not Applicable]
    \item (Optional) Anonymized source code, with specification of all dependencies, including external libraries. [Not Applicable]
  \end{enumerate}

  \item For any theoretical claim, check if you include:
  \begin{enumerate}
    \item Statements of the full set of assumptions of all theoretical results. [Yes]
    \item Complete proofs of all theoretical results. [Yes]
    \item Clear explanations of any assumptions. [Yes]     
  \end{enumerate}

  \item For all figures and tables that present empirical results, check if you include:
  \begin{enumerate}
    \item The code, data, and instructions needed to reproduce the main experimental results (either in the supplemental material or as a URL). [Yes]
    \item All the training details (e.g., data splits, hyperparameters, how they were chosen). [Yes]
    \item A clear definition of the specific measure or statistics and error bars (e.g., with respect to the random seed after running experiments multiple times). [Yes]
    \item A description of the computing infrastructure used. (e.g., type of GPUs, internal cluster, or cloud provider). [Not Applicable]
  \end{enumerate}

  \item If you are using existing assets (e.g., code, data, models) or curating/releasing new assets, check if you include:
  \begin{enumerate}
    \item Citations of the creator If your work uses existing assets. [Yes]
    \item The license information of the assets, if applicable. [Not Applicable]
    \item New assets either in the supplemental material or as a URL, if applicable. [Yes]
    \item Information about consent from data providers/curators. [Not Applicable]
    \item Discussion of sensible content if applicable, e.g., personally identifiable information or offensive content. [Not Applicable]
  \end{enumerate}

  \item If you used crowdsourcing or conducted research with human subjects, check if you include:
  \begin{enumerate}
    \item The full text of instructions given to participants and screenshots. [Not Applicable]
    \item Descriptions of potential participant risks, with links to Institutional Review Board (IRB) approvals if applicable. [Not Applicable]
    \item The estimated hourly wage paid to participants and the total amount spent on participant compensation. [Not Applicable]
  \end{enumerate}

\end{enumerate}

\beginsupplement
\clearpage
\appendix
\thispagestyle{empty}

\onecolumn

\section{Examples of noise simulation and regularization}
\label{app:noise_examples}

\begin{figure*}[h]
\centering
\includegraphics[width=\linewidth]{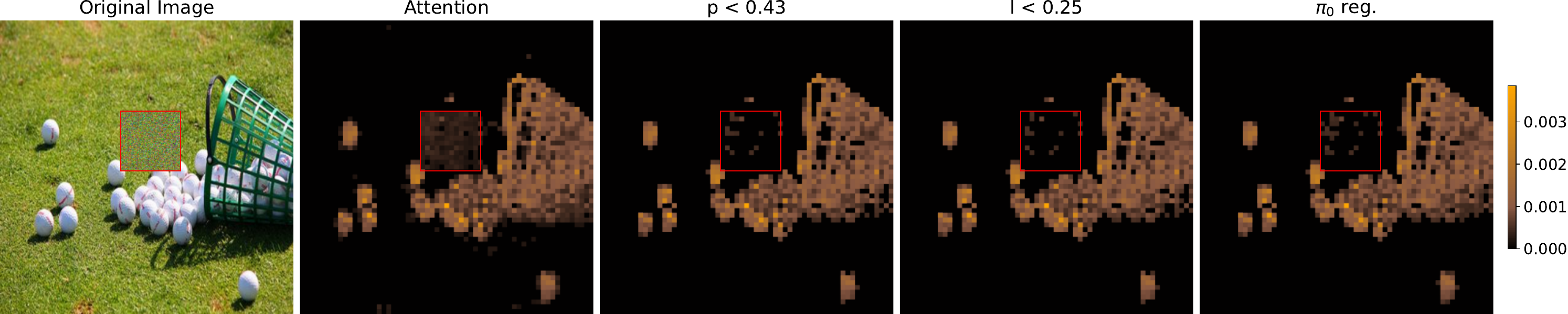}
\vfill
\includegraphics[width=\linewidth]{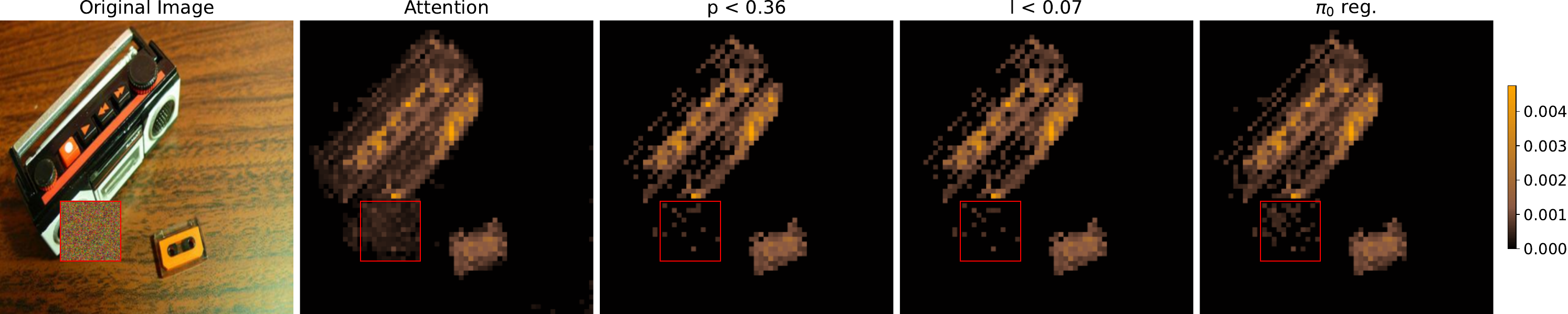}
\vfill
\includegraphics[width=\linewidth]{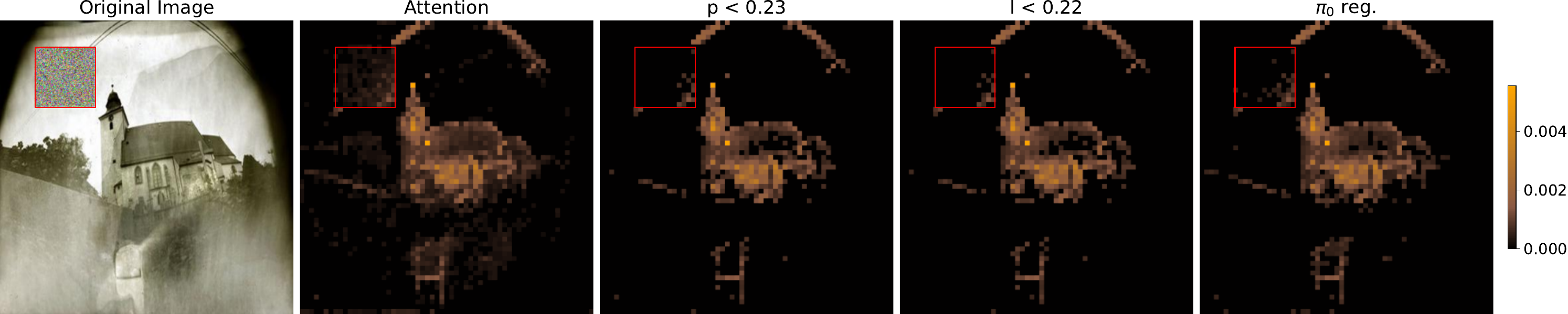}
\caption{Examples of perturbed images with the attention maps before and after regularization via different shrinkage methods: $p$-thresholding and $l$-thresholding with thresholds set at the 10-th percentile for $p$-values and LFDR respectively, and $\pi_0$-thresholding with $p$-values. The names of the images, the coordinates of the noise patch (bottom left corner of the patch) and the mean $z$ in ROI are as follows (from top to bottom): \texttt{n03445777\_3192.JPEG}, $(200, 150)$, $\langle z \rangle = 0.98$; \texttt{n02979186\_9450.JPEG}, $(100, 300)$, $\langle z \rangle = 1.03$; \texttt{n03028079\_8361.JPEG}, $(58, 44)$, $\langle z \rangle = 0.68$. The noise patch in the image and attention maps is highlighted with a red frame.}
\label{fig:perturbed_examples}
\end{figure*}

In Fig.~\ref{fig:perturbed_examples} we show the examples of regularization in our noise simulation study for different regularization methods and images from different categories. The figure layout and styling are exactly the same as in Fig.~\ref{fig:perturbed_image_attention_and_hist}. For $p$- and $l$-thresholds we again use the 10th quantile of the respective statistic. 

\begin{figure*}[h]
\centering
\begin{subfigure}{\linewidth}
    \includegraphics[width=\textwidth]{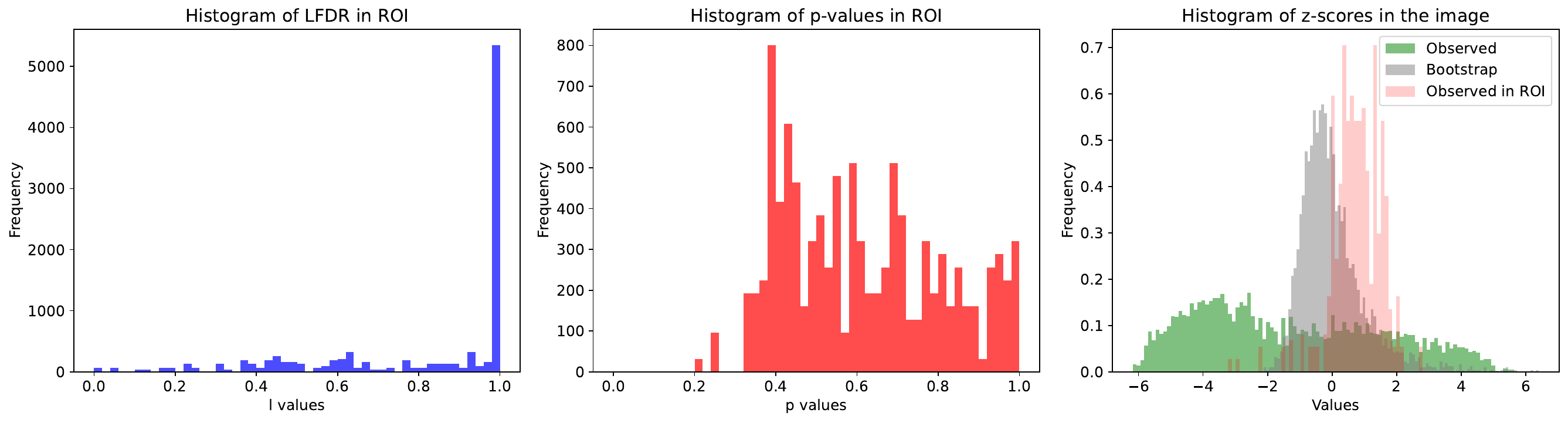}
\end{subfigure}
\vfill
\begin{subfigure}{\linewidth}
    \includegraphics[width=\textwidth]{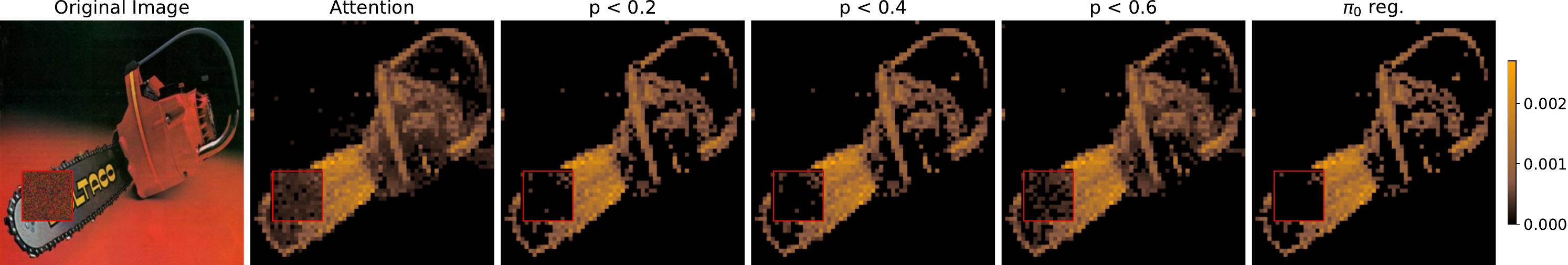}
\end{subfigure}
\vfill
\begin{subfigure}{\linewidth}
    \includegraphics[width=\textwidth]{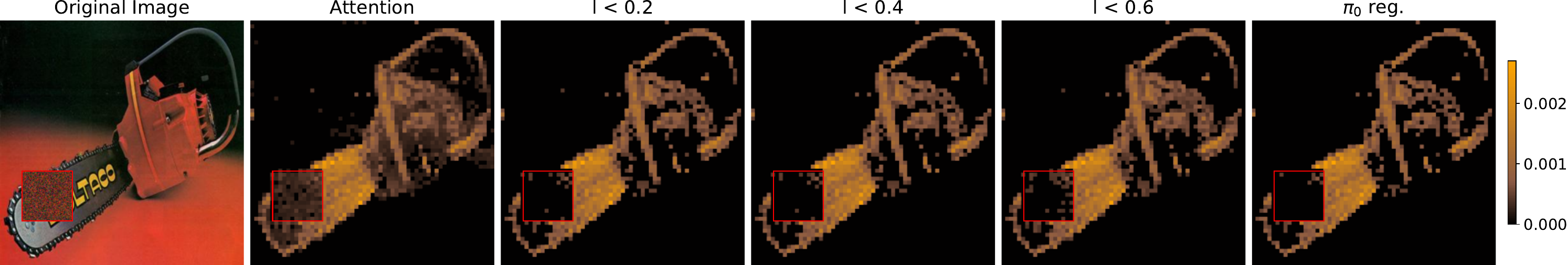}
\end{subfigure}
\caption{Example of a perturbed image (\texttt{n03000684\_5970.JPEG}) with the attention map before and after regularization via $p$-thresholding (middle row) and $l$-thresholding (bottom row) with the respective thresholds varying from $0.2$ to $0.6$. The last attention map in each of the rows to the right corresponds to $\pi_0$-thresholding. The top row shows the histograms of p-values in ROI, LFDR values in ROI and z-scores of the observed attention scores (green), bootstrap attention scores (gray) and the attention scores observed in ROI (red) corresponding to the image. 
The noise patch in the image and attention maps is highlighted with a red frame $(44, 300)$. The mean $z$-score in ROI is $\langle z \rangle = 0.54$.}
\label{fig:example_chainsaw}
\end{figure*}

\begin{figure*}[ht!]
\centering
\begin{subfigure}{\linewidth}
    \includegraphics[width=0.98\textwidth]{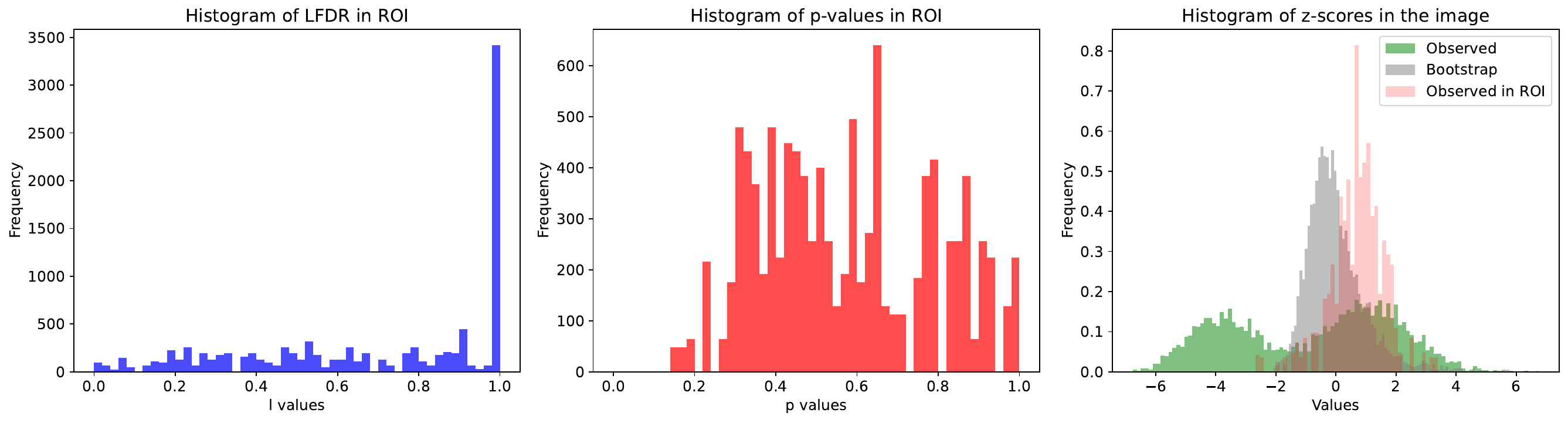}
\end{subfigure}
\vfill
\begin{subfigure}{\linewidth}
    \includegraphics[width=0.98\textwidth]{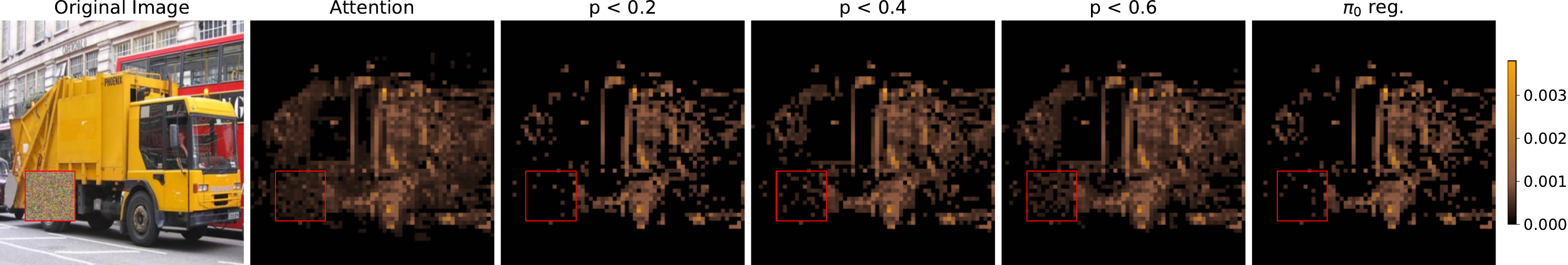}
\end{subfigure}
\vfill
\begin{subfigure}{\linewidth}
    \includegraphics[width=0.98\textwidth]{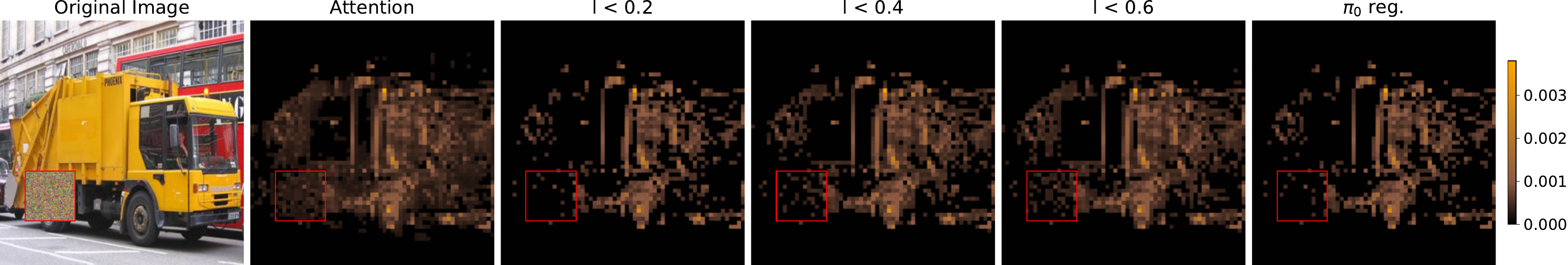}
\end{subfigure}
\caption{Example of a perturbed image (\texttt{n03417042\_3300.JPEG}) with the attention map before and after regularization via $p$-thresholding (middle row) and $l$-thresholding (bottom row) with the respective thresholds varying from $0.2$ to $0.6$. The last attention map in each of the rows to the right corresponds to $\pi_0$-thresholding. The top row shows the histograms of p-values in ROI, LFDR values in ROI and z-scores of the observed attention scores (green), bootstrap attention scores (gray) and the attention scores observed in ROI (red) corresponding to the image. The noise patch in the image and attention maps is highlighted with a red frame $(50, 300)$. The mean $z$-score in ROI is $\langle z \rangle = 0.88$.}
\label{fig:example_truck}
\end{figure*}

\begin{figure*}[h!]
\centering
\begin{subfigure}{\linewidth}
    \includegraphics[width=0.98\textwidth]{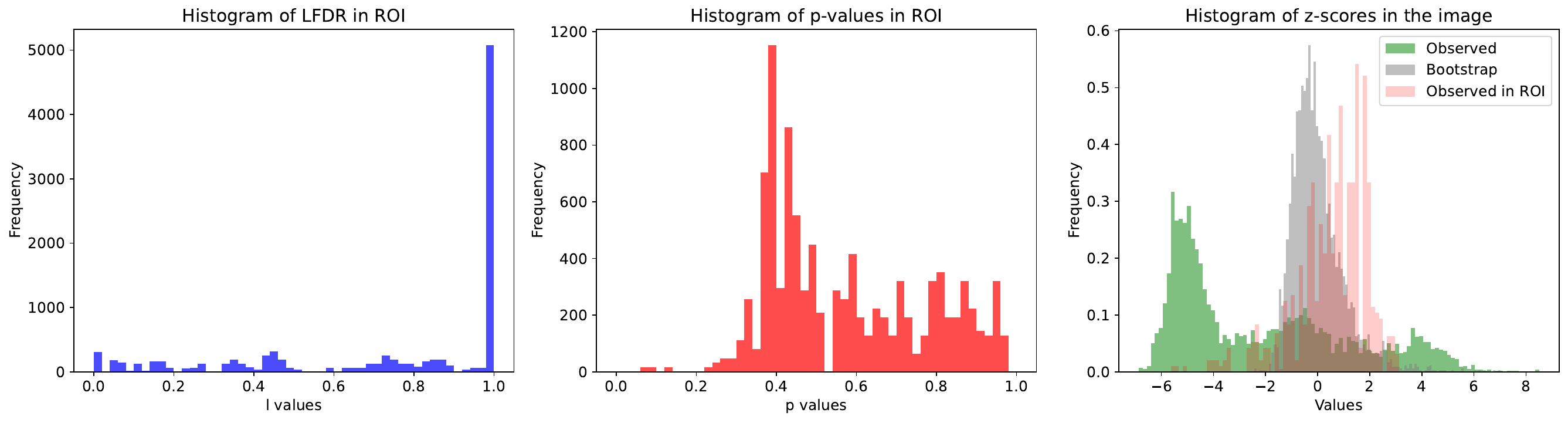}
\end{subfigure}
\vfill
\begin{subfigure}{\linewidth}
    \includegraphics[width=0.98\textwidth]{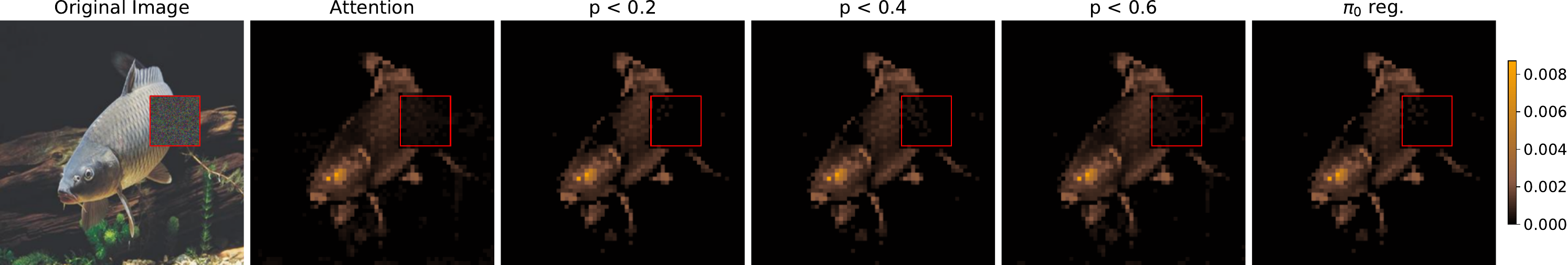}
\end{subfigure}
\vfill
\begin{subfigure}{\linewidth}
    \includegraphics[width=0.98\textwidth]{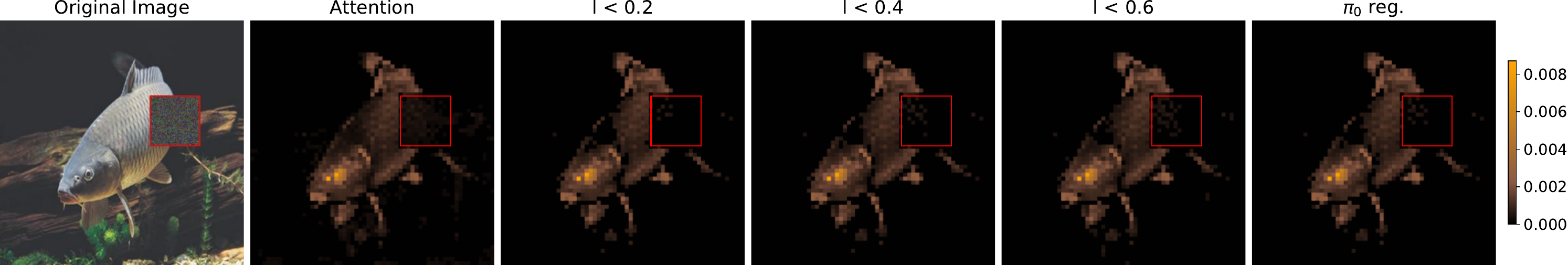}
\end{subfigure}
\caption{Example of a perturbed image (\texttt{n01440764\_1310}) with the attention map before and after regularization via $p$-thresholding (middle row) and $l$-thresholding (bottom row) with the respective thresholds varying from $0.2$ to $0.6$. The last attention map in each of the rows to the right corresponds to $\pi_0$-thresholding. The top row shows the histograms of p-values in ROI, LFDR values in ROI and z-scores of the observed attention scores (green), bootstrap attention scores (gray) and the attention scores observed in ROI (red) corresponding to the image. The noise patch in the image and attention maps is highlighted with a red frame $(300, 150)$. The mean $z$-score in ROI is $\langle z \rangle = 0.8$.}
\label{fig:example_fish}
\end{figure*}

\begin{figure*}[h!]
\centering
\begin{subfigure}{\linewidth}
    \includegraphics[width=\textwidth]{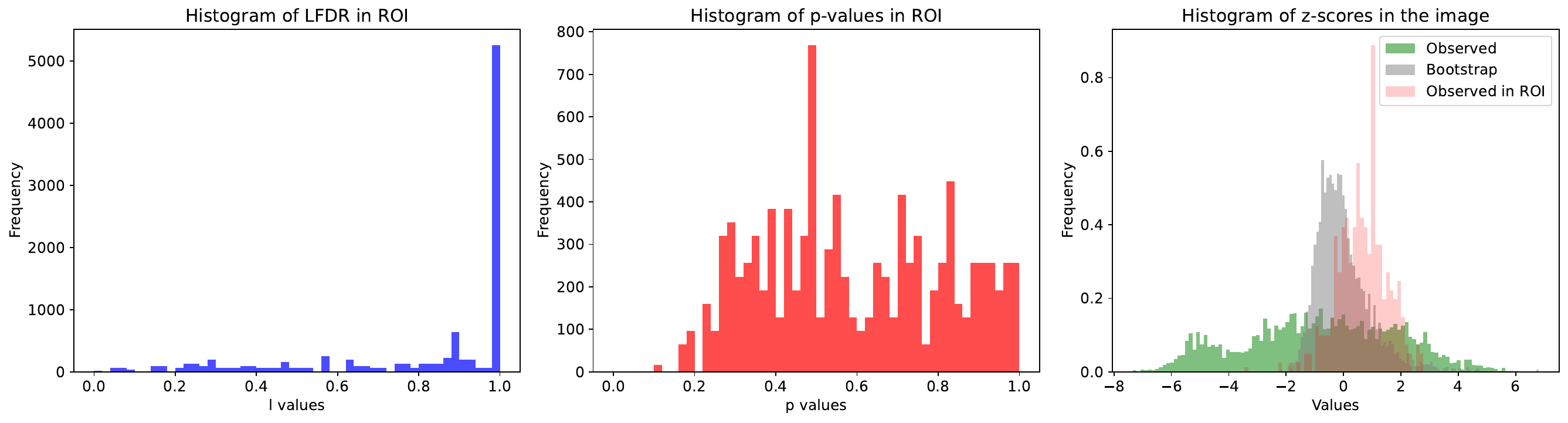}
\end{subfigure}
\vfill
\begin{subfigure}{\linewidth}
    \includegraphics[width=\textwidth]{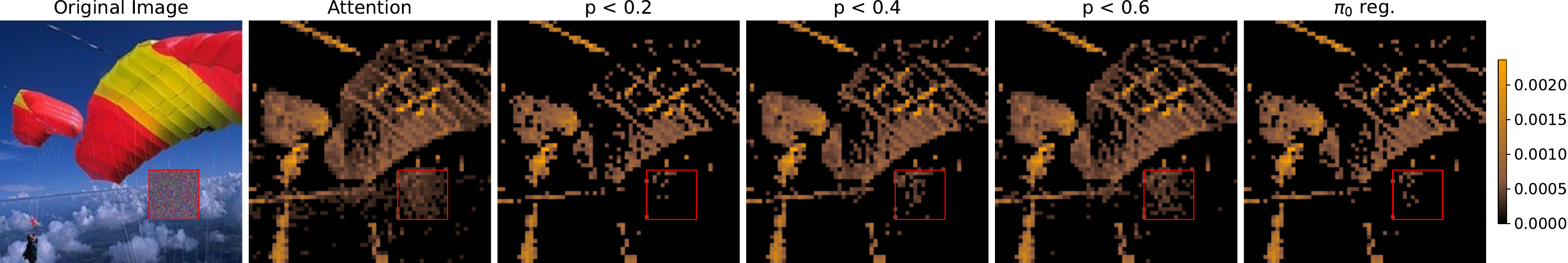}
\end{subfigure}
\vfill
\begin{subfigure}{\linewidth}
    \includegraphics[width=\textwidth]{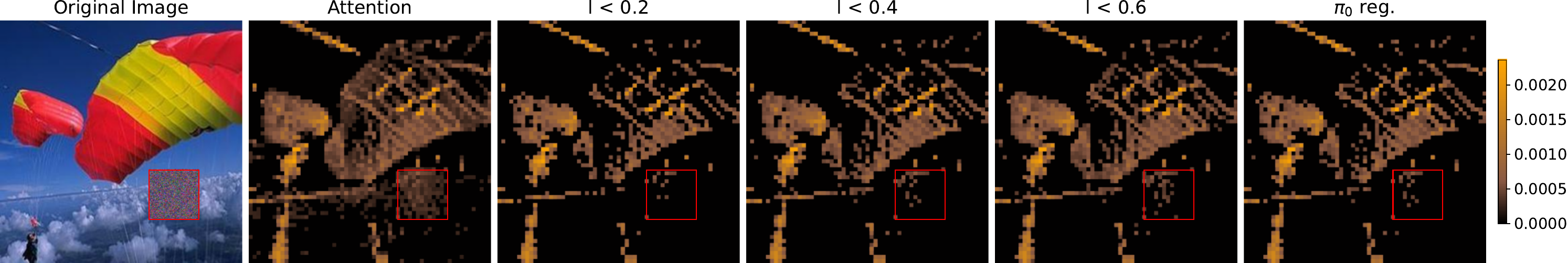}
\end{subfigure}
\caption{Example of a perturbed image (\texttt{n03888257\_6331.JPEG}) with the attention map before and after regularization via $p$-thresholding (middle row) and $l$-thresholding (bottom row) with the respective thresholds varying from $0.2$ to $0.6$. The last attention map in each of the rows to the right corresponds to $\pi_0$-thresholding. The top row shows the histograms of p-values in ROI, LFDR values in ROI and z-scores of the observed attention scores (green), bootstrap attention scores (gray) and the attention scores observed in ROI (red) corresponding to the image. The noise patch in the image and attention maps is highlighted with a red frame $(300, 300)$. The mean $z$-score in ROI is $\langle z \rangle = 0.71$.}
\label{fig:example_parasail}
\end{figure*}

\clearpage
In Figs.~\ref{fig:example_chainsaw}--\ref{fig:example_parasail}, we show the results of regularization for various images and for different threshold values for $p$- and $l$-regularization (middle row and bottom row respectively), as well as the corresponding histograms of $p$-values, LFDR and z-scores (top row), similarly to Fig.~\ref{fig:perturbed_image_attention_and_hist} (bottom). The last attention map on the right corresponds to $\pi_0$-thresholding. 

\clearpage
\section{Mean z-scores in ROI}
\label{app:mean_z_scores_roi}

\begin{figure*}[h]
    \centering
    \begin{subfigure}{0.49\textwidth}
    \centering
        \includegraphics[width=\textwidth]{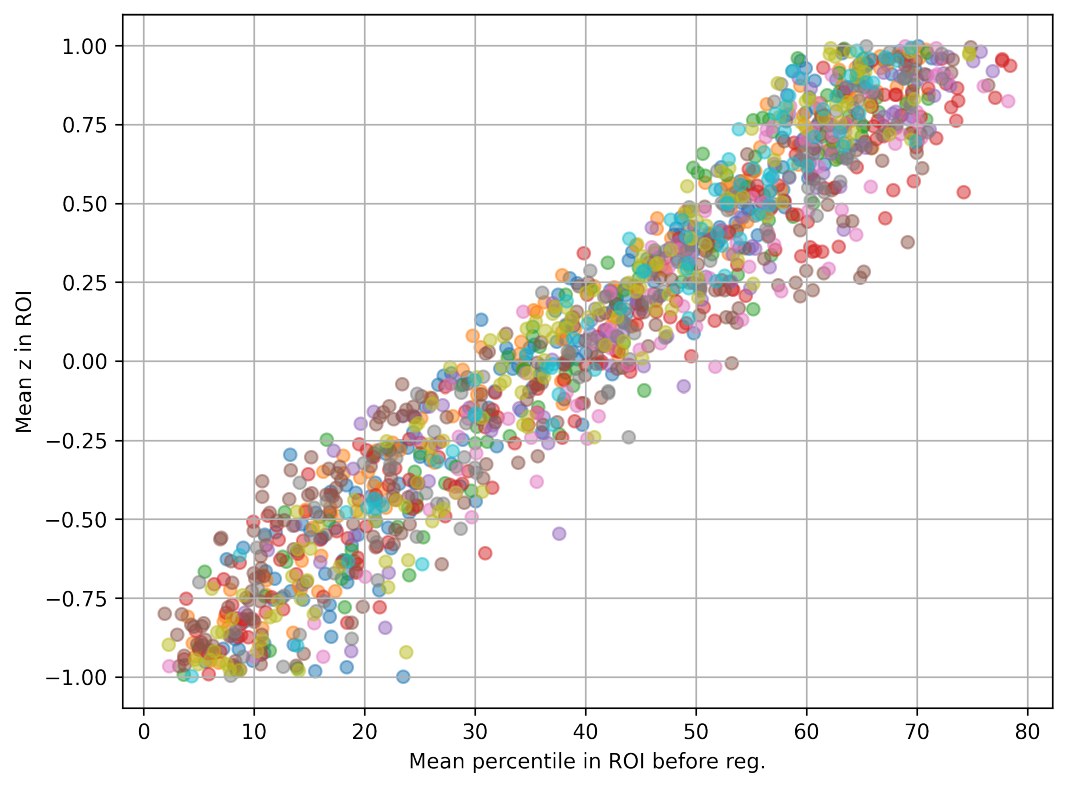}
        \caption{}
    \end{subfigure}
    \hfill
    \begin{subfigure}{0.47\textwidth}
        \includegraphics[width=\textwidth]{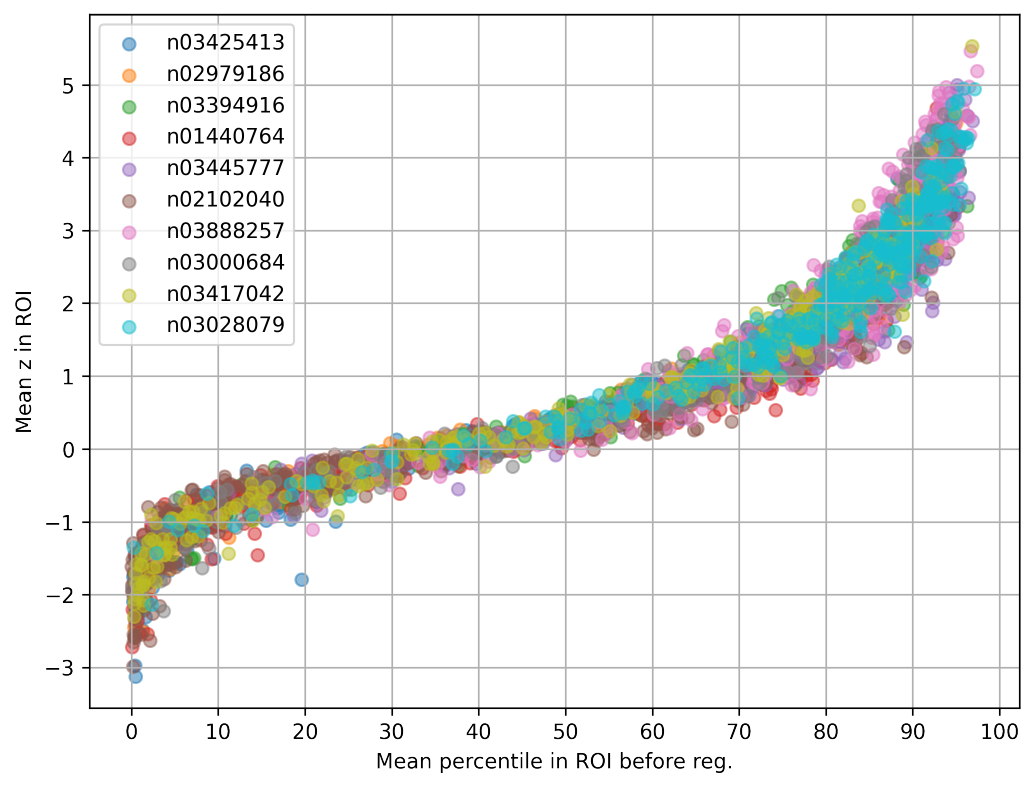}
        \caption{}
    \end{subfigure}
    \caption{Mean z-scores in ROI vs. mean percentile of attention scores in ROI w.r.t. the rest of the image before regularization for mean $|z| \leq 1$ (left) and for all of our noise simulation cases (right). Each point corresponds to one perturbed image with color denoting the category of the image from the Imagenette validation dataset.
    }
    \label{fig:z_offset}
\end{figure*}

In Fig.~\ref{fig:z_offset} we show the mean $z$-scores in ROI vs. mean percentile of attention scores in ROI w.r.t. the rest of the image before regularization for our noise simulation study. The left plot shows the distribution of $z$-scores for the cases that we actually include in our study described in Sec.~\ref{sec:noise_simulation}, where we select the perturbed images with $|z| \leq 1$, while the right plot shows the $z$-score distribution for all the perturbed images that we obtained. The results of regularization for all the images are presented in Appendix~\ref{app:non_filtered_results}.

\clearpage
\section{Uniformity of $p$-values and LFDR}
\label{app:p_lfdr_uniformity}

We use signed root mean square deviation (sRMSD) \citep{yao2023srmsd} to measure the uniformity of $p$-value distributions defined as follows:

\begin{equation}
\operatorname{sRMSD}(P) =
    \operatorname{Sign}\!\biggl[ \tfrac{1}{2} - \operatorname{med}(P) \biggr]
    \times \sqrt{\frac{1}{n} \sum_{i=1}^n \left( p_{(i)} - \tfrac{i - 0.5}{n} \right)^2 } \, ,
\end{equation}
where $P$ is the set of $p$-values, med($P$) denotes the median of the set and the sum goes over all the values in the set. The sign shows whether the distribution is tilted towards 0 (positive) or 1 (negative) w.r.t. the perfectly uniform case in which sRMSD is zero. The absolute upper limit for sRMSD is $\sqrt{1/3}$.

\begin{figure*}[h]
    \centering
    \begin{subfigure}{0.48\textwidth}
    \centering
        \includegraphics[width=\textwidth]{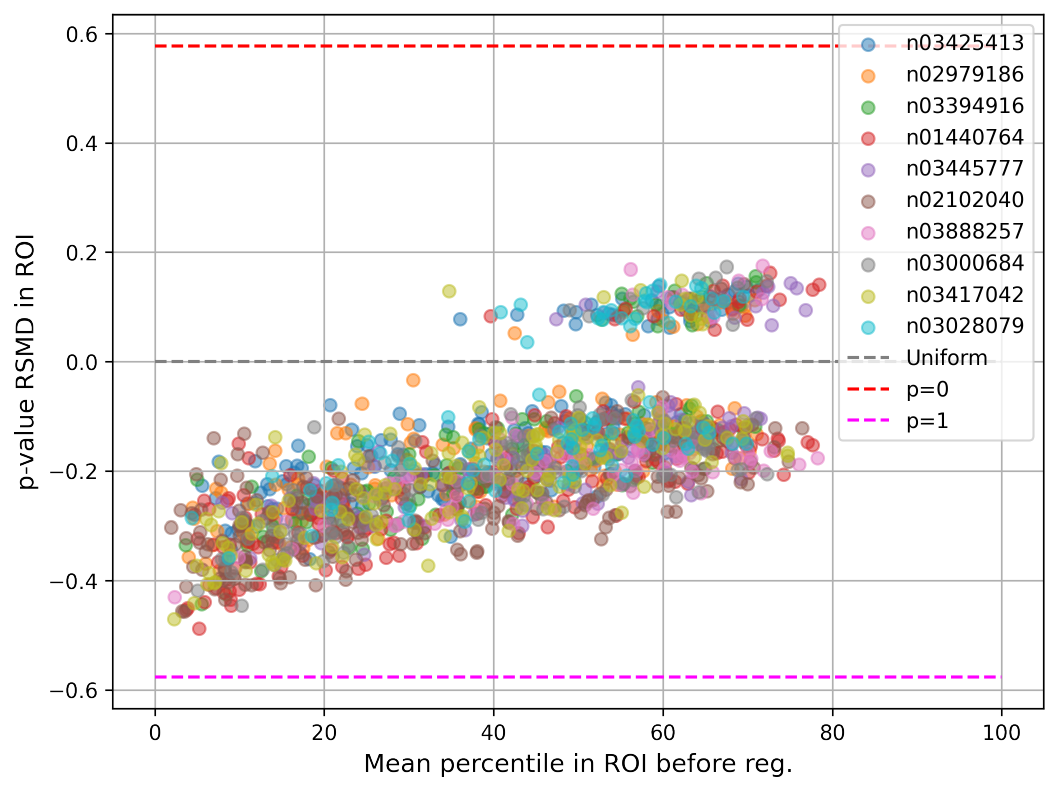}
        \caption{}
    \end{subfigure}
    \caption{sRSMDs in ROI for $p$-values vs. the mean percentile in ROI. Each point corresponds to one perturbed image and the color denotes the category of this image. Gray dashed line corresponds to a perfect uniformity of scores, while the red and magenta dashed lines correspond to the values being concentrated around 0 and 1 respectively. See text for more details.}
    \label{fig:uniformity}
\end{figure*}

In Fig.~\ref{fig:uniformity} we show the distributions of sRSMD values for $p$-values in ROI in our noise simulation study. Each dot corresponds to one perturbed image and the color denotes the category. The distribution of sRSMD values is plotted against the mean percentile in ROI before regularization to indicate how the shape of the distribution of these statistics changes with the level of noise. The distributions tend to be more skewed towards 0 as the mean percentile in ROI grows, which implies that the attention scores are further away from the bootstrap distribution. 

\section{Regularization efficiency for images not filtered by $|z| \leq 1$ }
\label{app:non_filtered_results}

\begin{figure*}[h]
    \centering
    \begin{subfigure}{0.49\textwidth}
    \centering
        \includegraphics[width=\textwidth]{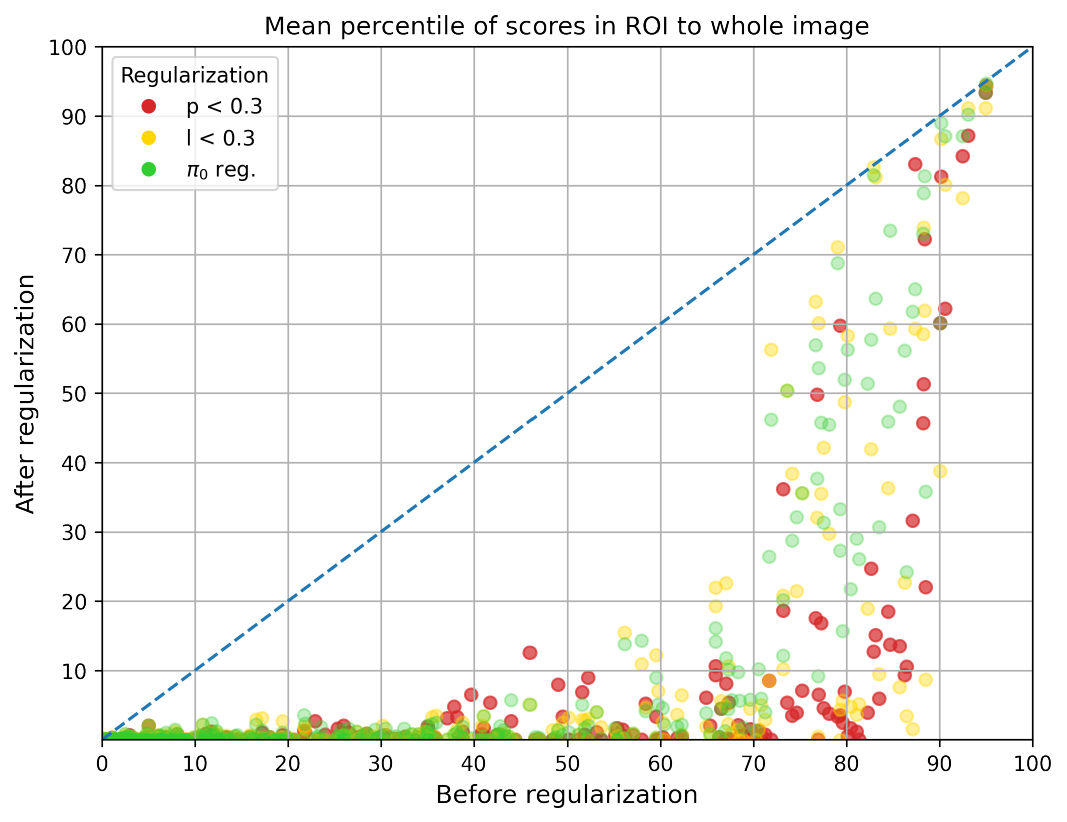}
        \caption{}
    \end{subfigure}
    \hfill
    \begin{subfigure}{0.49\textwidth}
        \includegraphics[width=\textwidth]{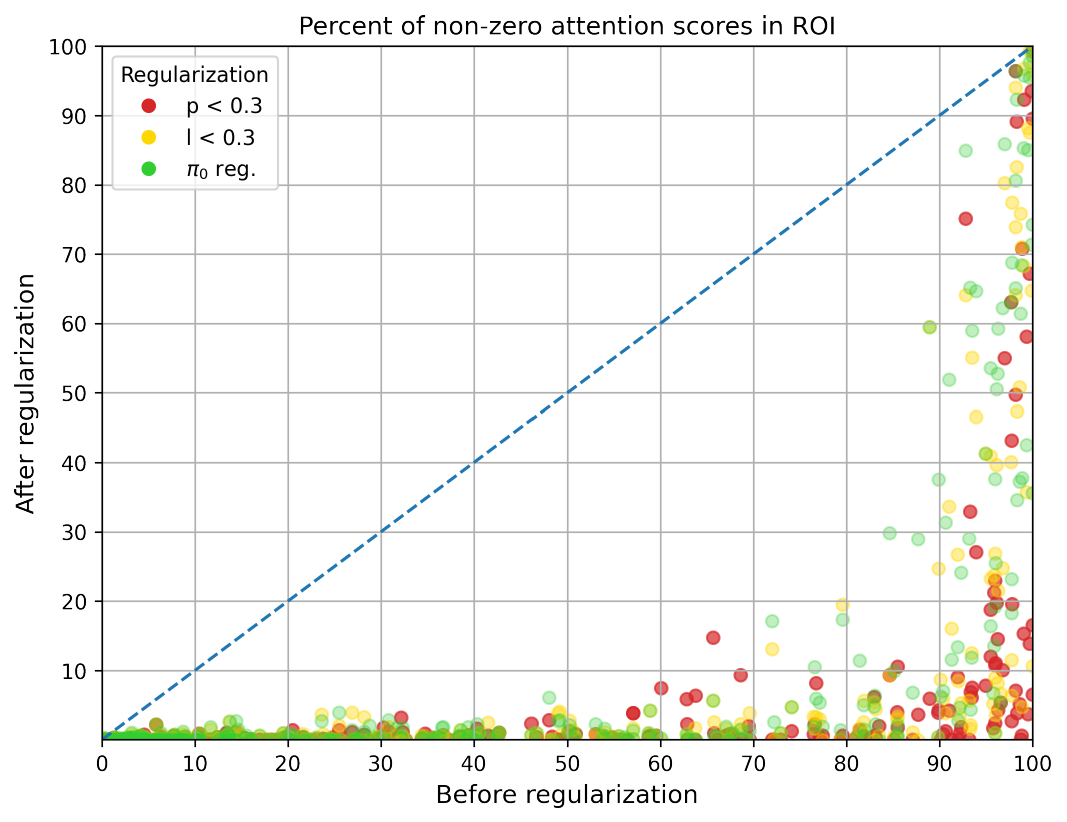}
        \caption{}
    \end{subfigure}
    \caption{Regularization efficiency in ROI for various images from a particular category of the Imagenette validation subset (n02102040) without $z$-filtering expressed in terms of: a) mean percentile of scores in ROI w.r.t. the whole image before and after regularization; b) percentage of non-zero attention scores in ROI before and after regularization. Each dot denotes the results of using one method for one perturbed image: $p$-thresholding (red), $l$-thresholding (yellow), $\pi_0$-thresholding (green). The threshold values for this example are $p_{\rm th} = 0.3$ and $l_{\rm th} = 0.3$. The blue dashed line corresponds to the same metric values before and after regularization.}
    \label{fig:reg_efficiency_n02102040}
\end{figure*}

\begin{figure}
    \centering
    \begin{subfigure}{0.49\textwidth}
    \centering
        \includegraphics[width=\textwidth]{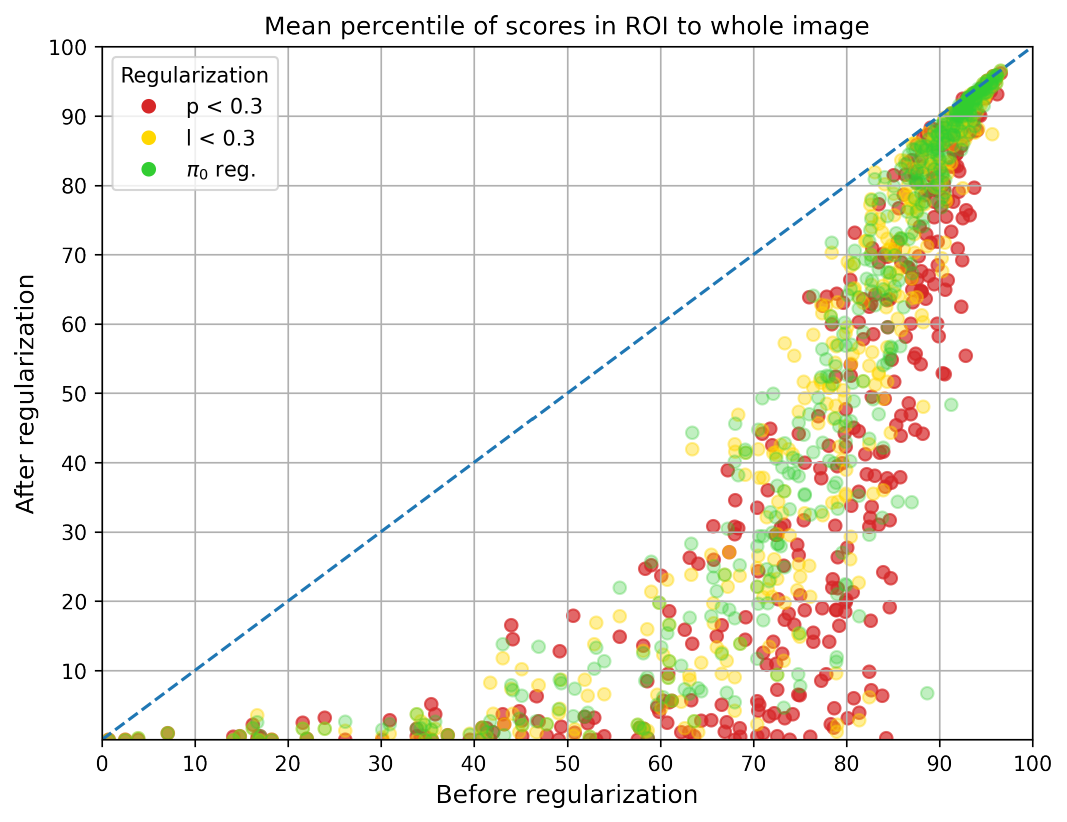}
        \caption{}
    \end{subfigure}
    \hfill
    \begin{subfigure}{0.49\textwidth}
        \includegraphics[width=\textwidth]{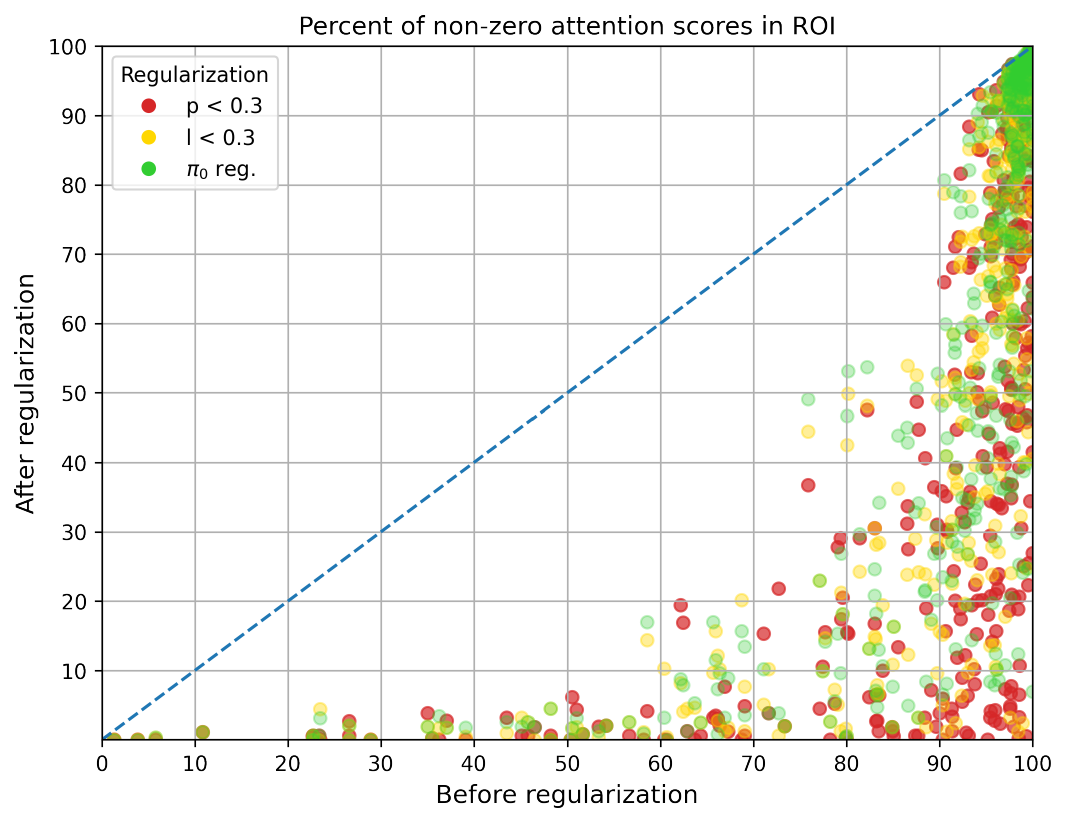}
        \caption{}
    \end{subfigure}
    \caption{Regularization efficiency in ROI for various images from a particular category of the Imagenette validation subset (n03028079) without $z$-filtering expressed in terms of: a) mean percentile of scores in ROI w.r.t. the whole image before and after regularization; b) percentage of non-zero attention scores in ROI before and after regularization. Each dot denotes the results of using one method for one perturbed image: $p$-thresholding (red), $l$-thresholding (yellow), $\pi_0$-thresholding (green). The threshold values for this example are $p_{\rm th} = 0.3$ and $l_{\rm th} = 0.3$. The blue dashed line corresponds to the same metric values before and after regularization.}
    \label{fig:reg_efficiency_n03028079}
\end{figure}

\begin{figure}
    \centering
    \includegraphics[width=0.8\textwidth]{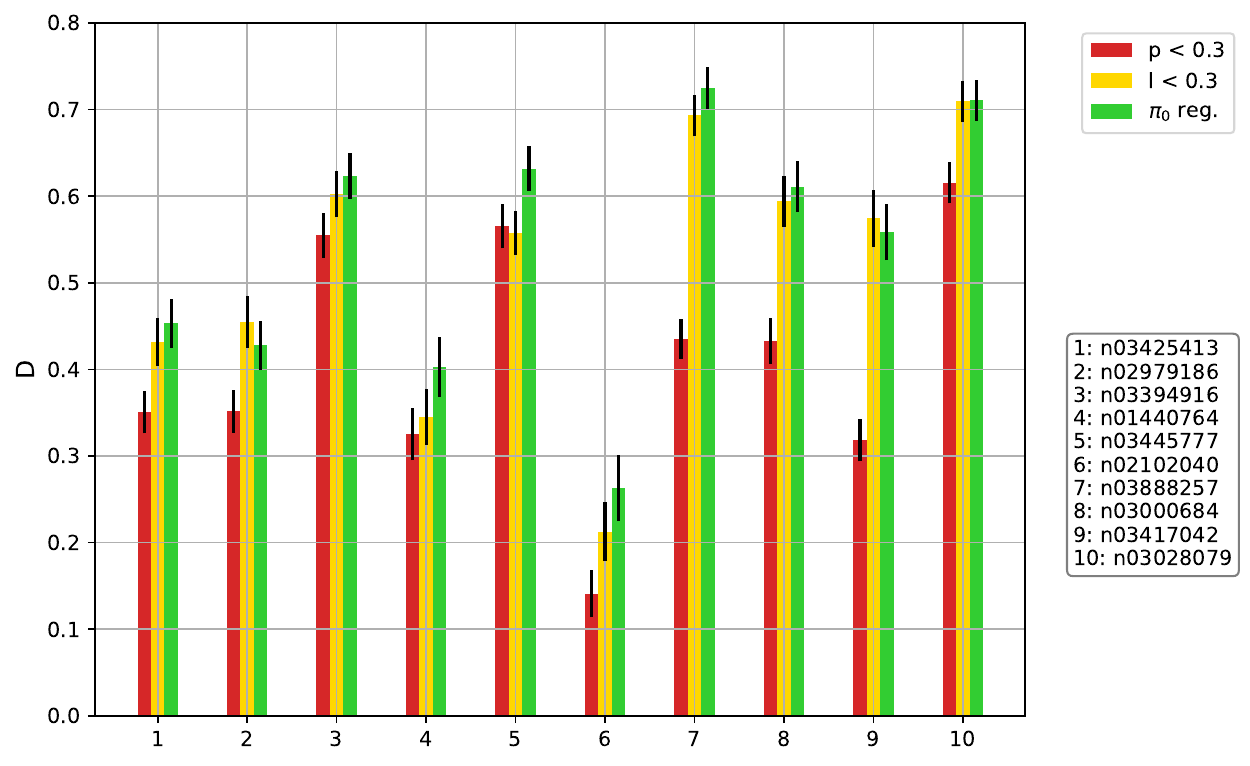}
    \caption{The average suppression factor $D$ for different categories of the Imagenette validation subset without $z$-filtering and different shrinkage methods. The lower is $D$ the better average regularization is achieved with $D=1$ being the case when all the points in Fig.~\ref{fig:mean_perc_roi_reg} located on the dashed blue curve. The relative uncertainty of the results are shown with the black lines.
    }
    \label{fig:D_histogram_non_filtered}
\end{figure}

In this appendix we present the results similar to the ones described in Section~\ref{sec:results} (Figs.~\ref{fig:reg_efficiency} and~\ref{fig:D_histogram}), but for all the perturbed images without $z$-filtering (Fig.~\ref{fig:z_offset}, right). Without $z$-filtering several images contain noise samples that are interpreted as positive signal by ViT and receive high attention scores (which result in large mean $z$ values in ROI). Here we present the mean percentile of scores in ROI and the percentage of non-zero attention scores in ROI before and after regularization separately for different categories: n02102040 (dogs) in Fig.~\ref{fig:reg_efficiency_n02102040} and n03028079 (churches) in Fig.~\ref{fig:reg_efficiency_n03028079}. All the figure layout and styling as well as the details of the study including the values of hyperparameters are the same as for results presented in Fig.~\ref{fig:reg_efficiency}. We show these two particular categories of images as they manifest different distributions of average attention scores in ROI before regularization. The results for the suppression factor $D$ for different categories, similar to Fig.~\ref{fig:D_histogram}, are shown in Fig.~\ref{fig:D_histogram_non_filtered}. Even though we do not separate the images with high mean $z$-scores in ROI, the regularization efficiency on average is still quite good.

\newcommand{\scaleone}{1.0}

\section{The impact of bootstrap hyperparameters}
\label{app:hyperparameters}

We examine how the choice of the bootstrap hyperparameters affects the regularization. We perform the same study as in Sec.~\ref{sec:noise_simulation} with the same parametric bootstrap method (normal distribution) and vary independently the number of bootstrap samples $B$ and the width of the bootstrap distribution, which we parameterize as $\sigma^j_B = {\rm \omega \times \sigma^j}$, where the index $j$ denotes RGB channels, $\sigma^j$ is the standard deviation of the image pixels in each channel and $\omega$ is the width parameter that we actually alter. We use the average suppression factor $D$ from Eq.~\eqref{eq:D_factor} as the main metric to evaluate the impact of different hyperparameter values on regularization. 

\begin{figure*}[h]
    \centering
    \begin{subfigure}{0.49\textwidth}
    \centering
        \includegraphics[width=\textwidth]{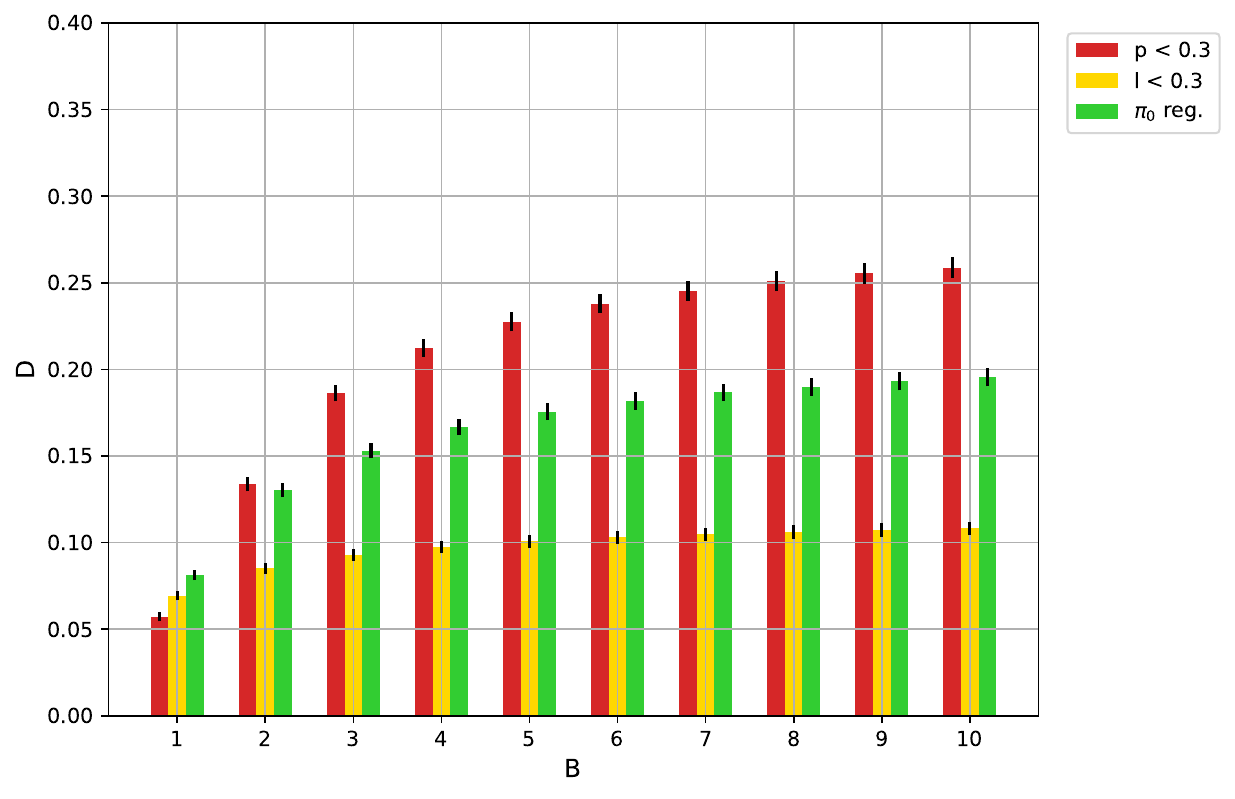}
        \caption{}
    \end{subfigure}
    \hfill
    \begin{subfigure}{0.49\textwidth}
        \includegraphics[width=\textwidth]{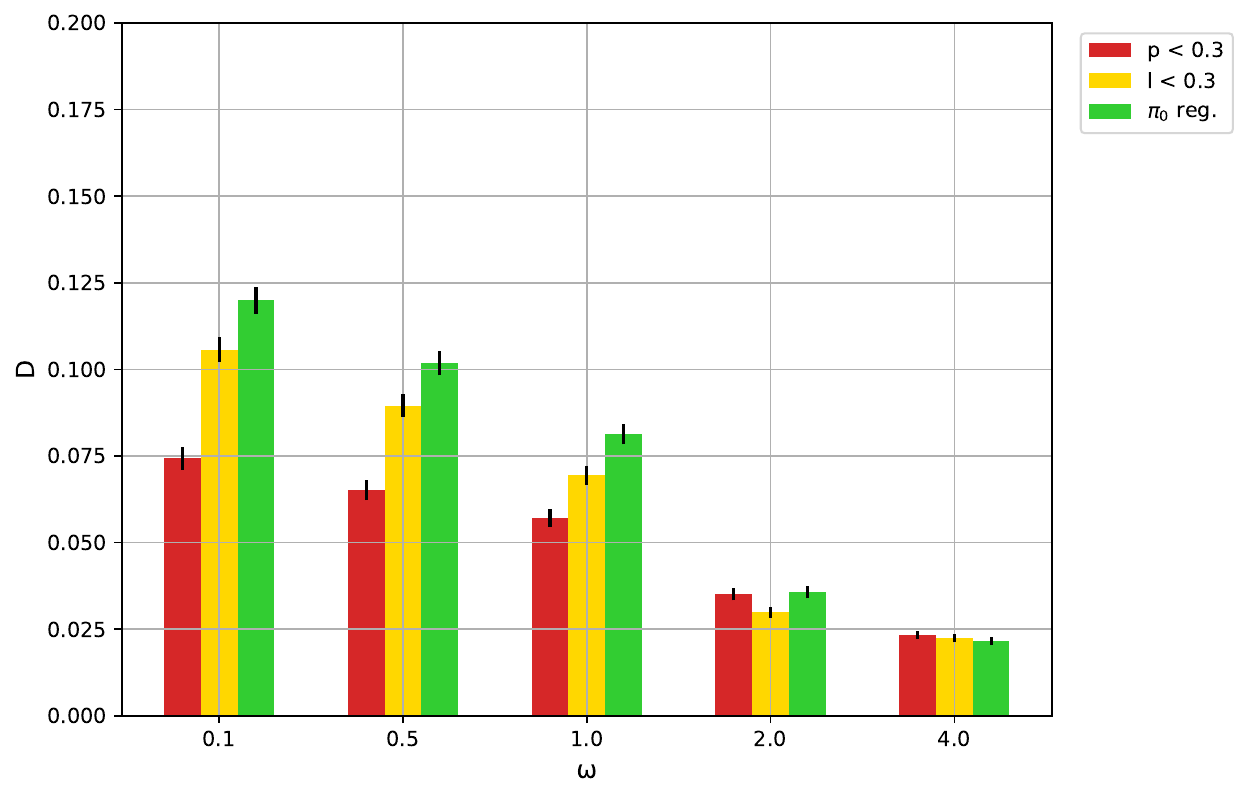}
        \caption{}
    \end{subfigure}
    \caption{The average suppression factor $D$ for different values of the $B$ hyperparameter (\textit{left}) and for different values of the $\omega$ hyperparameter (\textit{right}) for all the images from the Imagenette validation subset (with the mean $|z|$ in ROI $\leq 1$) and different shrinkage methods. The relative uncertainty of the results are shown with the black lines. See text for more details.
    }
    \label{fig:D_histogram_var_B_std}
\end{figure*}

\begin{figure*}[]
\centering
\includegraphics[width=\linewidth]{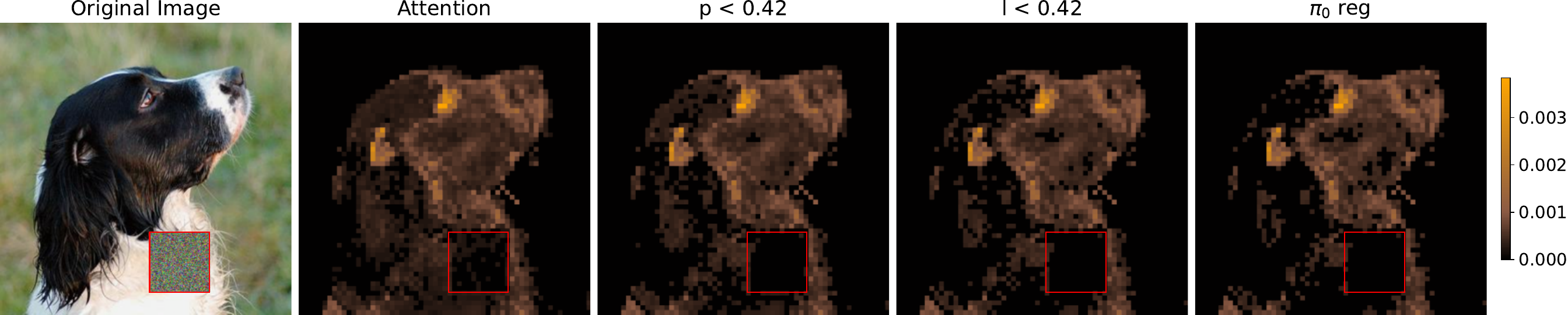}\\[0.5cm] 
\includegraphics[width=\linewidth]{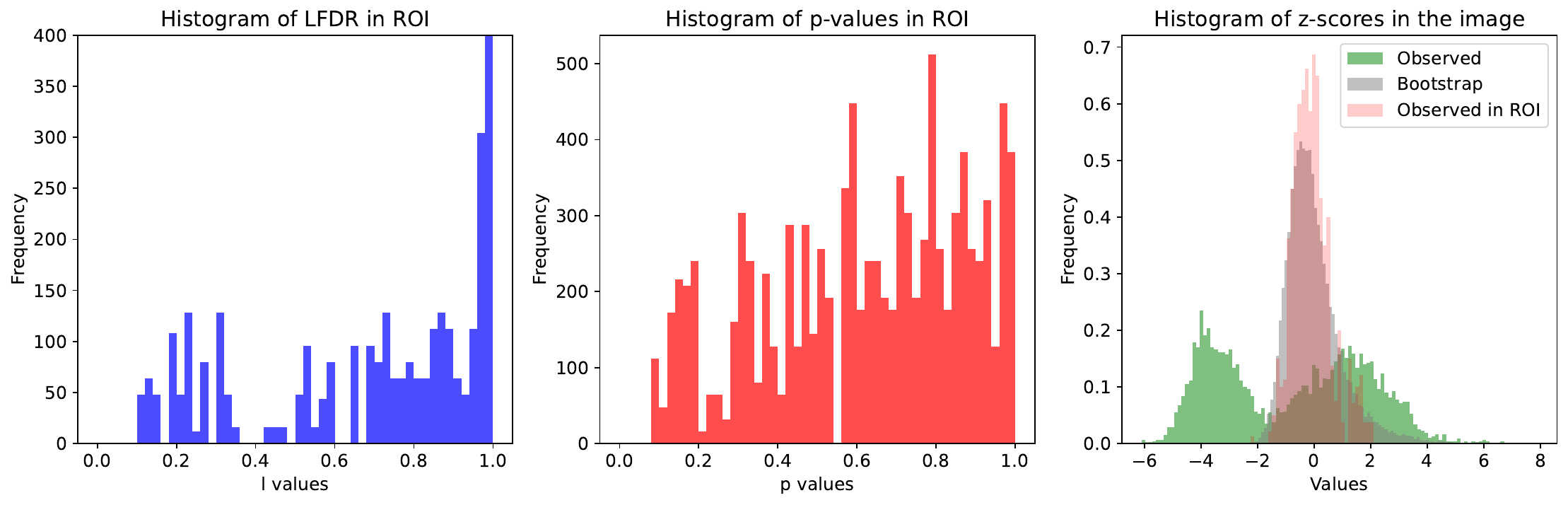}
\caption{Attention map before and after regularization via different shrinkage methods ($p$-thresholding, $l$-thresholding and $\pi_0$-thresholding) for the same image as in Fig.~\ref{fig:perturbed_image_attention_and_hist} (\texttt{n02102040\_821.JPEG}), but for $B = 10$ (top row); Histograms of p-values in ROI, LFDR values in ROI and z-scores of the observed attention scores (green), bootstrap attention scores (gray) and the attention scores observed in ROI (red) corresponding to the image (bottom row). 
The noise patch in the image and attention maps is highlighted with a red frame.}
\label{fig:perturbed_image_attention_and_hist_B=10}
\end{figure*}

In Fig.~\ref{fig:D_histogram_var_B_std} we present the results of this study for different values of $B$ (left plot) and $\omega$ (right plot). We show the D factors calculated on the whole Imagenette validation dataset (1286 images after filtering for mean $|z| \leq 1$ in ROI, see Sec.~\ref{sec:noise_simulation}) with 3 different regularization methods (similarly to Fig.~\ref{fig:D_histogram}) and for the same threshold values as in our main study. Recall that the lower is the average $D$ factor the better is the noise reduction in our experiments. The left plot in Fig.~\ref{fig:D_histogram_var_B_std} illustrates that the increase of $B$ leads to a weaker noise reduction for all the methods, which is especially pronounced for the methods based on $p$-values ($p$-thresholding and $\pi_0$-thresholding), as the increase of the number of bootstrap $z$-scores below a certain attention $z$-score shifts its $p$-value closer to 0 and enlarge the population of $p$-values below the threshold. The LFDR values are rather dependent on the shape of the null distribution, hence the bootstrap sample enlargement leaves $l$-values more intact. In Fig.~\ref{fig:perturbed_image_attention_and_hist_B=10} we show the histograms of $l$-values, $p$-values (in ROI) and $z$ scores, as well as the attention maps before and after regularization, for the same perturbed image and threshold values as in Fig.~\ref{fig:perturbed_image_attention_and_hist}, but for $B = 10$. Note that sensitivity is traded off against specificity with the increase of $B$.

\begin{figure*}[]
\centering
\includegraphics[width=\linewidth]{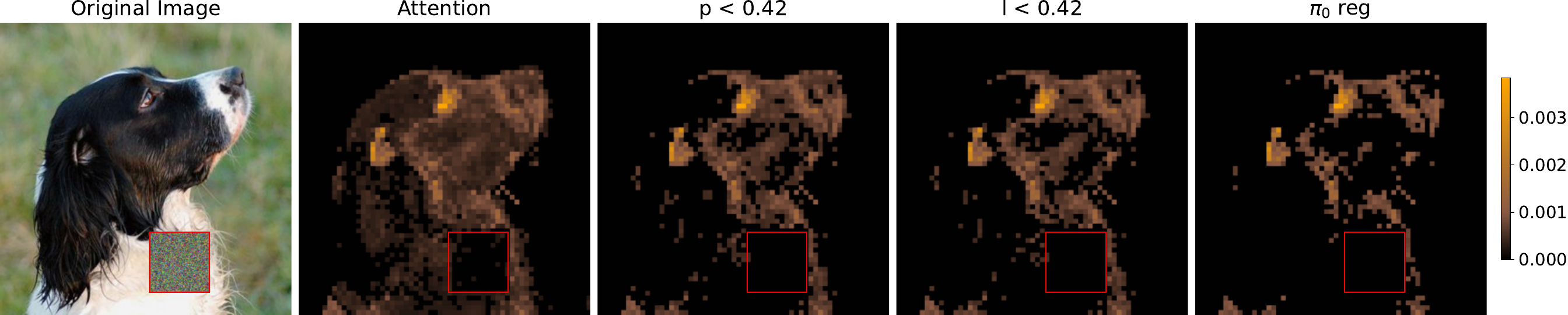}\\[0.5cm] 
\includegraphics[width=\linewidth]{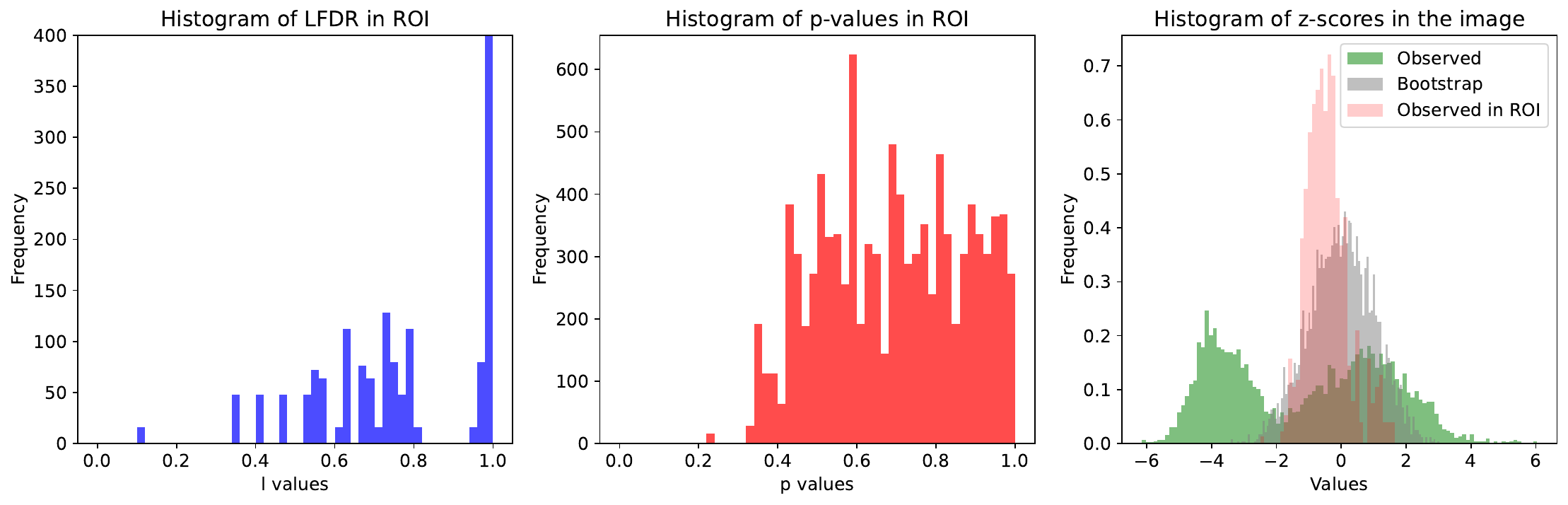}
\caption{Attention map before and after regularization via different shrinkage methods ($p$-thresholding, $l$-thresholding and $\pi_0$-thresholding) for the same image as in Fig.~\ref{fig:perturbed_image_attention_and_hist} (\texttt{n02102040\_821.JPEG}), but for standard deviation of $= 4$ (top row); Histograms of p-values in ROI, LFDR values in ROI and z-scores of the observed attention scores (green), bootstrap attention scores (gray) and the attention scores observed in ROI (red) corresponding to the image (bottom row). 
The noise patch in the image and attention maps is highlighted with a red frame.}
\label{fig:perturbed_image_attention_and_hist_std=4}
\end{figure*}

The right plot in Fig.~\ref{fig:D_histogram_var_B_std} shows that the wider is the bootstrap distribution the more attention scores in ROI are attributed to noise, hence it leads to the decrease of the average $D$ factor. In Fig.~\ref{fig:perturbed_image_attention_and_hist_std=4} we show the attention maps and statistics histogram as in Fig.~\ref{fig:perturbed_image_attention_and_hist_B=10}, but for $B=1$ and $\omega = 4$. Comparing the attention maps between the two cases (as well as the case in Fig.~\ref{fig:perturbed_image_attention_and_hist}) we can again notice the trade off between sensitivity and specificity. These considerations can be used to adjust the hyperparameters of the bootstrap distribution to reach a desired level of regularization. 
 
\clearpage
\section{Evaluation with non-parametric bootstrap}
\label{app:non_parametric_bootstrap}

\begin{figure*}[h]
    \centering
    \begin{subfigure}{0.329\textwidth}
    \centering
        \includegraphics[width=\textwidth]{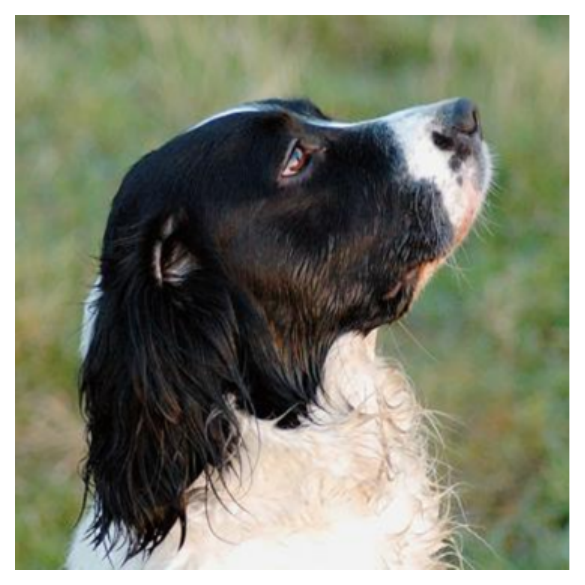}
        \caption{}
    \end{subfigure}
    \hfill
    \begin{subfigure}{0.329\textwidth}
    \centering
        \includegraphics[width=\textwidth]{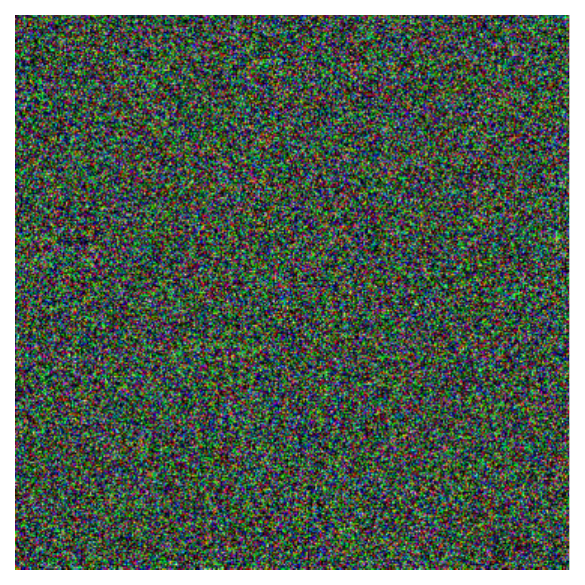}
        \caption{}
    \end{subfigure}
    \hfill
    \begin{subfigure}{0.329\textwidth}
        \centering
        \includegraphics[width=\textwidth]{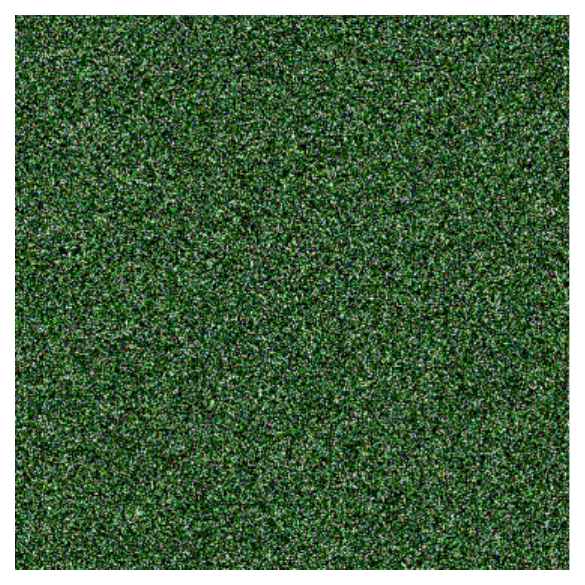}
        \caption{}
    \end{subfigure}
    \caption{Examples of bootstrap images generated from the original image (\texttt{n02102040\_821.JPEG}, subfigure \textit{a}) via parametric (gaussian) bootstrap (\textit{b}) and non-parametric bootstrap (\textit{c}).} 
    \label{fig:bootstrap_examples}
\end{figure*}

We conduct a study similar to the one described in Sec.~\ref{sec:noise_simulation}, but with a non-parametric bootstrapping. We prepare the null image by sampling with repetitions from the set of pixels of the original image. In Fig.~\ref{fig:bootstrap_examples} we show the examples of the bootstrap images sampled from the original image (left plot) via normal bootstrapping that we use in all our studies (center plot) and via non-parametric bootstrapping (right plot). 

We calculate the average $D$ factor for 1286 images from the Imagenette validation dataset (with mean $|z| \leq 1$ in ROI) for parametric and non-parametric bootstrapping. For both bootstrap methods we obtain the same values of $D = \left[ 0.057; 0.069 ; 0.081 \right] \pm 0.003$ for three shrinkage methods ($p$-thresholding, $l$-thresholding and $\pi_0$-thresholding) within the margin of error.

\clearpage
\section{Diffuse noise simulation}
\label{app:diffuse_noise}

In Sec.~\ref{sec:noise_simulation} we describe the noise simulation procedure, in which we inject a square of the gaussian noise into a random location in an image. Here we perform a different type of noise simulation, in which the noise patches are spread over the image in a more random and diffuse manner. 

For that we generate a $\mathcal{N}(0,1)$ field of the size of the image, smooth it using an FFT-based gaussian filter $S \propto \exp\left( -(x_f + y_f)^2*\lambda^2 \right)$, where $||x_f + y_f||$ is the distance from the center in the frequency space and $\lambda$ is the clustering parameter. We further apply min-max normalization, select the top $N_p$ pixels and create an ROI mask out of them. For the sake of comparison with our main noise simulation method we choose $N_p = 100 \times 100$. We take $\lambda = 20$, which produces multiple small, but visually noticeable clusters of noise. These clusters are filled with the same gaussian noise generated with the mean and standard deviation of the distribution of pixels in each RGB channel as we used in all of our experiments. 

\begin{figure*}[htb]
\centering
\includegraphics[width=\linewidth]{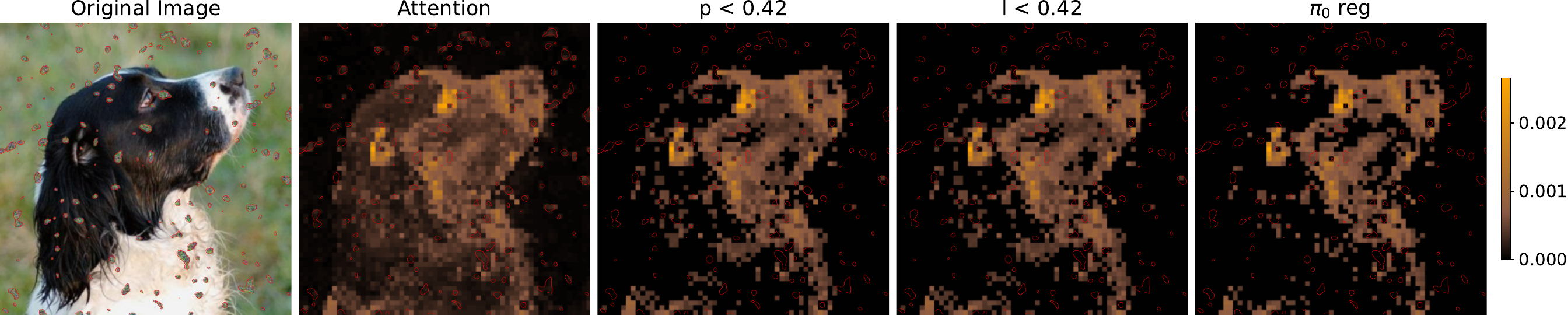}\\[0.5cm] 
\includegraphics[width=\linewidth]{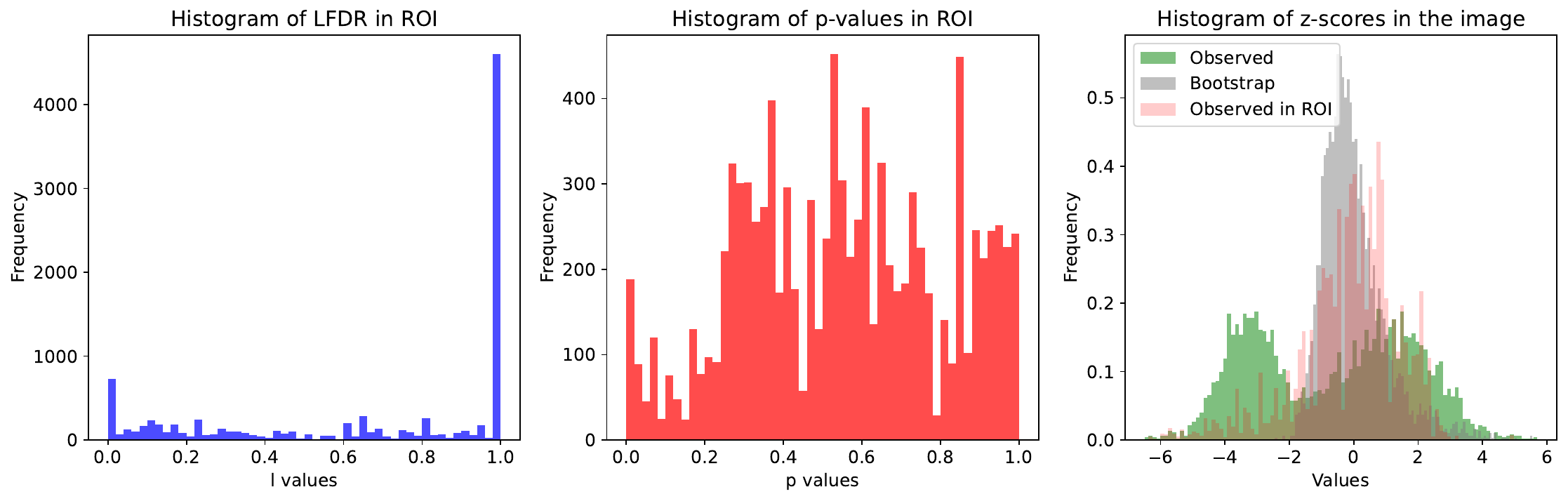}
\caption{Attention map before and after regularization via different shrinkage methods ($p$-thresholding, $l$-thresholding and $\pi_0$-thresholding) for an image (\texttt{n02102040\_821.JPEG}) perturbed with diffuse gaussian noise, (top row); Histograms of p-values in ROI, LFDR values in ROI and z-scores of the observed attention scores (green), bootstrap attention scores (gray) and the attention scores observed in ROI (red) corresponding to the image (bottom row). 
The noise patches in the image and attention maps are highlighted with a red frame. See text for more details.}
\label{fig:perturbed_image_attention_and_hist_diffuse}
\end{figure*}

In Fig.~\ref{fig:perturbed_image_attention_and_hist_diffuse} we show an example of an image (\texttt{n02102040 821.JPEG}, which we commonly adopt as a benchmark image in our studies) perturbed with the injection of such diffuse noise, as well as the corresponding attention maps before and after regularization and the histograms of $l$-,$p$- and $z$-values (similarly to Fig.~\ref{fig:perturbed_image_attention_and_hist}). 

\begin{figure*}[h]
    \centering
    \includegraphics[width=0.8\textwidth]{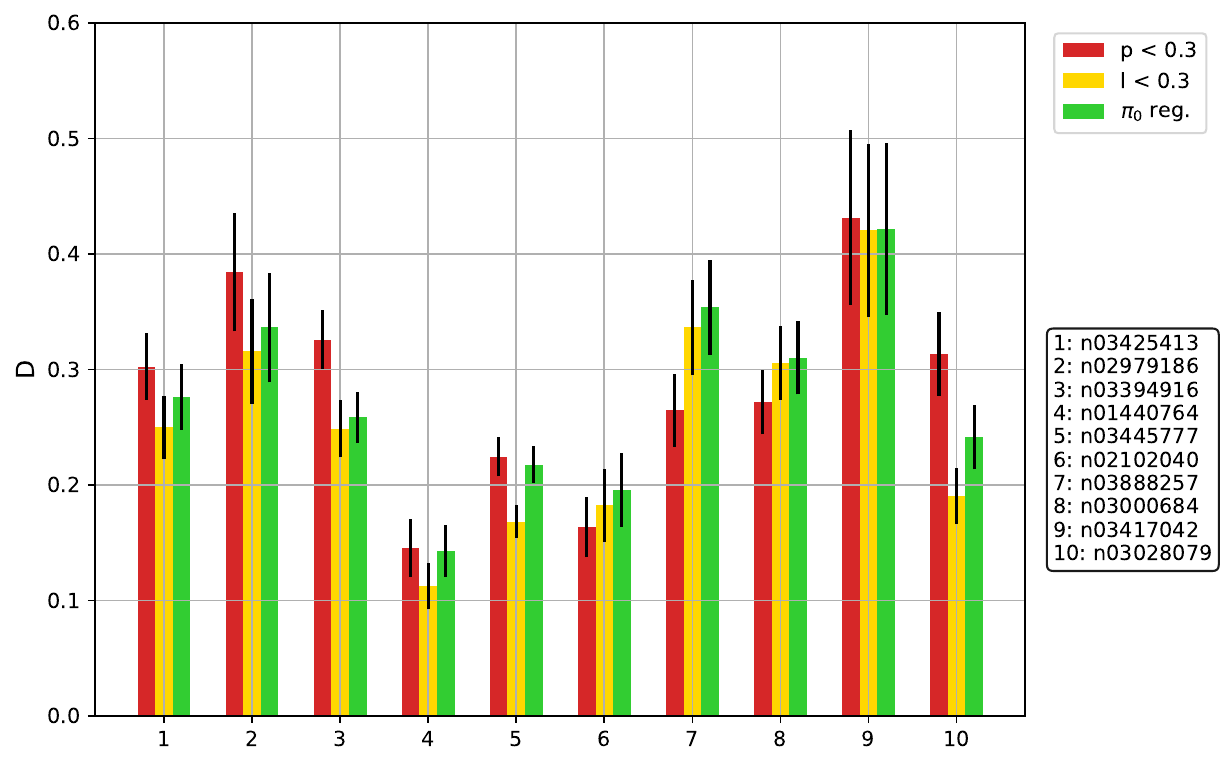}
    \caption{The average suppression factor $D$ for different categories of images from the Imagenette validation subset perturbed with diffuse gaussian noise (with the mean $|z|$ in ROI $\leq 1$) and different shrinkage methods. The relative uncertainty of the results are shown with the black lines. See text for more details.
    }
    \label{fig:D_histogram_diffuse}
\end{figure*}

In Fig.~\ref{fig:D_histogram_diffuse} we show the average suppression factors $D$ calculated for each category of the Imagenette validation dataset (similarly to Fig.~\ref{fig:D_histogram}). We injected the diffuse noise into each image and selected those with the mean $|z| \leq 1$ in ROI (resulting in the total of 528 images). Comparing this plot with Fig.~\ref{fig:D_histogram} one can notice that the average regularization efficiency for all methods in the case of diffuse noise is a few times worse. The reason is the presence of multiple small clusters within the objects of attention in the images - these clusters would typically contain high attention scores and assigned low $p$- and $l$-values. This is well illustrated in Fig.~\ref{fig:perturbed_image_attention_and_hist_diffuse}. The efficiency of regularization should depend on the size of the noise clusters, however the analysis of this dependence is out of scope of our paper. 

\clearpage
\section{Evaluation with DINOv2}
\label{app:dino_v2}

We conduct a study of how the proposed regularization method works with a different type of ViT. For that we adopt a DINOv2 ViT with a different patch size of $14 \times 14$ pixels \citep{oquab2023dinov2}. 

\begin{figure*}[htb]
\centering
\includegraphics[width=\linewidth]{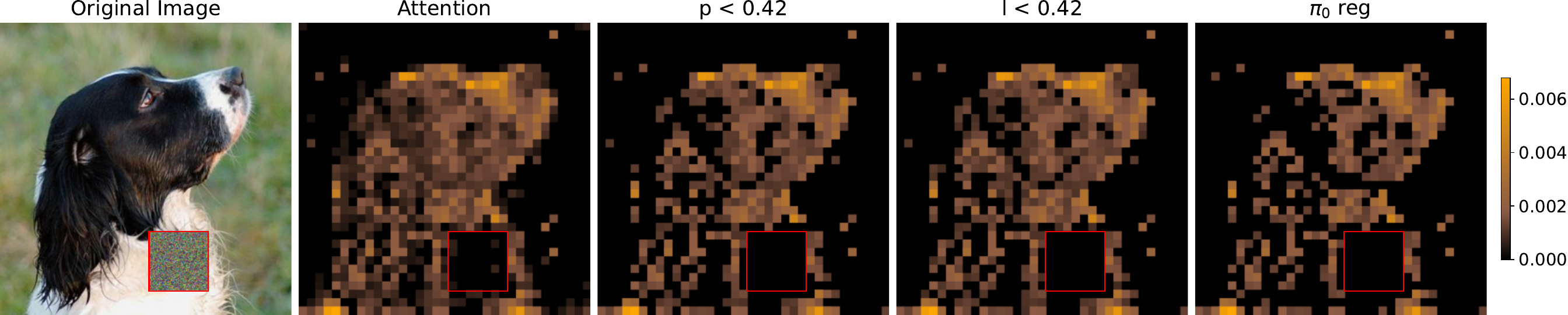}\\[0.5cm] 
\includegraphics[width=\linewidth]{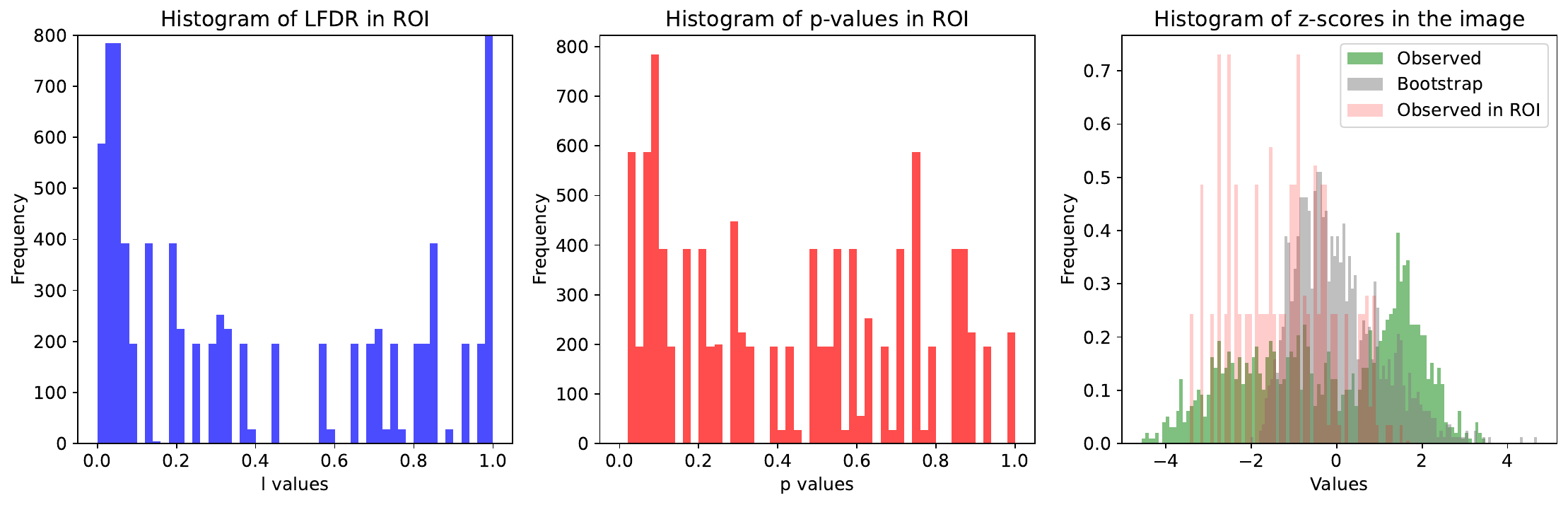}
\caption{Attention map before and after regularization via different shrinkage methods ($p$-thresholding, $l$-thresholding and $\pi_0$-thresholding) for the same image as in Fig.~\ref{fig:perturbed_image_attention_and_hist} (\texttt{n02102040\_821.JPEG}), but processed with a DINOv2 ViT with patch size $14 \times 14$ (top row); Histograms of p-values in ROI, LFDR values in ROI and z-scores of the observed attention scores (green), bootstrap attention scores (gray) and the attention scores observed in ROI (red) corresponding to the image (bottom row). 
The noise patch in the image and attention maps is highlighted with a red frame.}
\label{fig:perturbed_image_attention_and_hist_dino_v2}
\end{figure*}

\begin{figure*}[p]
\centering
\begin{subfigure}{\linewidth}
    \includegraphics[width=\textwidth]{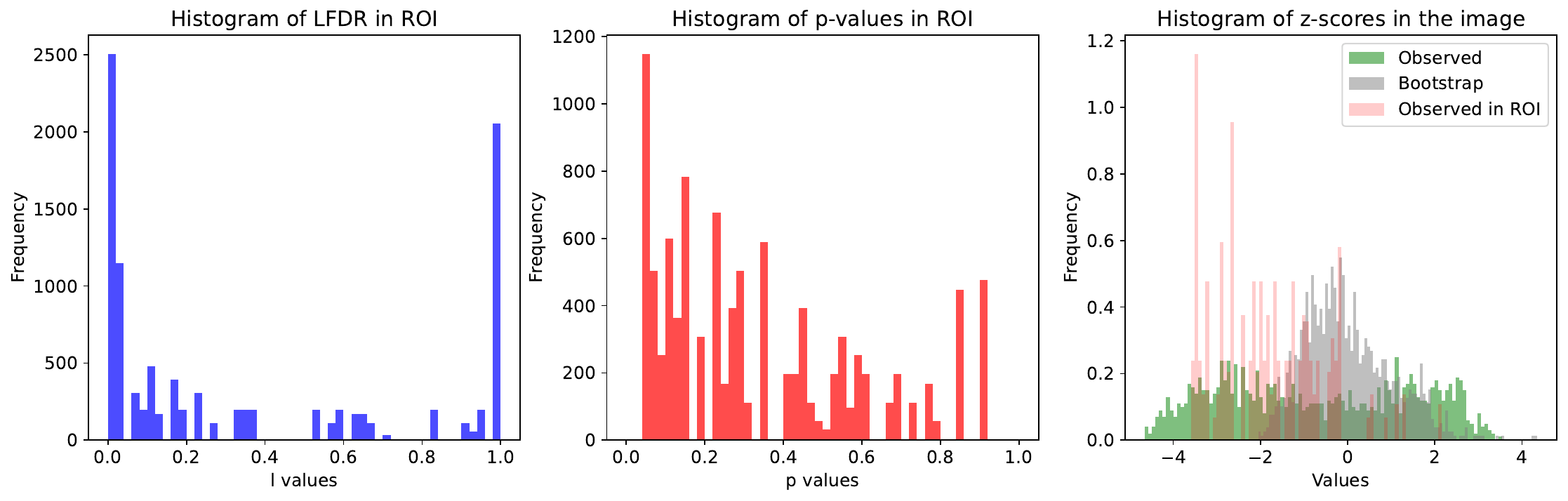}
\end{subfigure}
\vfill
\begin{subfigure}{\linewidth}
    \includegraphics[width=\textwidth]{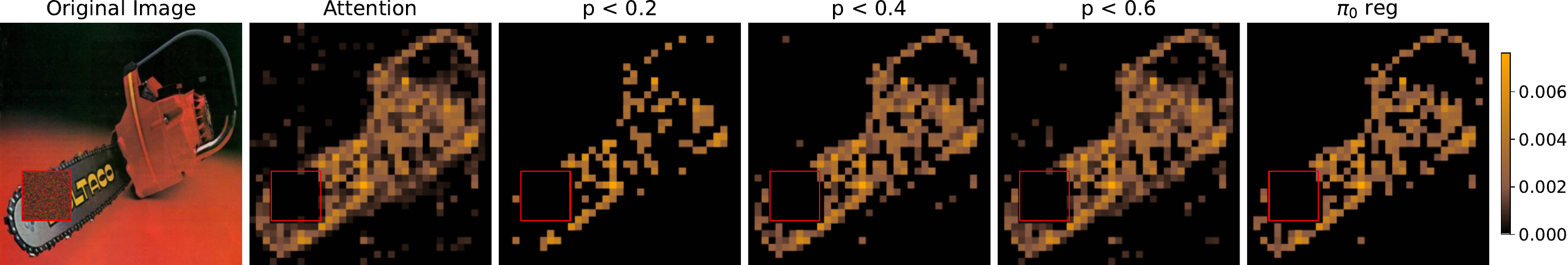}
\end{subfigure}
\vfill
\begin{subfigure}{\linewidth}
    \includegraphics[width=\textwidth]{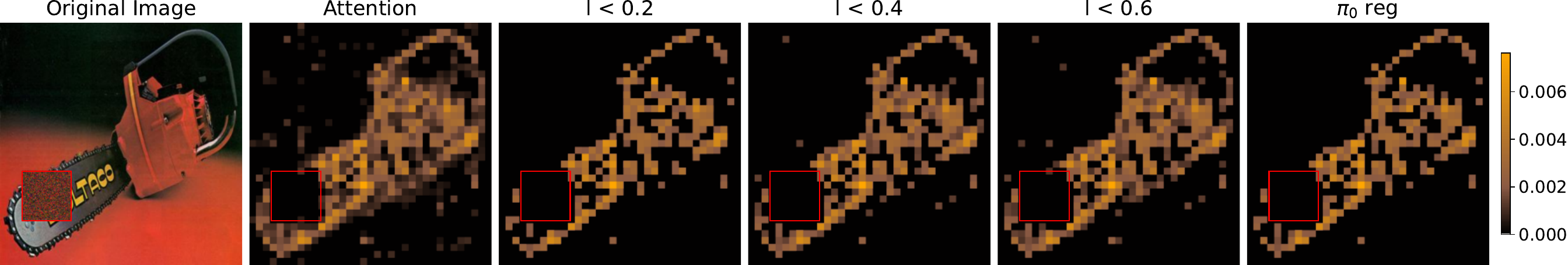}
\end{subfigure}
\caption{Example of a perturbed image (\texttt{n03000684\_5970.JPEG}) with the attention map before and after regularization with the DINO v2 ViT (patch size $14 \times 14$) via $p$-thresholding (middle row) and $l$-thresholding (bottom row) with the respective thresholds varying from $0.2$ to $0.6$. The last attention map in each of the rows to the right corresponds to $\pi_0$-thresholding. The top row shows the histograms of p-values in ROI, LFDR values in ROI and z-scores of the observed attention scores (green), bootstrap attention scores (gray) and the attention scores observed in ROI (red) corresponding to the image. 
The noise patch in the image and attention maps is highlighted with a red frame $(44, 300)$. }
\label{fig:example_chainsaw_dino_v2}
\end{figure*}

\begin{figure*}[h]
    \centering
    \includegraphics[width=0.8\textwidth]{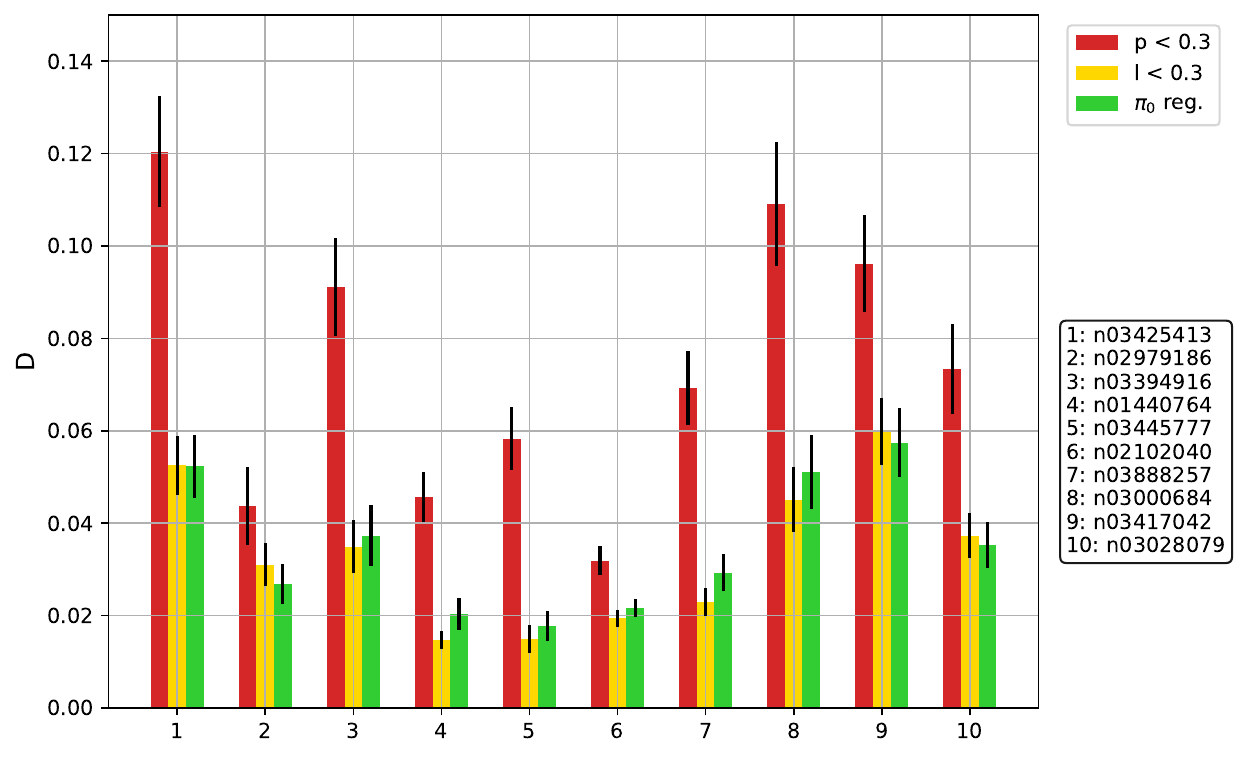}
    \caption{The average suppression factor $D$ for different categories of images from the Imagenette validation subset processed with the DINO-v2 ViT model with patch size 14 and for different shrinkage methods (without $z$-filtering). The relative uncertainty of the results are shown with the black lines.
    }
    \label{fig:D_histogram_dino_v2_no_filters}
\end{figure*}

In Fig.~\ref{fig:perturbed_image_attention_and_hist_dino_v2} we show the attention maps before and after regularization for the same benchmark image and the same threshold values as in Fig.~\ref{fig:perturbed_image_attention_and_hist} (\texttt{n02102040\_821.JPEG}), as well as the corresponding histograms of $l$-, $p$- and $z$-values. Compared to our base ViT model the $z$-scores in ROI are less sparsely distributed because of a larger patch size. In fact, DINOv2 ViT displays a strong tendency to disregard the injected square noise regions, hence the attention scores in ROI are mostly low even prior to regularization (as can be seen e.g. in Fig.~\ref{fig:example_chainsaw_dino_v2}, to be compared with Fig.~\ref{fig:example_chainsaw}). Nevertheless, bootstrapping regularization is efficient in removing the noisy attention scores further, as demonstrated in Fig.~\ref{fig:D_histogram_dino_v2_no_filters}, where we show the average D factors for each category of Imagenette validation dataset. The perturbed images that we included in this study were not filtered by the mean $|z|$ in ROI.  

\clearpage
\section{Evaluations on the medical image dataset}
\label{app:applications}

We apply our regularization method to the images from the IQ-OTH/NCCD dataset as described in Section~\ref{sec:medical_data_applications}. We use the same ViT model that we use for noise simulation study. 

\begin{figure}[h!]
    \centering
    \begin{subfigure}{0.49\textwidth}
    \centering
        \includegraphics[width=\textwidth]{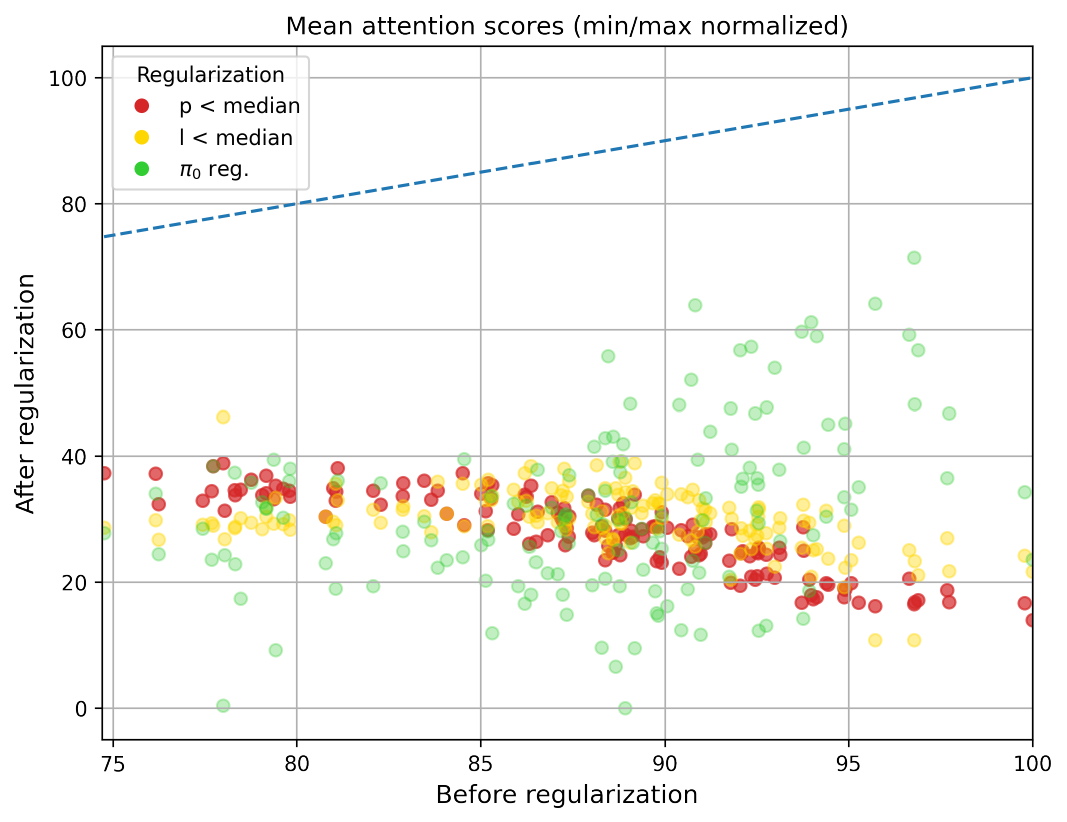}
        \caption{}
    \end{subfigure}
    \hfill
    \begin{subfigure}{0.48\textwidth}
        \includegraphics[width=\textwidth]{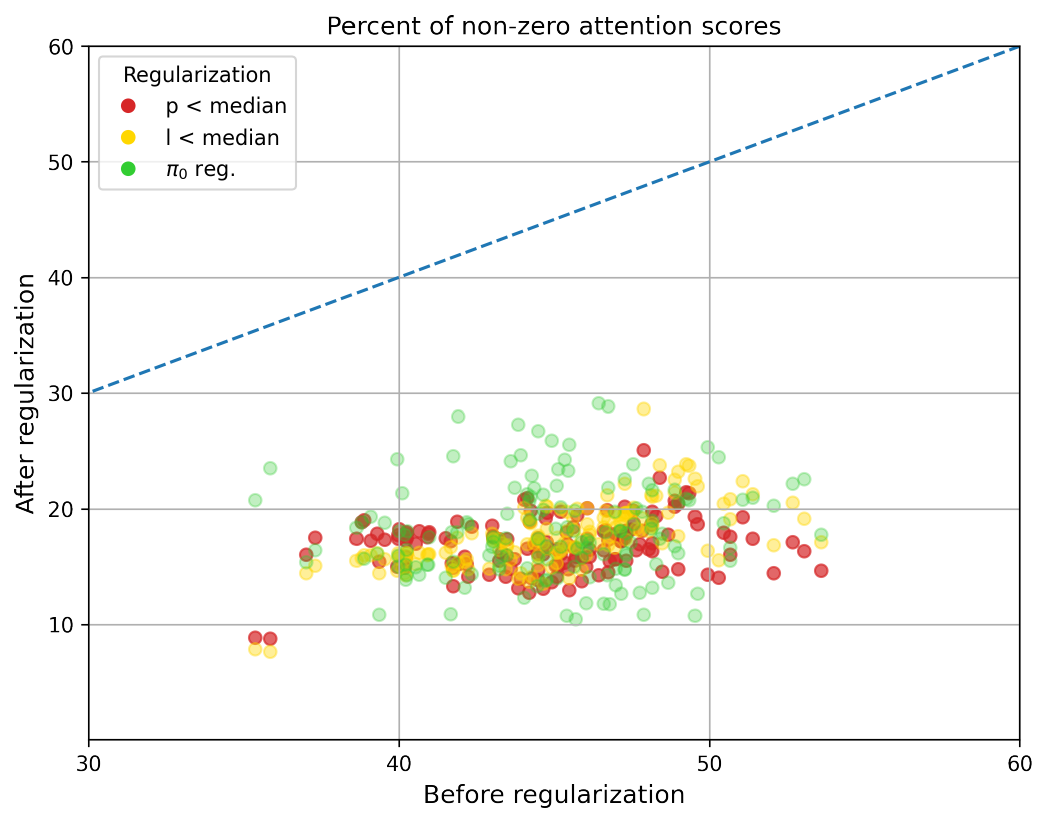}
        \caption{}
    \end{subfigure}
    \caption{Regularization efficiency in for various images from the IQ-OTH/NCCD dataset expressed in terms of: a) mean attention scores before and after regularization scaled via min-max normalization b) percentage of non-zero attention scores before and after regularization. Each dot denotes the results of using one method for one perturbed image: $p$-thresholding (red), $l$-thresholding (yellow), $\pi_0$-thresholding (green). For each point we use the corresponding median of the $p$-value and LFDR distributions as the respective regularization threshold. The blue dashed line corresponds to the same metric values before and after regularization.}
    \label{fig:reg_efficiency_medical}
\end{figure}

In Fig.~\ref{fig:reg_efficiency_medical} we show the results of the regularization that we perform on 50 random images from each category (benign, malignant and normal): left plot shows the mean attention scores before and after regularization rescaled via min-max normalization for three different methods, right plot shows the percentage of non-zero attention scores before and after regularization. The blue dashed line corresponds to the same values before and after regularization. For $p$- and $l$-thresholds we use the median values of the respective statistic distributions. 

\begin{figure*}[hb!]
    \centering
    \includegraphics[width=\scaleone\linewidth]{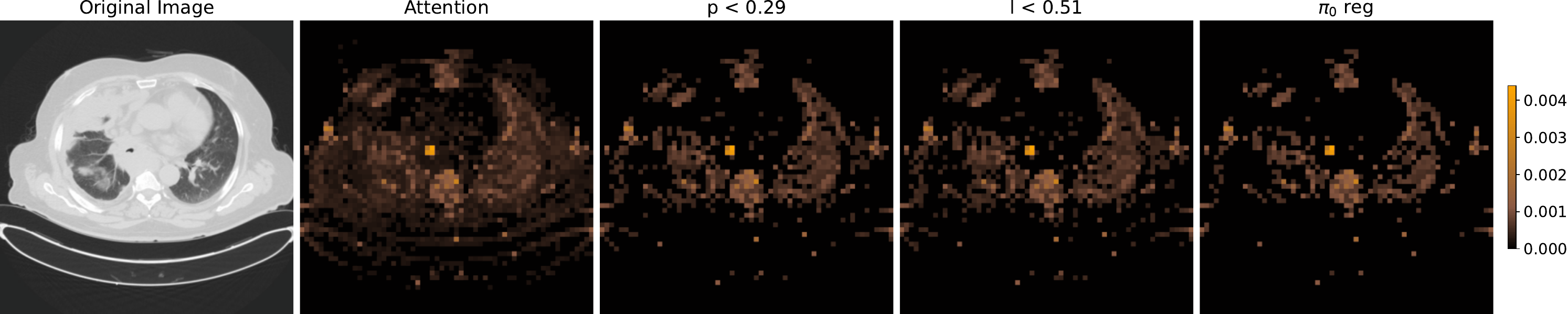}
    \vfill
    \includegraphics[width=\scaleone\linewidth]{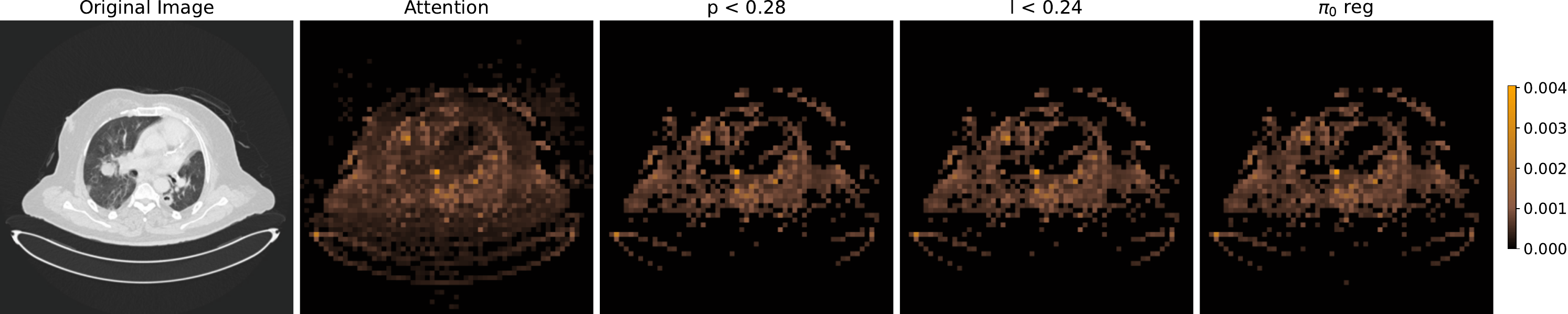}
    \vfill
    \includegraphics[width=\scaleone\linewidth]{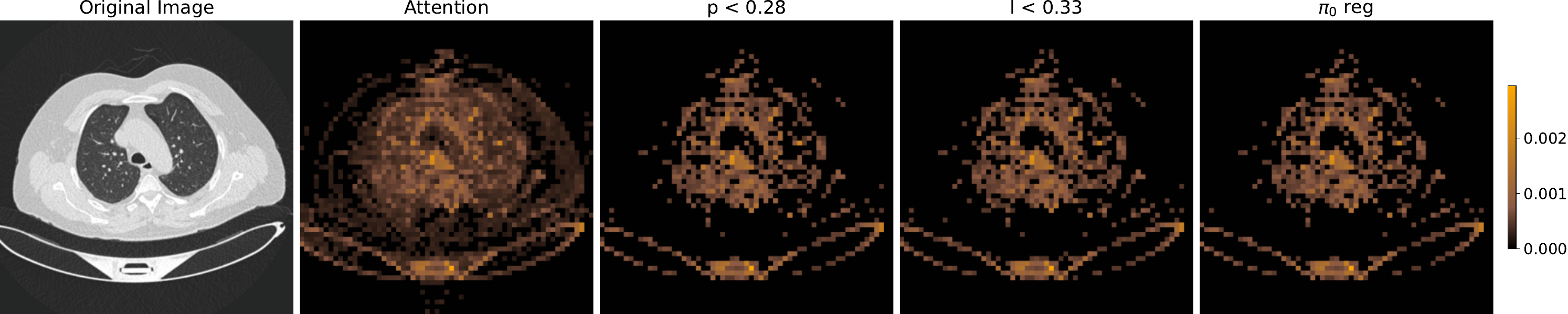}
    \vfill
    \includegraphics[width=\scaleone\linewidth]{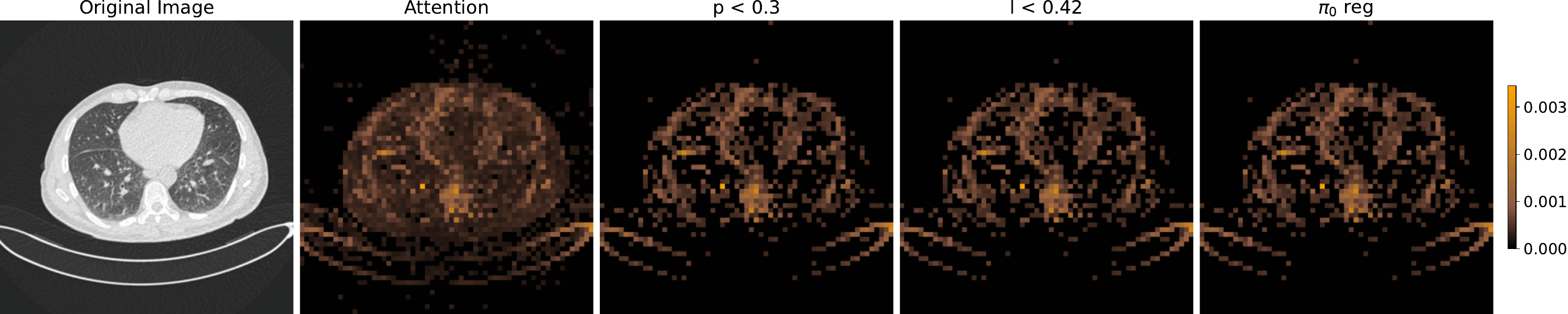}
    \vfill
    \includegraphics[width=\scaleone\linewidth]{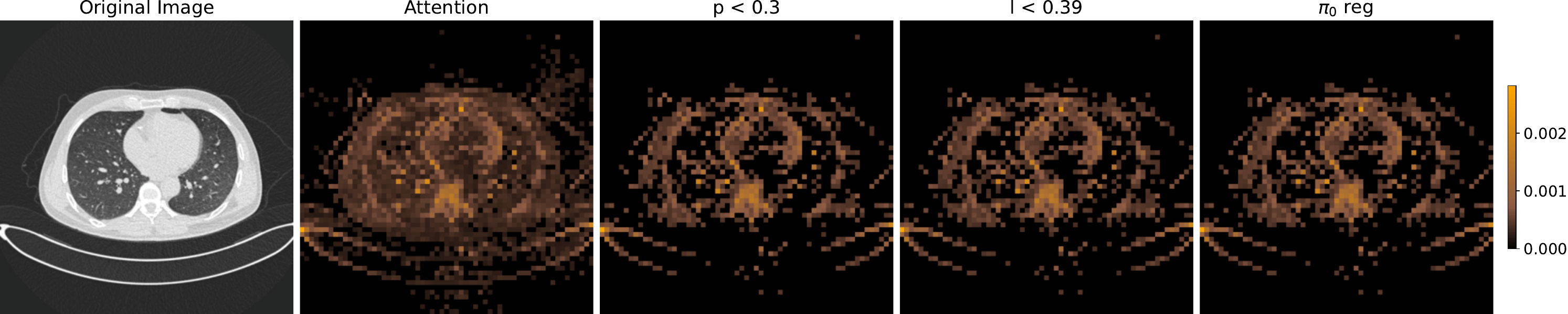}
    \vfill
    \includegraphics[width=\scaleone\linewidth]{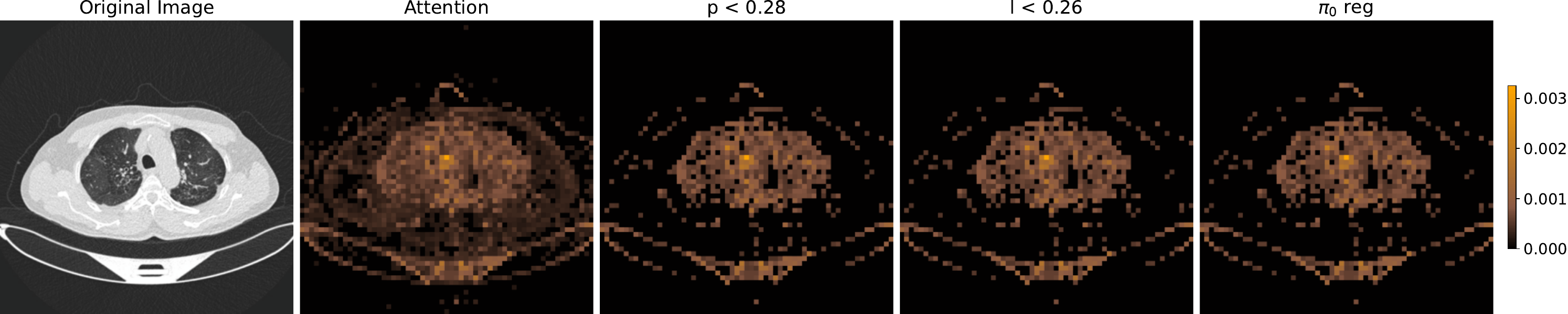}
    \caption{The examples of applying bootstrap regularization to different images from the IQ-OTH/NCCD dataset from top to bottom: malignant case (23 and 90), benign case (28 and 44) and normal case (9 and 18). The original image is displayed on the left of each row and is succeeded by the attention map before regularization and the attentions maps after: $p$-thresholding, $l$-thresholding and $\pi_0$-thresholding regularizations. For $p$- and $l$-thresholds we use the median values of the respective statistic distributions. }
    \label{fig:reg_efficiency_applications}
\end{figure*}

In Fig.~\ref{fig:reg_efficiency_applications} we demonstrate several examples of attention maps for different images from the considered dataset before and after regularization via different methods. Each pair of images is taken from each subset, namely malignant, benign and normal in the presented order from top to bottom. For $p$- and $l$-thresholds we again use the median values of the respective statistic distributions. 

\begin{figure*}[p]
\centering
\begin{subfigure}{\linewidth}
    \includegraphics[width=\textwidth]{Plots/image_histograms_mal_case21.pdf}
\end{subfigure}
\vfill
\begin{subfigure}{\linewidth}
    \includegraphics[width=\textwidth]{Plots/mal_case21_reg_p.pdf}
\end{subfigure}
\vfill
\begin{subfigure}{\linewidth}
    \includegraphics[width=\textwidth]{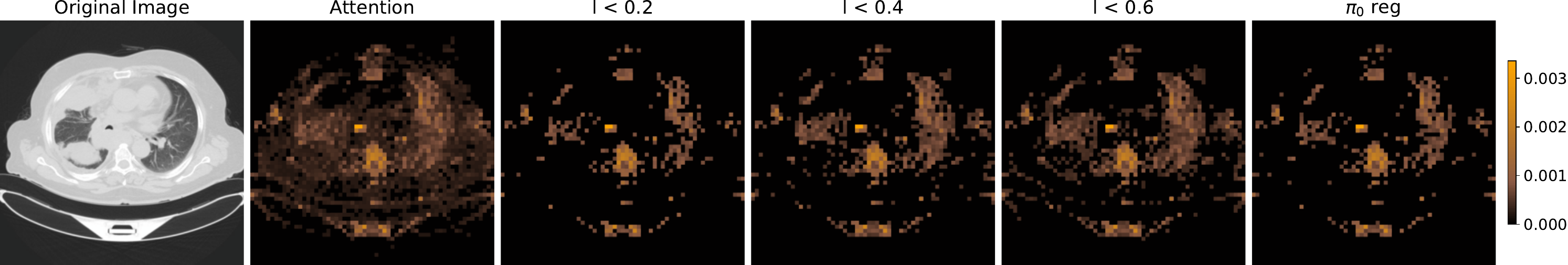}
\end{subfigure}
\vfill
\begin{subfigure}{\linewidth}
    \includegraphics[width=\textwidth]{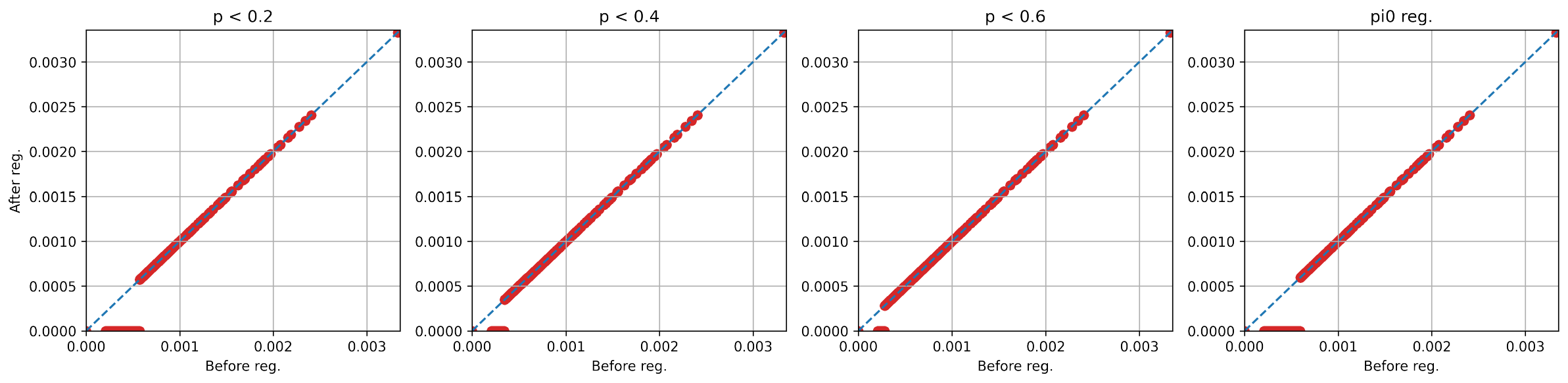}
\end{subfigure}
\vfill
\begin{subfigure}{\linewidth}
    \includegraphics[width=\textwidth]{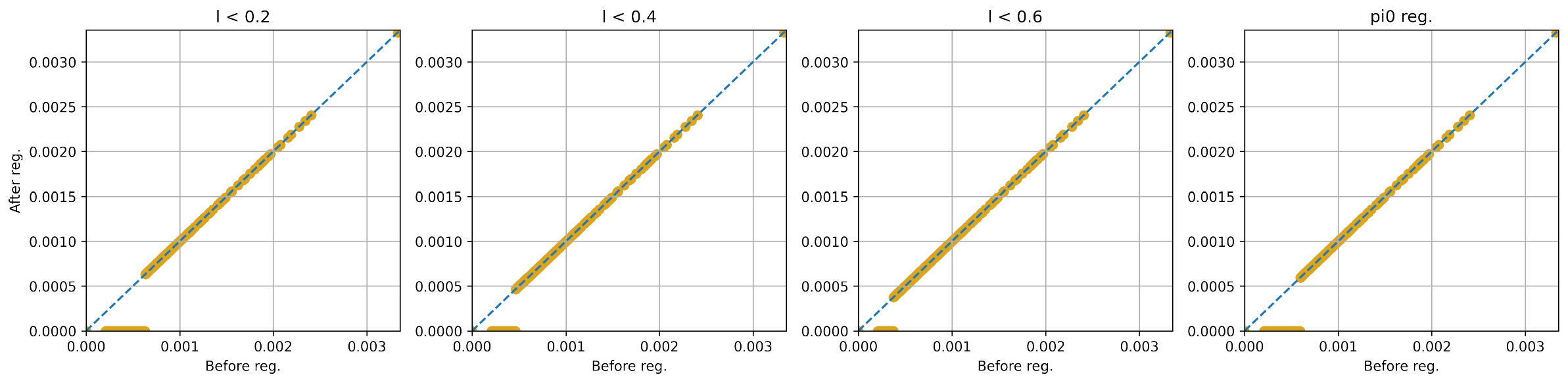}
\end{subfigure}
\caption{Results of applying $p$-thresholding (second row) and $l$-thresholding (third row) regularizations to the malignant case 21 image with the respective thresholds varying from $0.2$ to $0.6$. The last attention map in each of the rows to the right corresponds to $\pi_0$-thresholding. The top row shows the histograms of p-values, LFDR values and z-scores of the observed attention scores (green) and bootstrap attention scores (gray). The two bottom rows display the attention scores after regularization vs. before regularization for the regularized attention maps considered above, for $p$-thresholding (red) and $l$-thresholding (gold). The blue dashed line corresponds to the same attention score values before and after regularization.}
\label{fig:example_malignant_21}
\end{figure*}

\begin{figure*}[p]
\centering
\begin{subfigure}{\linewidth}
    \includegraphics[width=\textwidth]{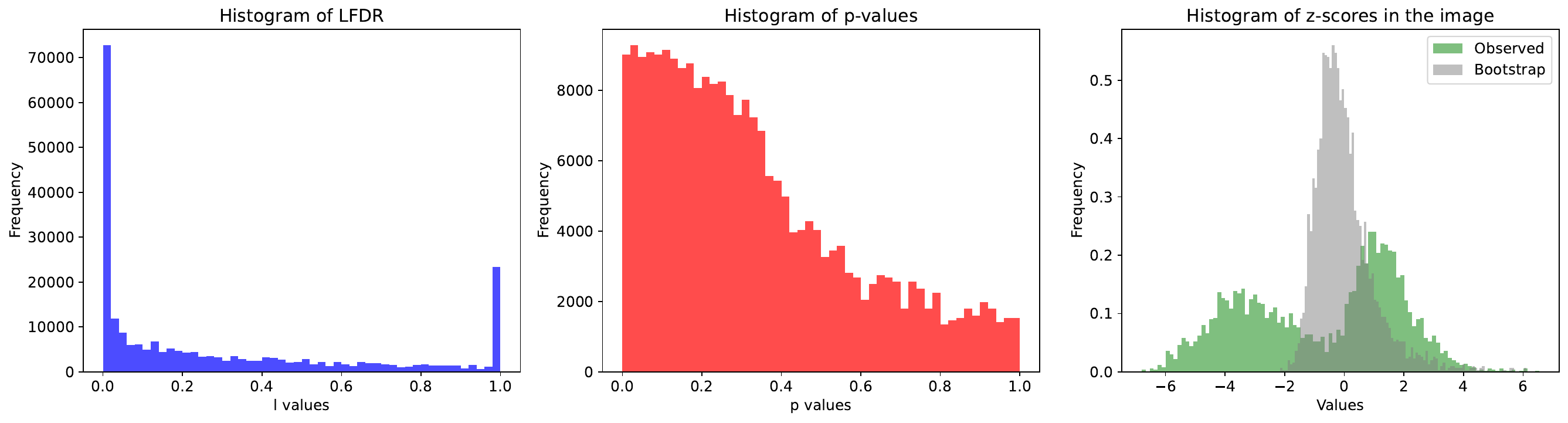}
\end{subfigure}
\vfill
\begin{subfigure}{\linewidth}
    \includegraphics[width=\textwidth]{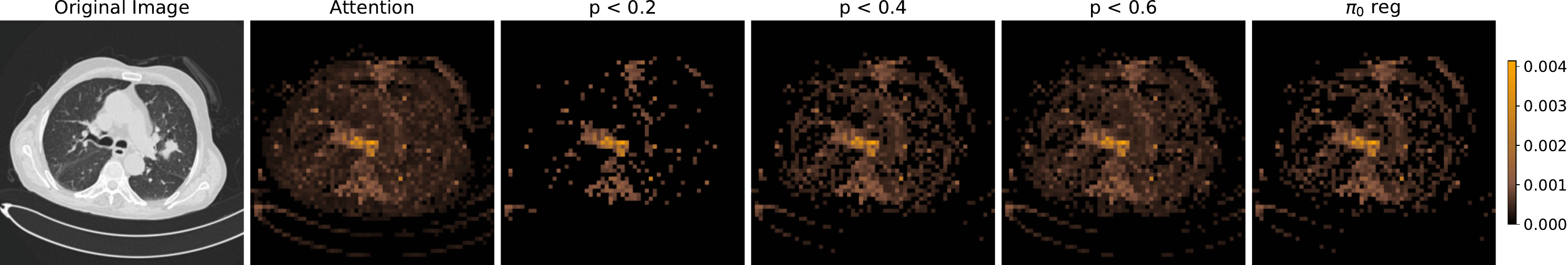}
\end{subfigure}
\vfill
\begin{subfigure}{\linewidth}
    \includegraphics[width=\textwidth]{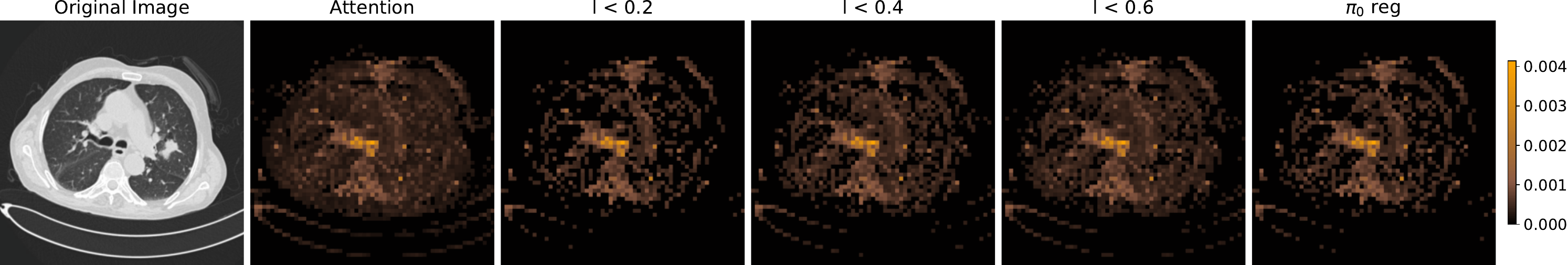}
\end{subfigure}
\vfill
\begin{subfigure}{\linewidth}
    \includegraphics[width=\textwidth]{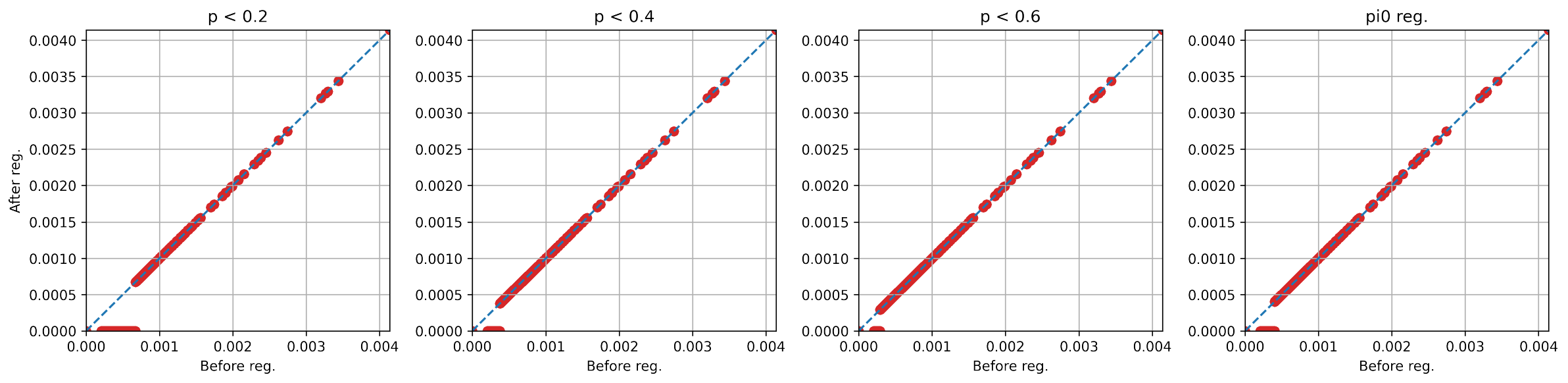}
\end{subfigure}
\vfill
\begin{subfigure}{\linewidth}
    \includegraphics[width=\textwidth]{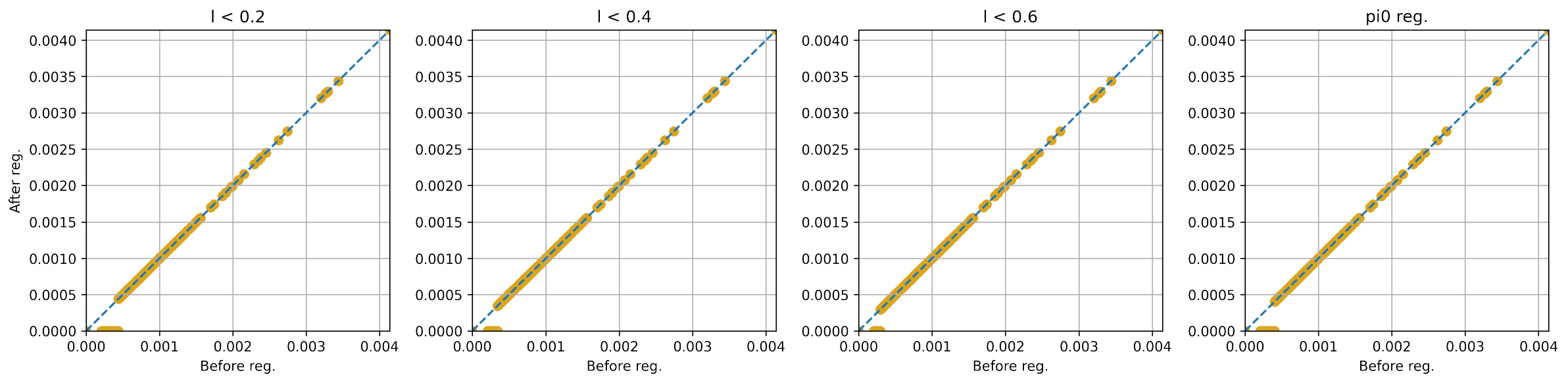}
\end{subfigure}
\caption{Results of applying $p$-thresholding (second row) and $l$-thresholding (third row) regularizations to the malignant case 73 image with the respective thresholds varying from $0.2$ to $0.6$. The last attention map in each of the rows to the right corresponds to $\pi_0$-thresholding. The top row shows the histograms of p-values, LFDR values and z-scores of the observed attention scores (green) and bootstrap attention scores (gray). The two bottom rows display the attention scores after regularization vs. before regularization for the regularized attention maps considered above, for $p$-thresholding (red) and $l$-thresholding (gold). The blue dashed line corresponds to the same attention score values before and after regularization.}
\label{fig:example_malignant_73}
\end{figure*}

\begin{figure*}[p]
\centering
\begin{subfigure}{\linewidth}
    \includegraphics[width=\textwidth]{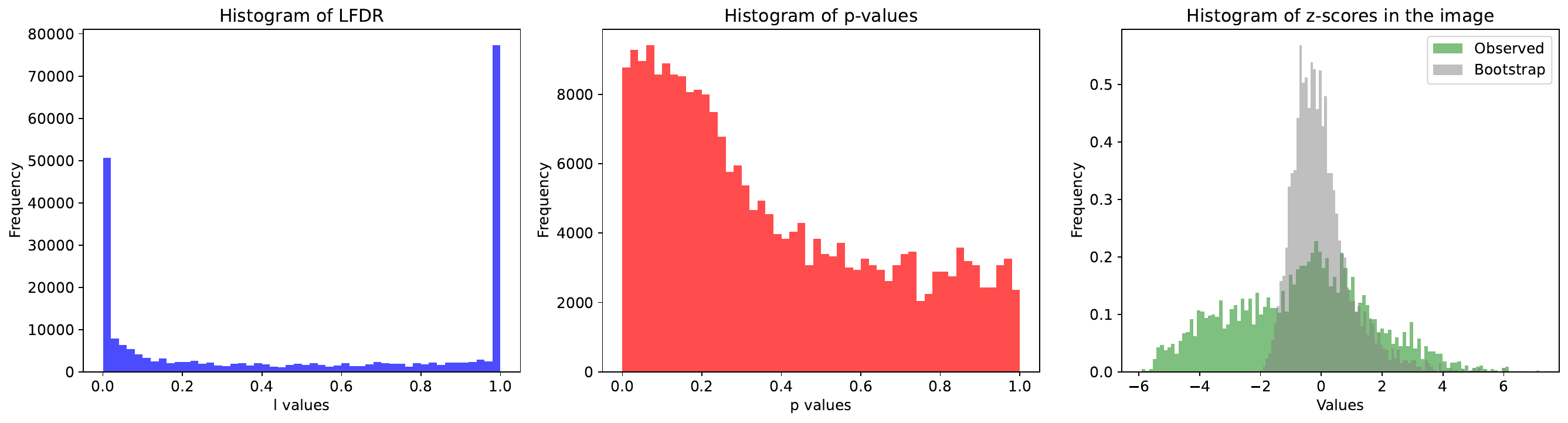}
\end{subfigure}
\vfill
\begin{subfigure}{\linewidth}
    \includegraphics[width=\textwidth]{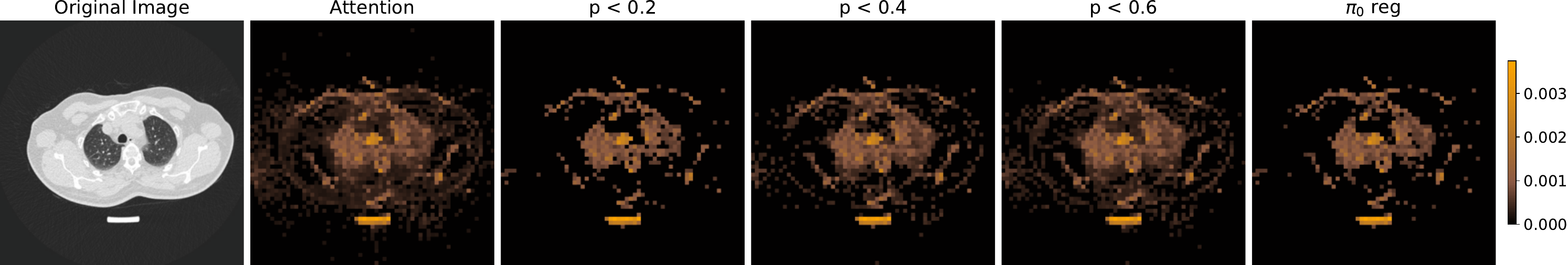}
\end{subfigure}
\vfill
\begin{subfigure}{\linewidth}
    \includegraphics[width=\textwidth]{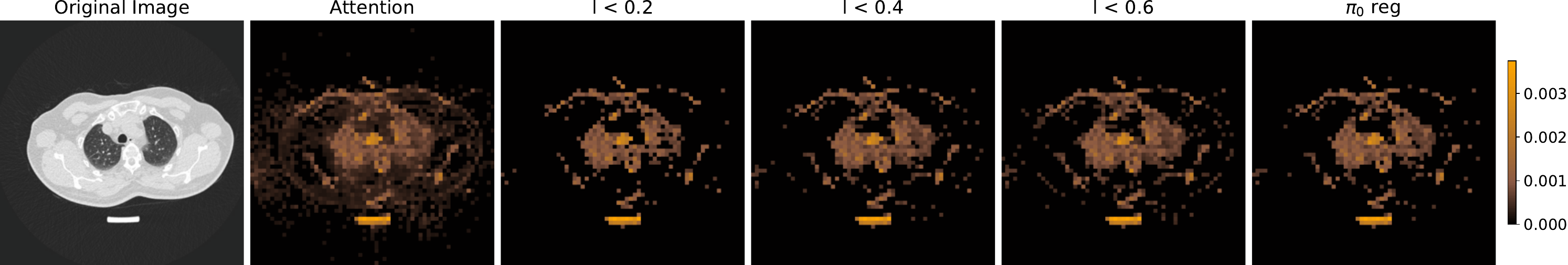}
\end{subfigure}
\vfill
\begin{subfigure}{\linewidth}
    \includegraphics[width=\textwidth]{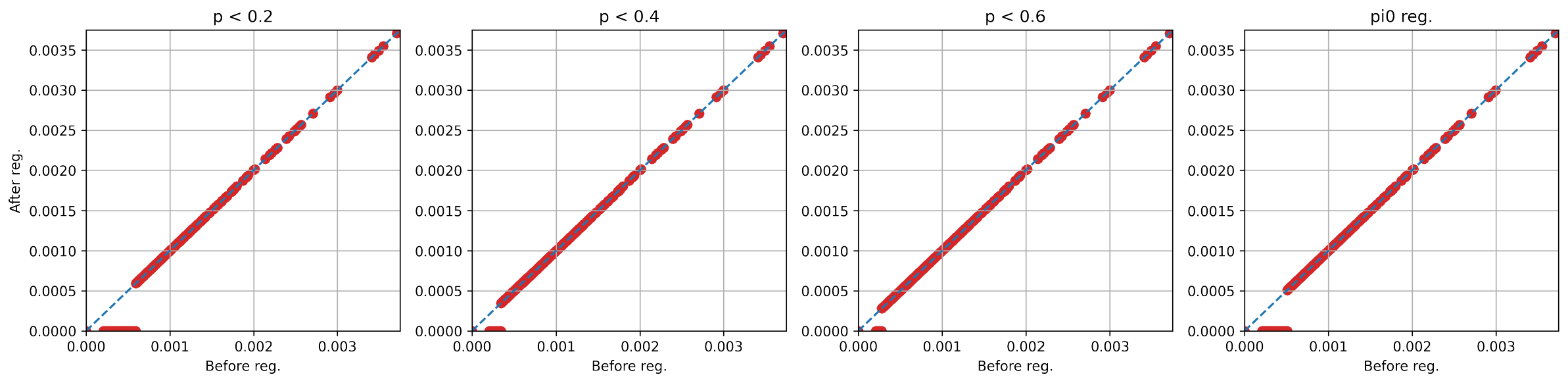}
\end{subfigure}
\vfill
\begin{subfigure}{\linewidth}
    \includegraphics[width=\textwidth]{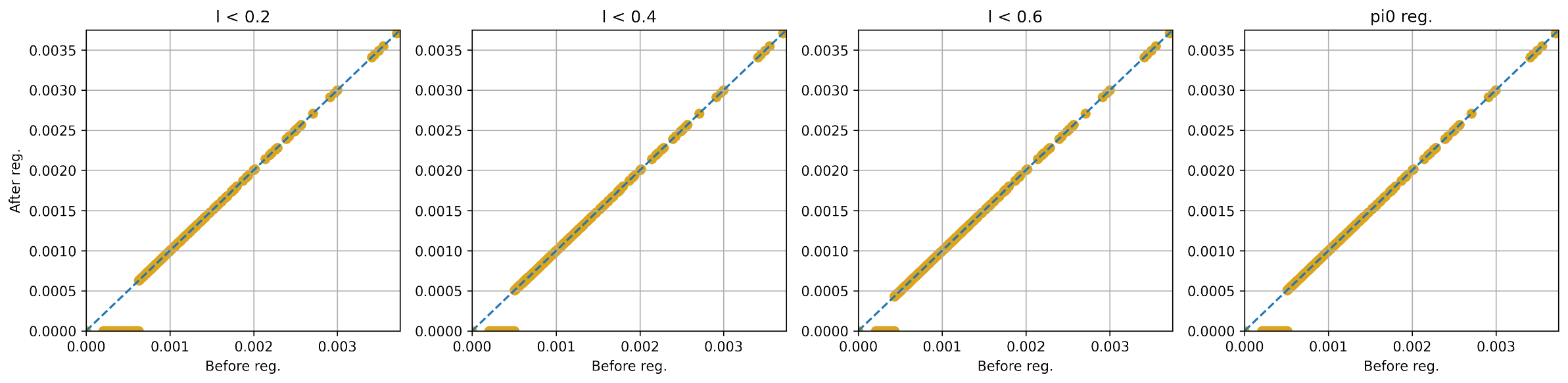}
\end{subfigure}
\caption{Results of applying $p$-thresholding (second row) and $l$-thresholding (third row) regularizations to the benign case 5 image with the respective thresholds varying from $0.2$ to $0.6$. The last attention map in each of the rows to the right corresponds to $\pi_0$-thresholding. The top row shows the histograms of p-values, LFDR values and z-scores of the observed attention scores (green) and bootstrap attention scores (gray). The two bottom rows display the attention scores after regularization vs. before regularization for the regularized attention maps considered above, for $p$-thresholding (red) and $l$-thresholding (gold). The blue dashed line corresponds to the same attention score values before and after regularization.}
\label{fig:example_benign_5}
\end{figure*}

\begin{figure*}[p]
\centering
\begin{subfigure}{\linewidth}
    \includegraphics[width=\textwidth]{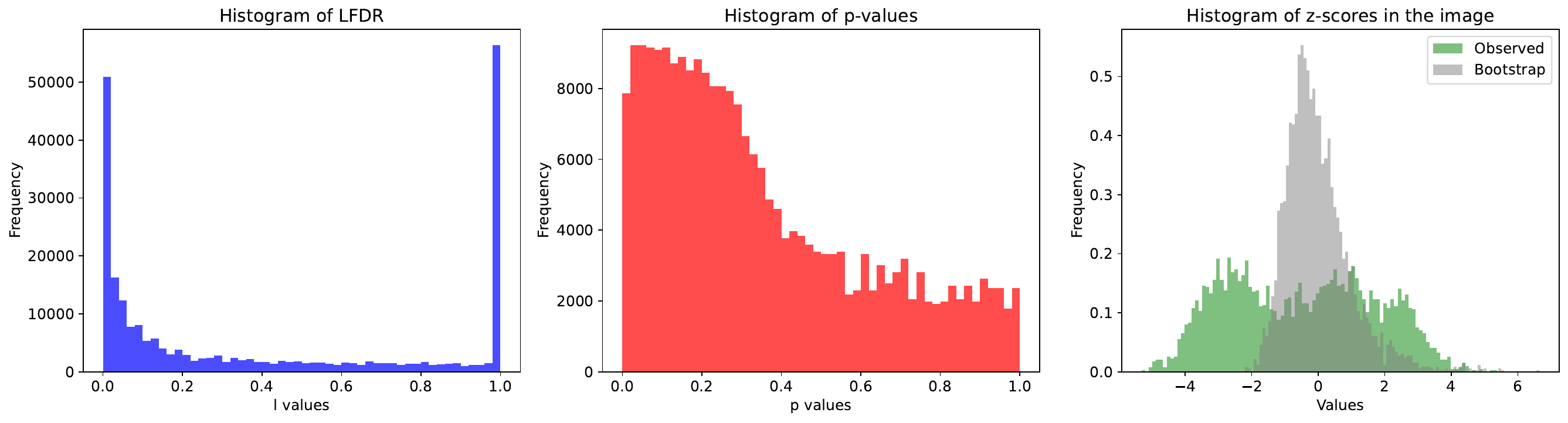}
\end{subfigure}
\vfill
\begin{subfigure}{\linewidth}
    \includegraphics[width=\textwidth]{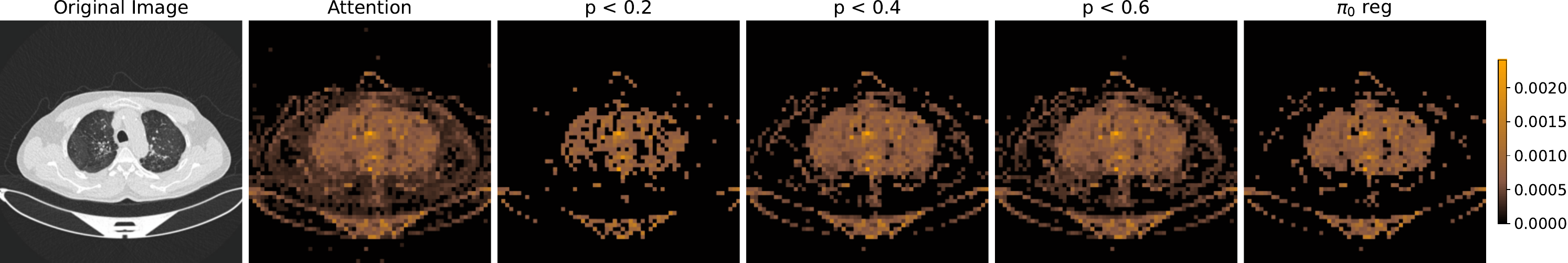}
\end{subfigure}
\vfill
\begin{subfigure}{\linewidth}
    \includegraphics[width=\textwidth]{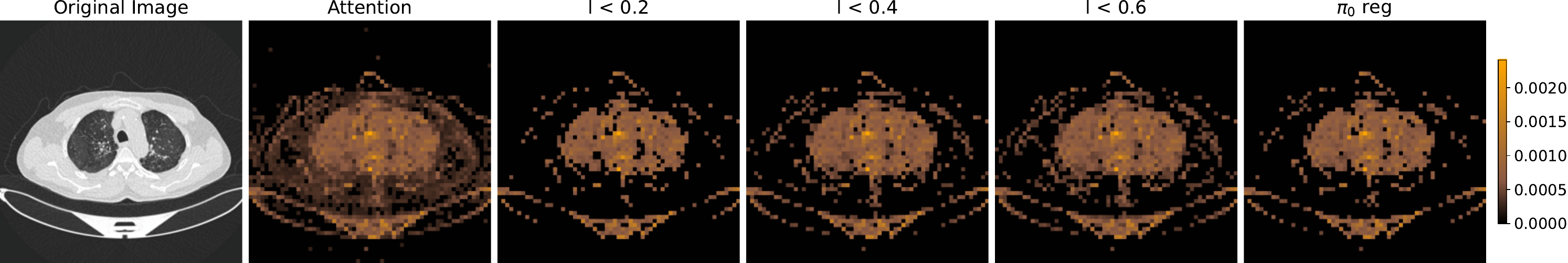}
\end{subfigure}
\vfill
\begin{subfigure}{\linewidth}
    \includegraphics[width=\textwidth]{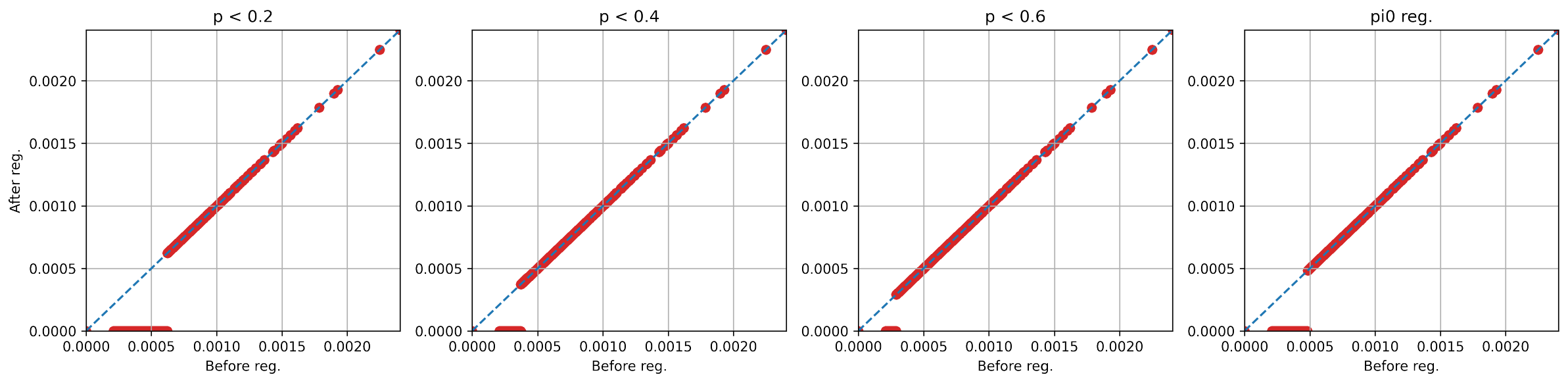}
\end{subfigure}
\vfill
\begin{subfigure}{\linewidth}
    \includegraphics[width=\textwidth]{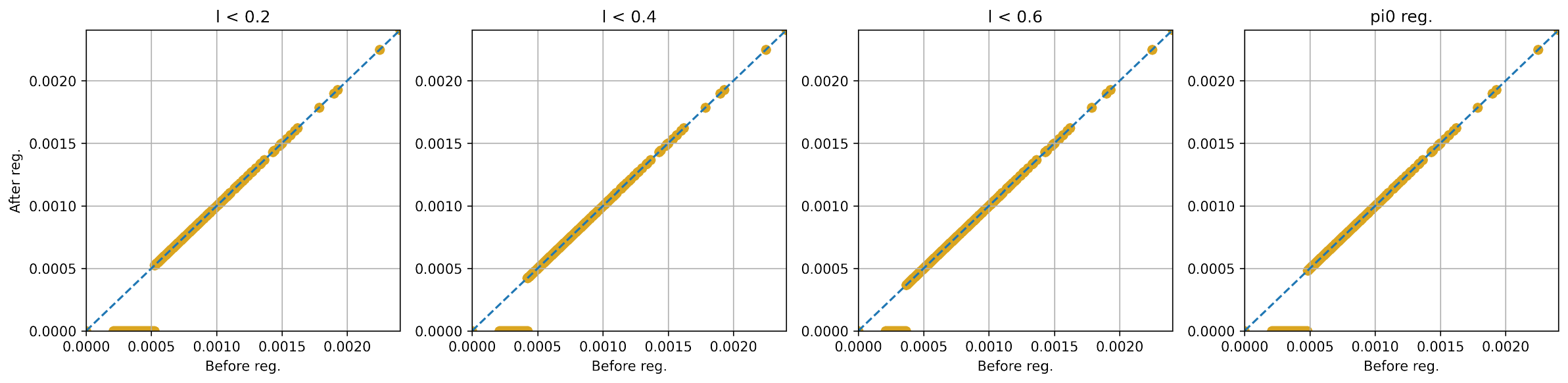}
\end{subfigure}
\caption{Results of applying $p$-thresholding (second row) and $l$-thresholding (third row) regularizations to the normal case 17 image with the respective thresholds varying from $0.2$ to $0.6$. The last attention map in each of the rows to the right corresponds to $\pi_0$-thresholding. The top row shows the histograms of p-values, LFDR values and z-scores of the observed attention scores (green) and bootstrap attention scores (gray). The two bottom rows display the attention scores after regularization vs. before regularization for the regularized attention maps considered above, for $p$-thresholding (red) and $l$-thresholding (gold). The blue dashed line corresponds to the same attention score values before and after regularization.}
\label{fig:example_normal_17}
\end{figure*}

In Figs.~\ref{fig:example_malignant_21}--\ref{fig:example_normal_17} we present the examples of regularization using different $p$- and $l$-thresholds for medical images similar to what we display in Figs.~\ref{fig:example_chainsaw}--\ref{fig:example_parasail}. In addition for each example we also show the distribution of attention scores before and after regularization for different $p$- and $l$-thresholds.

\end{document}